\documentclass[11pt, a4paper, logo, copyright]{deepmind}

\usepackage[authoryear, sort&compress, round]{natbib}
\usepackage{subcaption}[skip=-10pt]
\usepackage{svg}
\usepackage{graphicx}
\usepackage{longtable}
\usepackage{cleveref} 
\usepackage{chngcntr}
\usepackage{multirow} 
\usepackage{array}

\usepackage{amsmath,amsfonts,bm}

\title{Human-Timescale Adaptation in an Open-Ended Task Space}

\correspondingauthor{Feryal Behbahani (\href{mailto:feryal@deepmind.com}{feryal@deepmind.com}) and Edward Hughes (\href{mailto:edwardhughes@deepmind.com}{edwardhughes@deepmind.com}).\linebreak 
}

\reportnumber{001} 

\renewcommand{\today}{}

\author[1]{\hyperref[sec:authors]{Adaptive Agents Team}}
\affil[1]{DeepMind}

\begin{abstract}

Foundation models have shown impressive adaptation and scalability in supervised and self-supervised learning problems, but so far these successes have not fully translated to reinforcement learning (RL). In this work, we demonstrate that training an RL agent at scale leads to a general in-context learning algorithm that can adapt to open-ended novel embodied 3D problems as quickly as humans.  In a vast space of held-out environment dynamics, our adaptive agent (AdA) displays on-the-fly hypothesis-driven exploration, efficient exploitation of acquired knowledge, and can successfully be prompted with first-person demonstrations. Adaptation emerges from three ingredients: (1) meta-reinforcement learning across a vast, smooth and diverse task distribution, (2) a policy parameterised as a large-scale attention-based memory architecture, and (3) an effective automated curriculum that prioritises tasks at the frontier of an agent's capabilities. We demonstrate characteristic scaling laws with respect to network size, memory length, and richness of the training task distribution. We believe our results lay the foundation for increasingly general and adaptive RL agents that perform well across ever-larger open-ended domains.  

\end{abstract}

\begin{document}
\maketitle

\begin{center}
  \includegraphics[width=1\textwidth]{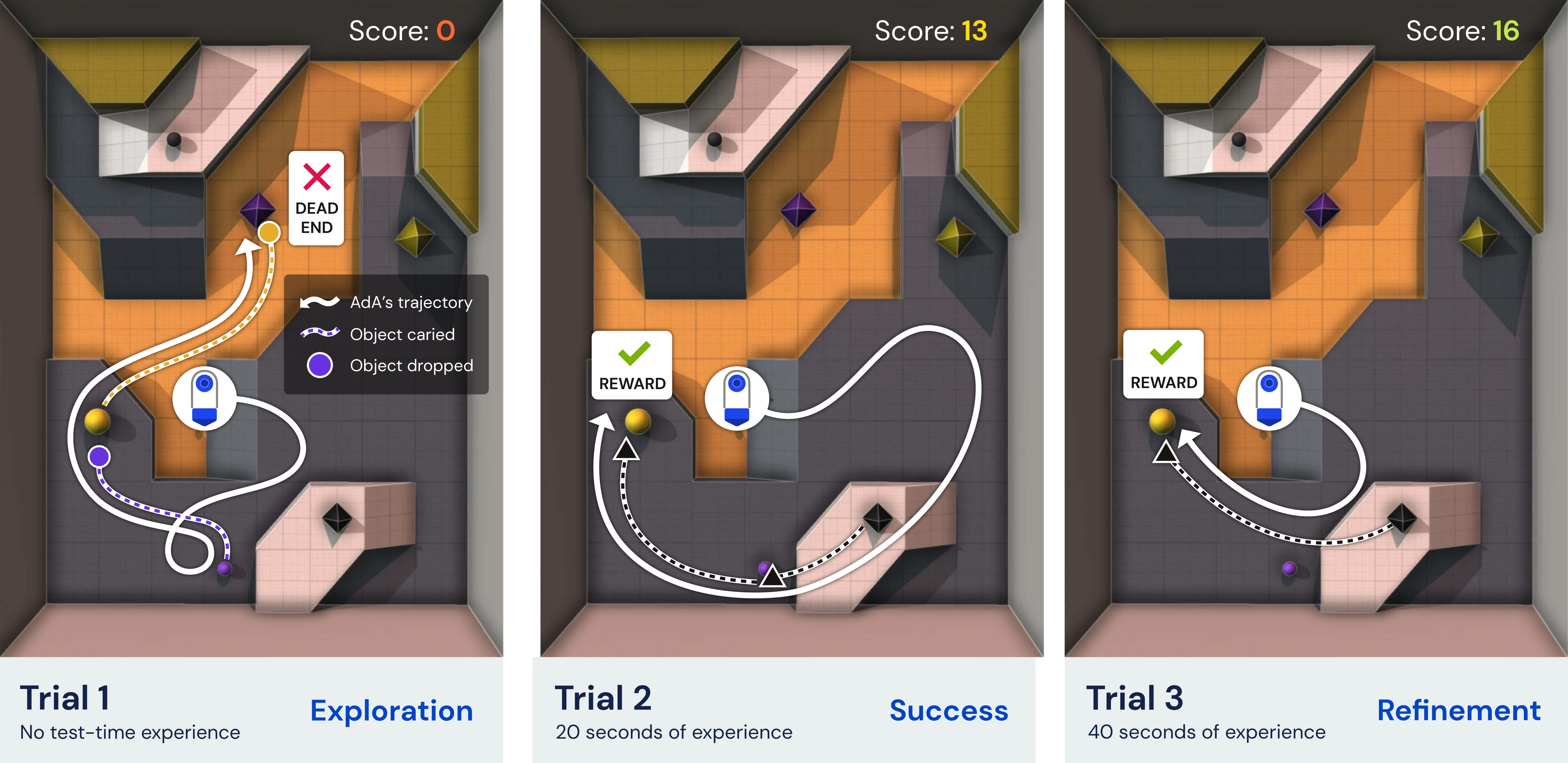}
  \captionof{figure}{\textbf{Human timescale adaptation.} Example trajectories of our agent (AdA) solving a held-out task in a complex
  3D environment within minutes of test-time experience without any further agent training. Initial trials (\textbf{Exploration}) show a policy that uncovers hidden environment dynamics. After just seconds of test-time experience (\textbf{Success}), AdA finds a valid solution to the task. Later (\textbf{Refinement}), it improves this solution, gradually finding a more rewarding behaviour. The solid white lines show agent movement. The dashed coloured lines show the agent carrying an object of the corresponding colour. For a full description of the task, see Figure \ref{fig:wrong-pair-disappears-explained}. Videos of AdA's behaviour are available on our \href{http://sites.google.com/view/adaptive-agent/}{microsite} and accompanying \href{https://youtu.be/U93bUQ1roiw}{results reel}.}
  \label{fig:intro-pyramid-in-a-haystack-explained}
\end{center}

\newpage
\section{Introduction} 

The ability to adapt in minutes is a defining characteristic of human intelligence and an important milestone on the path towards general intelligence. Given any level of bounded rationality, there will be a space of tasks in which it is impossible for agents to succeed by just generalising their policy zero-shot, but where progress is possible if the agent is capable of very fast in-context learning from feedback. To be useful in the real world, and in interaction with humans, our artificial agents should be capable of fast and flexible adaptation given only a few interactions, and should continue to adapt as more data becomes available. Operationalising this notion of adaptation, we seek to train an agent that, given few episodes in an unseen environment at test time, can accomplish a task that requires trial-and-error exploration and can subsequently refine its solution towards optimal behaviour.

Meta-RL has been shown to be effective for fast in-context adaptation (e.g. \citet{yu2020learning, zintgraf2022fast}). However, meta-RL has had limited success in settings where the reward is sparse and the task space is vast and diverse \citep{yang2019single}. Outside RL, \emph{foundation models} in semi-supervised learning have generated significant interest \citep{DBLP:journals/corr/abs-2108-07258} due to their ability to adapt in few shots from demonstrations across a broad range of tasks. These models are designed to provide a strong foundation of general knowledge and skills that can be built upon and adapted to new situations via fine-tuning or prompting with demonstrations~\citep{brown2020gpt3}. Crucial to this success has been attention-based memory architectures like Transformers \citep{vaswani2017attention}, which show power-law scaling in performance with the number of parameters \citep{https://doi.org/10.48550/arxiv.2207.10551}.

\begin{figure}[htb]
    \centering
    \includegraphics[width=\linewidth]{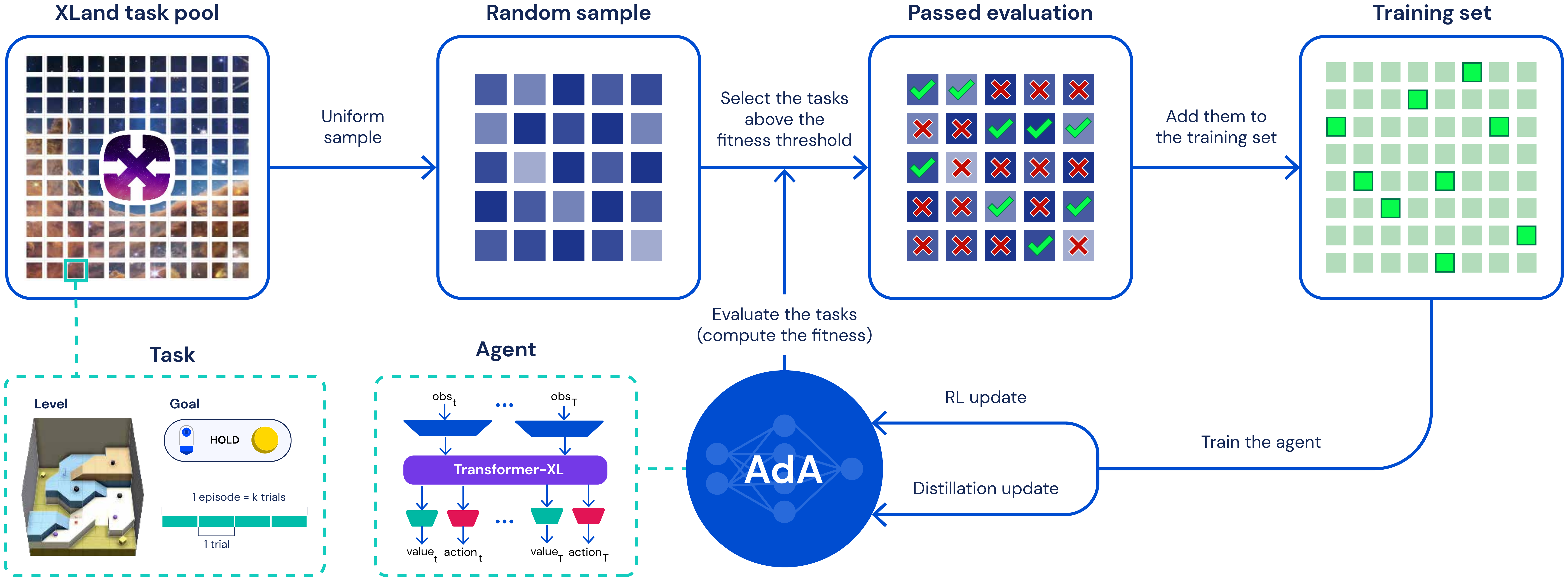}
    \caption{
        \textbf{Training our Adaptive Agent (AdA).} We train a large Transformer model with meta-RL in XLand. During training, tasks are uniformly sampled, and subsequently filtered to produce an ever-changing training pool of tasks at the frontier of the agent's capabilities. After training on these tasks, the agent is capable of adapting to unseen hand-authored tasks as effectively and efficiently as humans.}
    \label{fig:method_overview}
\end{figure}

In this work, we pave the way for training an RL foundation model; that is, an agent that has been pre-trained on a vast task distribution and that, at test time, can adapt few-shot to a broad range of downstream tasks. We introduce \emph{Adaptive Agent} (AdA), an agent capable of human-timescale adaptation in a vast open-ended task space with sparse rewards. AdA does not require any prompts \citep{reed2022a}, fine-tuning \citep{lee2022multigame} or access to offline datasets \citep{laskin2022algorithmdistillation, reed2022a}. Instead, AdA exhibits hypothesis-driven exploratory behaviour, using information gained on-the-fly to refine its policy and to achieve close to optimal performance. AdA acquires knowledge efficiently, adapting in minutes on challenging held-out sparse-reward tasks in a partially-observable 3D environment with a first-person pixel observation. A human study confirms that the timescale of AdA's adaptation is comparable to that of trained human players. AdA's adaptation behaviour in a representative held-out task can be seen in Figure \ref{fig:intro-pyramid-in-a-haystack-explained}. AdA can also achieve improved performance through zero-shot prompting with first-person demonstrations, analogously to foundation models in the language domain.

We use Transformers as an architectural choice to scale in-context fast adaptation via model-based RL${^2}$ \citep{duan2017rl, wang2016learning, melo2022transformers}. Foundation models typically require large, diverse datasets to achieve their generality \citep{8237359, mahajan2018, brown2020gpt3, 9880094, schuhmann2022laionb}. To make this possible in an RL setting, where agents collect their own data, we extend the recent XLand environment \citep{xland}, producing a vast open-ended world with over $10^{40}$ possible tasks. These tasks require a range of different online adaptation capabilities, including experimentation, navigation, coordination, division of labour and coping with irreversibility. Given the wide range of possible tasks, we make use of adaptive auto-curricula, which prioritise tasks at the frontier of an agent's capabilities~\citep{xland, jiang2021robustplr}. Finally, we make use of distillation \citep{schmitt2018kickstarting}, which enables scaling to models with over 500M parameters, to the best of our knowledge the largest model trained from scratch with RL at the time of publication \citep{https://doi.org/10.48550/arxiv.2102.07920}. A high level overview of our method is shown in Figure~\ref{fig:method_overview}. 

Our main contributions are as follows:

\begin{itemize}
    \item We introduce AdA, an agent capable of human-timescale adaptation in a wide range of challenging tasks.
    \item We train AdA using meta-RL at scale in an open-ended task space with an automated curriculum.
    \item We show that adaptation is
    influenced by memory architecture, curriculum, and the size and complexity of the training task distribution.
    \item We produce scaling laws in both model size and memory, and demonstrate that AdA improves its performance with zero-shot first-person prompting. 
\end{itemize}

\section{Adaptive Agent (AdA)} 
To achieve human timescale adaptation across a vast and diverse task space, we propose a general and scalable approach for memory-based meta-RL, producing an \textit{Adaptive Agent} (AdA). We train and test AdA in XLand 2.0, an environment supporting procedural generation of diverse 3D worlds and multi-player games, with rich dynamics that necessitate adaptation. Our training method combines three key components: a curriculum to guide the agent's learning, a model-based RL algorithm to train agents with large-scale attention-based memory, and distillation to enable scaling. An overview of our approach is shown in Figure \ref{fig:method_overview}. In the following sections, we describe each component and how it contributes to efficient few-shot adaptation.

\subsection{Open-ended task space: XLand 2.0} \label{sec:methods_xverse}

In order to demonstrate fast adaptation across an open-ended task space, we extend the procedurally-generated 3D environment XLand \citep{xland}, which we refer to here as XLand 1.0. In XLand, a task consists of a game, a world, and a list of co-player policies (if any). The game consists of a goal per player, defined as a boolean function (predicate) on the environment state. An agent receives reward if and only if the goal is satisfied. Goals are defined in a synthetic language, and the agent receives an encoding. The world specifies a static floor topology, objects the player can interact with, and spawn locations for players. The agent observes the world, and any co-players therein, via a first-person pixel observation. All fundamental details of the game, world and co-player system are inherited from the original XLand; see~\cite{xland} for a full description and Appendix~\ref{app:xland_changes} for details of the new features we added.

XLand 2.0 extends XLand 1.0 with a system called \textit{production rules}. Each production rule expresses an additional environment dynamic, leading to a much richer and more diverse array of different transition functions than in XLand 1.0. The production rules system can be thought of as a domain-specific language (DSL) to express this diverse array of dynamics. Each production rule consists of:
\begin{enumerate}
    \item A \texttt{condition}, which is a predicate, for example \texttt{near(yellow sphere,black cube)},
    \item A (possibly empty) list of \texttt{spawn}s, which are objects, like \texttt{purple cube, black cube}.
\end{enumerate}

\begin{figure}[t!]
    \centering
    \includegraphics[width=\linewidth]{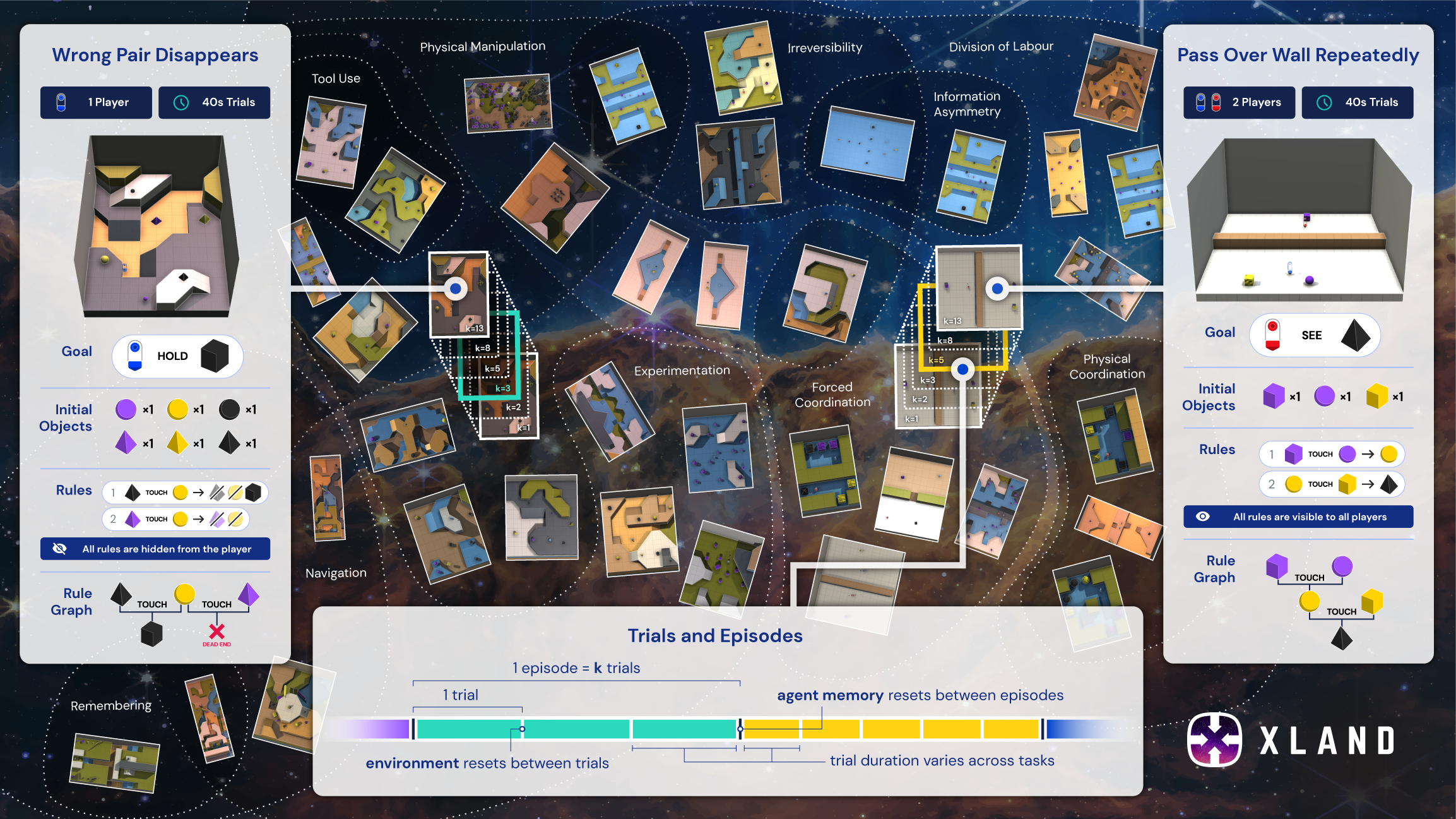}
    \caption{\textbf{XLand 2.0: a vast, smooth and diverse task space of adaptation problems.} Different tasks have different adaptation requirements, such as experimentation, tool use or division of labour. For instance, in a task requiring experimentation, a player might be required to identify which objects can usefully combine, avoiding dead-ends, and then optimise the way in which they combine objects, like a toy version of experimental chemistry. Each task can be run for one or more trials, where the environment is reset between trials, but agent memory is not. Highlighted are two example tasks, \texttt{Wrong Pair Disappears} and \texttt{Pass Over Wall Repeatedly}, showing the goal, initial objects, production rules (``rules'' in the figure) and how agents need to interact with them to solve the task. For full task descriptions see Appendix \ref{appendix:probe_tasks}.}
    \label{fig:xland-overview}
\end{figure}

When \texttt{condition} is satisfied, the objects present in \texttt{condition} get removed from the environment, and the ones in \texttt{spawns} appear.
Each game can have multiple production rules. Production rules can be observable to players, or partially or fully masked, depending on the task configuration. More precisely, there are three distinct mechanisms for hiding production rule information from the players: 

\begin{enumerate}
    \item Hiding a full production rule, where the player only gets information that a rule exists, but neither knows the condition nor what spawns.
    \item Hiding an object, where a particular object is hidden from all production rules. The hidden objects are numbered such that if multiple objects are hidden, the agent can distinguish them.
    \item Hiding a condition's predicate, where the agent gets to know the objects that need to satisfy \textit{some} predicate, but it does not know which one. The hidden predicates are also numbered.
\end{enumerate}

Instead of procedurally generating tasks on the fly, we pre-sample a large pool of tasks. For more details about the specific mechanism we use for pre-sampling tasks, see Appendix \ref{app:presample_tasks}. We visualise the XLand 2.0 task space in Figure \ref{fig:xland-overview}.

\subsection{Meta-RL} \label{sec:methods_problem} 

We use a black-box meta-RL problem setting \citep{duan2017rl, wang2016learning}. We define the task space $\mathcal{M}$ to be a set of partially-observable Markov decision processes (POMDPs). For a given task $m \in \mathcal{M}$ we define a \emph{trial} to be any sequence of transitions from an initial state $s_0$ to a terminal state $s_T$.\footnote{Note that we use a reversed naming convention to \citet{duan2017rl}. In our convention, the term ``trial'' maps well onto the related concept in the human behavioural literature \citep{barbosa2022practical}.} In XLand, tasks terminate if and only if a certain time period $T \in [10\textrm{s}, 40\textrm{s}]$ has elapsed, specified per-task. The environment ticks at $30$ frames-per-second and the agent observes every $4$\textsuperscript{th} frame, so task lengths in units of timesteps lie in the range $[75, 300]$.

An \emph{episode} consists of a sequence of $k$ trials for a given task $m$. At trial boundaries, the task is reset to an initial state. In our domain, initial states are deterministic except for the rotation of the agent, which is sampled uniformly at random. The trial and episode structure is depicted in Figure \ref{fig:xland-overview}.

In black-box meta-RL training, an agent uses experience of interacting with a wide distribution of tasks to update the parameters of its neural network, which parameterises the agent's policy distribution over actions given a state observation. If an agent possesses dynamic internal state (memory), then meta-RL training endows that memory with an implicit online learning algorithm, by leveraging the structure of repeated trials \citep{mikulik2020metatrained}. 

At test time, this online learning algorithm enables the agent to adapt its policy without any further updates to the neural network weights. Therefore, the memory of the agent is not reset at trial boundaries, but is reset at episode boundaries. To generate an episode, we sample a pair ($m$, $k$) where $k \in \{1,2, \dots 6\}$. As we will discuss later, at test time AdA is evaluated on unseen, held-out tasks across a variety of $k$ values, including on held-out $k$ not seen during training. For full details on AdA's meta-RL method, see Appendix \ref{appendix:metarl}. 
\subsection{Auto-curriculum learning} \label{sec:methods_auto_curriculum}

Given the vastness and diversity of our pre-sampled task pool, it is challenging for an agent to learn effectively with uniform sampling. Most randomly sampled tasks are likely going to be too hard (or too easy) to benefit an agent's learning progress. Instead, we use automatic approaches to select ``interesting'' tasks at the frontier of the agent's capabilities, analogous to the ``zone of proximal development'' in human cognitive development~\citep{vygotsky1978}. We propose extensions to two existing approaches, both of which strongly improve agent performance and sample efficiency (see Section~\ref{sec:results_curriculum}), and lead to an emergent curriculum, selecting tasks with increasing complexity over time.

\paragraph{No-op filtering.} We extend the dynamic task generation method proposed in \citet[Section~5.2]{xland} to our setup. When a new task is sampled from the pool, it is first evaluated to assess whether AdA can learn from it. We evaluate AdA's policy and a ``No-op'' control policy (which takes no action in the environment) for a number of episodes. The task is used for training if and only if the scores of the two policies meet a number of conditions. We expanded the list of conditions from the original no-op filtering and used normalised thresholds to account for different trial durations. See Appendix~\ref{app:autocurriculum} for further details.

\paragraph{Prioritised level replay (PLR).} We modify ``Robust PLR'' (referred to here as \emph{PLR},~\citet{jiang2021robustplr}) to fit our setup. By contrast to no-op filtering, PLR uses a \emph{fitness score} \citep{Schmidhuber1991surprise} that approximates the agent's regret for a given task. We consider several potential estimates for agent regret, ranging from TD errors as used in \citet{jiang2020prioritized}, to novel approaches using dynamics-model errors from AdA (see Appendix~\ref{app:autocurriculum} and Figure~\ref{fig:autocurriculum_abl1}). 

PLR operates by maintaining a fixed-sized archive containing tasks with the highest fitness. We only train AdA on tasks sampled from the archive, which occurs with probability $p$. With probability $1-p$, a new task is randomly sampled and evaluated, and the fitness is compared to the lowest value in the archive. If the new task has higher fitness, it is added to the archive, and the lowest fitness task is dropped. Thus, PLR can also be seen as a form of filtering, using a dynamic criteria (the lowest fitness value of the archive). It differs to no-op filtering in that tasks can be repeatedly sampled from the archive as long as they maintain high fitness. To apply PLR in our heterogeneous task space, we normalise fitness at each trial index by using rolling means and variances, and use the mean per-timestep fitness value rather than the sum, to account for varying trial duration. Finally, since we are interested in tasks at the frontier of an agent's capabilities after across-trial adaptation, we use only the fitness from the last trial. See Appendix~\ref{app:autocurriculum} for further details.

\subsection{RL agent} \label{sec:methods_rl_agent}

\paragraph{Learning algorithm.} 

We use Muesli ~\citep{hessel2021muesli} as our RL algorithm. We briefly describe the algorithm here, but refer the reader to the original publication for details. Taking a history-dependent encoding as input, in our case the output of an RNN or Transformer, AdA learns a sequence model (an LSTM) to predict the values $\hat{v_i}$, action-distributions $\hat{\pi_i}$ and rewards $\hat{r_i}$ for the next $I$ steps. Here, $i = 0, \dots, I$ denotes the prediction $i$ steps ahead. $I$ is typically small and in our case $I = 4$. For each observed step $t$, the model is unrolled for $I$ steps and updated towards respective targets: \begin{equation}
\mathcal{L}_r^t=\sum_{i=0}^I\left(\hat{r}^t_i - r_{t+i}\right)^2, \; \mathcal{L}_v^t=\sum_{i=0}^I\left(\hat{v}^t_i - G_{t+i}\right)^2, \; \mathcal{L}_{\pi}^t=\sum_{i=0}^I \textrm{KL}\left(\left. \left. \pi^{t+i}_{\textrm{CMPO}} \ \right|\right|\  \hat{\pi}^t_i \right).\label{eq:muesli_losses}
\end{equation}
Here, $r_{t+i}$ refers to the observed rewards. $G_{t+i}$ refers to value-targets which are obtained using Retrace~\citep{munos2016safe} based on Q-values obtained from one-step predictions of the model.

The action-targets $\pi^{t}_{\textrm{CMPO}}$ are obtained by re-weighting the current policy\footnote{The prior distribution is actually a mixture of the current estimate of the policy, the (outdated) policy used to produce the sample and the uniform distribution where the latter two are mixed in as regularisers.} using clipped, normalised, exponentially transformed advantages. Muesli furthermore incorporates an additional auxiliary policy-gradient loss based on these advantages to help optimise immediate predictions of action-probabilities. Finally, Muesli maintains a target network which trails the sequence model and is used for acting and to compute Retrace targets and advantages.

\begin{figure}
    \centering
    \includegraphics[width=0.7\textwidth]{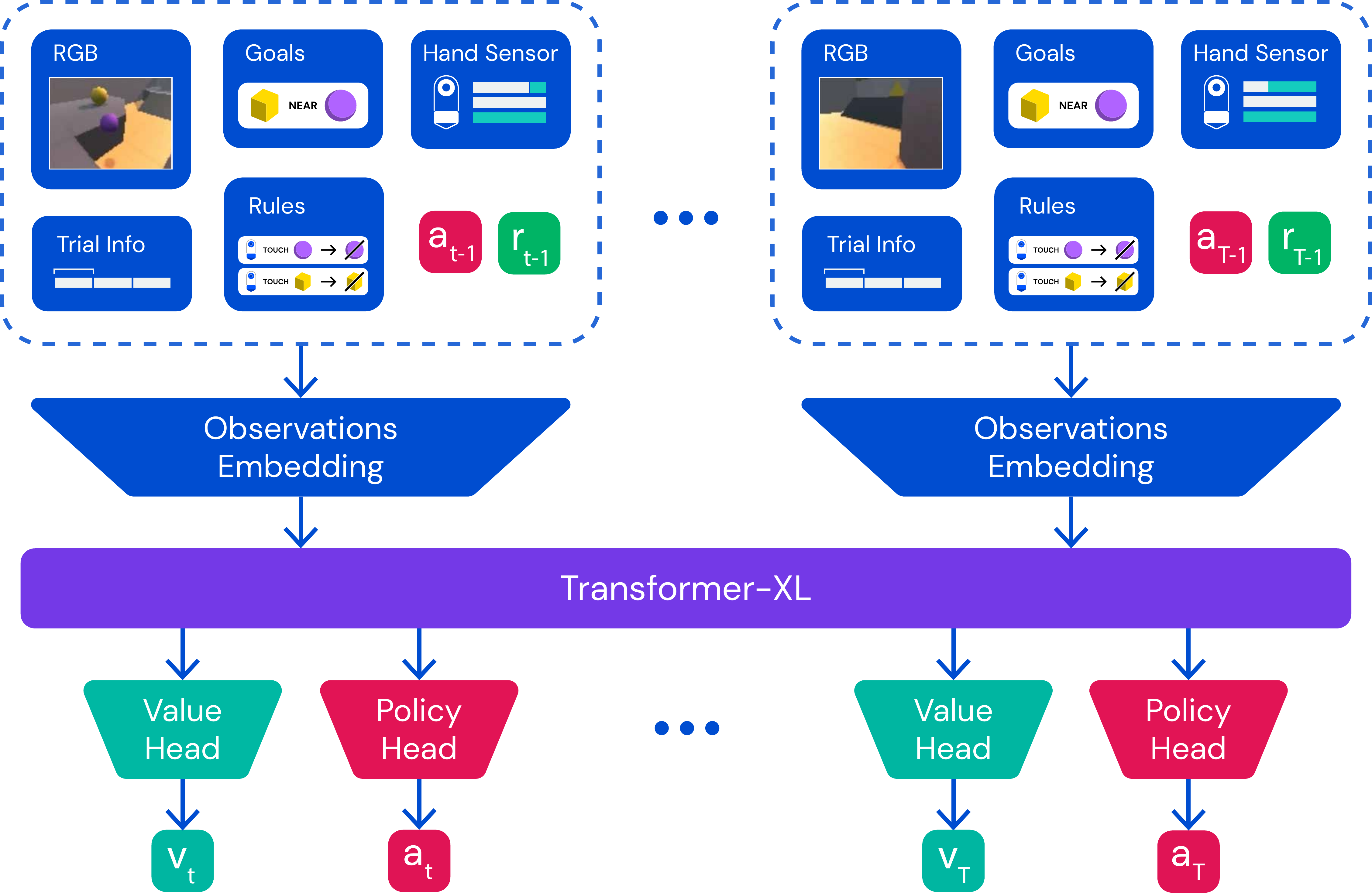}
    \caption{\textbf{Agent architecture.} For each timestep, we embed and combine the pixel observation, goal, hand, trial and time information, production rules, previous action, and previous reward into a single vector. These observations embeddings pass in sequence to the Transformer-XL, whose output embeddings feed into an MLP value head, MLP policy head, and the Muesli LSTM model step (omitted in the diagram for brevity). See Appendix \ref{app:agent-architecture} for more details about our agent architecture.}
    \label{fig:transformer_viz}
\end{figure}

\paragraph{Memory architecture.} 
\label{sec:memory-architectures}
Memory is a crucial component for adaptation as it allows the agent to store and recall information learned and experienced in the past. In order for agents to effectively adjust to the changes in task requirements, memory should allow the agent to recall information from both the very recent and the more distant past. While slow gradient-based updates are able to capture the latter, they are often not fast enough to capture the former, i.e. fast adaptation. The majority of work on memory-based meta-RL has relied on RNNs as a mechanism for fast adaptation \citep{parisotto2021meta}. In this work, we show that RNNs are not capable of adaptation in our challenging partially-observable embodied 3D task space. We experiment with two memory architectures to address this problem:

\begin{enumerate}\setlength{\itemsep}{6pt}
\item \emph{RNN with Attention} stores a number of past activations (in our case 64) in an episodic memory and attends over it, using the current hidden state as query. The output of the attention module is then concatenated with the hidden state and fed into the RNN. We increase effective memory length of the agent by storing only every 8\textsuperscript{th} activation in its episodic memory.\footnote{We arrived at these numbers as a compromise between performance and speed. Note that the resulting architecture is slower than an equivalently sized Transformer.}
\item \emph{Transformer-XL (TXL)} \citep{dai2019transformer} is a variant of the Transformer architecture \citep{vaswani2017attention} which enables the use of longer, variable-length context windows to increase the model's ability to capture long-term dependencies. To increase the stability of training Transformers with RL, we follow \citet{parisotto2020stabilizing} in performing  normalisation \textit{before} each layer, and use gating on the feedforward layers as in \citet{shazeer2020glu}.
\end{enumerate}

Both memory modules operate on a sequence of learned timestep embeddings, and produce a sequence of output embeddings that are fed into the Muesli architecture, as shown in Figure \ref{fig:transformer_viz} with a Transformer-XL module. In Section \ref{sec:results_architecture} we show that both attention-based memory modules significantly outperform a vanilla RNN in tasks that require adaptation. Transformer-XL performs the best and therefore is used as the default memory architecture in all our experiments unless stated otherwise. 

\textbf{Going beyond few shots.} We propose a simple modification to our Transformer-XL architecture to increase the effective memory length without additional computational cost. Since observations in visual RL environments tend to be highly temporally correlated, we propose sub-sampling the sequence as described for RNN with Attention, allowing the agent to attend over 4 times as many trials. To ensure that observations which fall between the sub-sampled points can still be attended to, we first encode the entire trajectory using an RNN with the intention of summarising recent history at every step. We show that the additional RNN encoding does not affect the performance of our Transformer-XL variant but enables longer range memory (see Section \ref{sec:results_few_to_many_shot}). 

\subsection{Distillation} \label{sec:methods_distillation}

For the first four billion steps of training, we use an additional distillation loss~\citep{Schmidhuber1992distillation, schmitt2018kickstarting, czarnecki2019distilling} to guide AdA's learning with the policy of a pre-trained teacher, in a process known as kickstarting; iterating this process leads to a generational training regime~\citep{xland,wang2102alchemy}. The teacher is pre-trained from scratch via RL, using an identical training procedure and hyperparameters as AdA, apart from the lack of initial distillation and a smaller model size (23M Transformer parameters for the teacher and 265M for multi-agent AdA). Unlike aforementioned prior work, we do not employ shaping rewards or Population Based Training (PBT,~\citet{jaderberg2017population}) in earlier generations. During distillation, AdA acts according to its own policy and the teacher provides target logits given the trajectories observed by AdA. Distillation allows us to amortise an otherwise costly initial training period, and it allows the agent to overcome harmful representations acquired in the initial phases of training; see Section \ref{sec:results_kickstarting}. 

To integrate the distillation loss with Muesli, we unroll the model from every transition observed by the student. We minimise the KL-divergence between all of the action-probabilities predicted by the model and the action-probabilities predicted by the teacher's policy at the corresponding timestep. Analogously to Muesli's policy-loss $\mathcal{L}_\pi$ defined in \eqref{eq:muesli_losses}, we define
\begin{equation}
\mathcal{L}_{\textrm{dist}}=\sum_{i=0}^I \textrm{KL}\left(\left. \left.\tilde{\pi}^{t+i}_0 \ \right|\right|\ \hat{\pi}^t_i \right)\, ,
\end{equation}
where $\tilde{\pi}$ corresponds to the predicted action-logits provided by the teacher given the same observed history. Furthermore, we found it useful to add additional $L^2$ regularisation during distillation.

\begin{figure}[htb]
    \centering
    \includegraphics[width=0.8\linewidth]{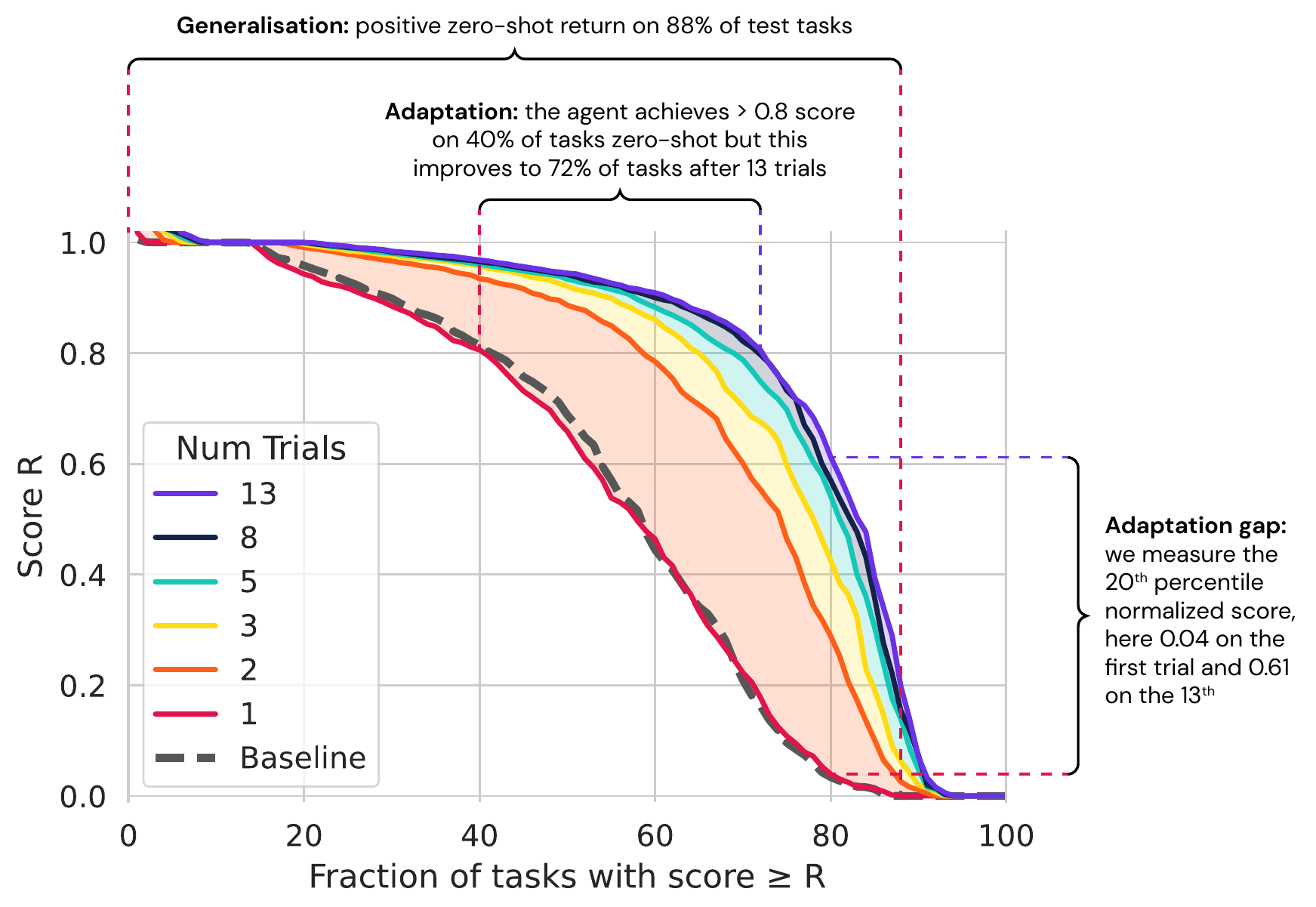}
    \caption{\textbf{Zero-shot generalisation and few-shot adaptation.} We report the distribution of normalised task scores over the single-agent test set when evaluated with various numbers of trials. On the $y$-axis is the total last-trial reward relative to that of an agent fine-tuned on the test tasks (approximating ``infinite trials'' performance). Curves moving further towards the top right corner indicate better performance. When given more trials, the agent achieves higher scores in the last trial, showing test-time adaptation across most of the task distribution (shaded regions). The dashed line indicates the zero-shot performance of an agent trained in a regime where every episode consists of only a single trial.}
    \label{fig:single_agent_percentiles}
\end{figure}

\section{Experiments and Results}

We evaluate our agents in two distinct regimes: on a set of $1000$ \textit{test tasks} sampled from the same distribution as the training tasks, and on a set of $30$ single-agent and $28$ multi-agent \emph{hand-authored probe tasks}. A rejection sampling procedure guarantees that the procedural test tasks and probe tasks are outside the training set. The probe tasks represent situations that are particularly intuitive to humans, and deliberately cover a wide range of qualitatively different adaptation behaviours. Example probe tasks are depicted in Figures \labelcref{fig:wrong-pair-disappears-explained,fig:pyramid-in-a-haystack-explained,fig:no-touchy-explained} in the Appendix, and a full description of every probe task is available in Appendix \ref{appendix:human_scale_adaptation}. 

The total achievable reward on each task varies, so whenever we present aggregated results on the test or hand-authored task set, we normalise the total per-trial reward for each task against the reward obtained by fine-tuning AdA on the respective task set. We refer to this normalised reward as a \textit{score}. We stipulate that an adaptive agent must have two capabilities: zero-shot generalisation and few-shot adaptation. Zero-shot \textit{generalisation} is assessed by the score in the case of only being given $1$ trial of interaction with a held-out task. Few-shot \textit{adaptation} is assessed by the improvement in score as the agent is given progressively more trials ($k$) of interaction with the task. More precisely, for each $k$ we report the score in the last trial, showing whether or not an agent is able to make use of additional experience on-the-fly to perform better, i.e. measuring adaptation. 

We aggregate scores across a task set using (one or more) percentiles. When presenting individual probe tasks we report unnormalised total last trial rewards per task for agents and for human players where applicable. For full details of our evaluation methodology see Appendix~\ref{app:evaluation}. 

The space of training configurations for AdA is large, comprising model size, auto-curriculum, memory architecture, memory length, number of tasks in the XLand task pool, single vs multi-agent tasks, distillation teacher, and number of training steps. We use a consistent training configuration within each experimental comparison, but different configurations across different experimental comparisons. We therefore caution the reader against directly comparing results between different sections. For convenience, all experimental configurations are tabulated in Appendix \ref{app:training-details}.

\subsection{AdA shows human-timescale adaptation} \label{sec:results_human_scale_adaptation}

\begin{table}[h!]
  \begin{center}
    \caption{Experimental setup for agent experiments in Section \ref{sec:results_human_scale_adaptation}.}
    \begin{tabular}{c|c|c|c|c|c|c}
    \label{tab:human-timescale-settings}
      \# players & Model parameters & Memory & Task pool & Curriculum & Teacher & Steps\\
      \hline
      1 & 169M TXL / 353M total & 1800 & 25B & PLR \ref{tab:plr_hyperparameters} & \ref{tab:single-agent-teacher} & 100B \\
      2 & 265M TXL / 533M total & 1800 & see App.~\ref{app:multi_agent_training} & PLR \ref{tab:plr_hyperparameters} & \ref{tab:multi-agent-teacher} & 70B \\
    \end{tabular}
  \end{center}
\end{table}

\paragraph{Single-agent.}

In Figure~\ref{fig:single_agent_percentiles} we show the performance of AdA when trained in the single-agent setting described in Table \ref{tab:human-timescale-settings}. Examine first AdA's zero-shot performance ($k=1$, red line). This matches the performance of a baseline agent, trained only in a regime where each episode consists of a single trial. In other words, AdA does not suffer any degradation in zero-shot performance, despite being trained on a distribution over number of trials $k \in \{1, 2, \dots 6\}$. Now turn your attention to AdA's few-shot performance ($k \in \{2, 3, 5, 8, 13\}$, orange to purple lines). Given more trials, AdA improves its performance on over $80\%$ of the task set, clearly adapting at test time. The improvements are particularly strong when comparing zero-shot performance to the two trial setting, but AdA keeps on improving when given more trials. 

\begin{figure}[t]
    \centering
    \begin{subfigure}[b]{0.47\textwidth}
        \centering
        \includegraphics[width=\linewidth]{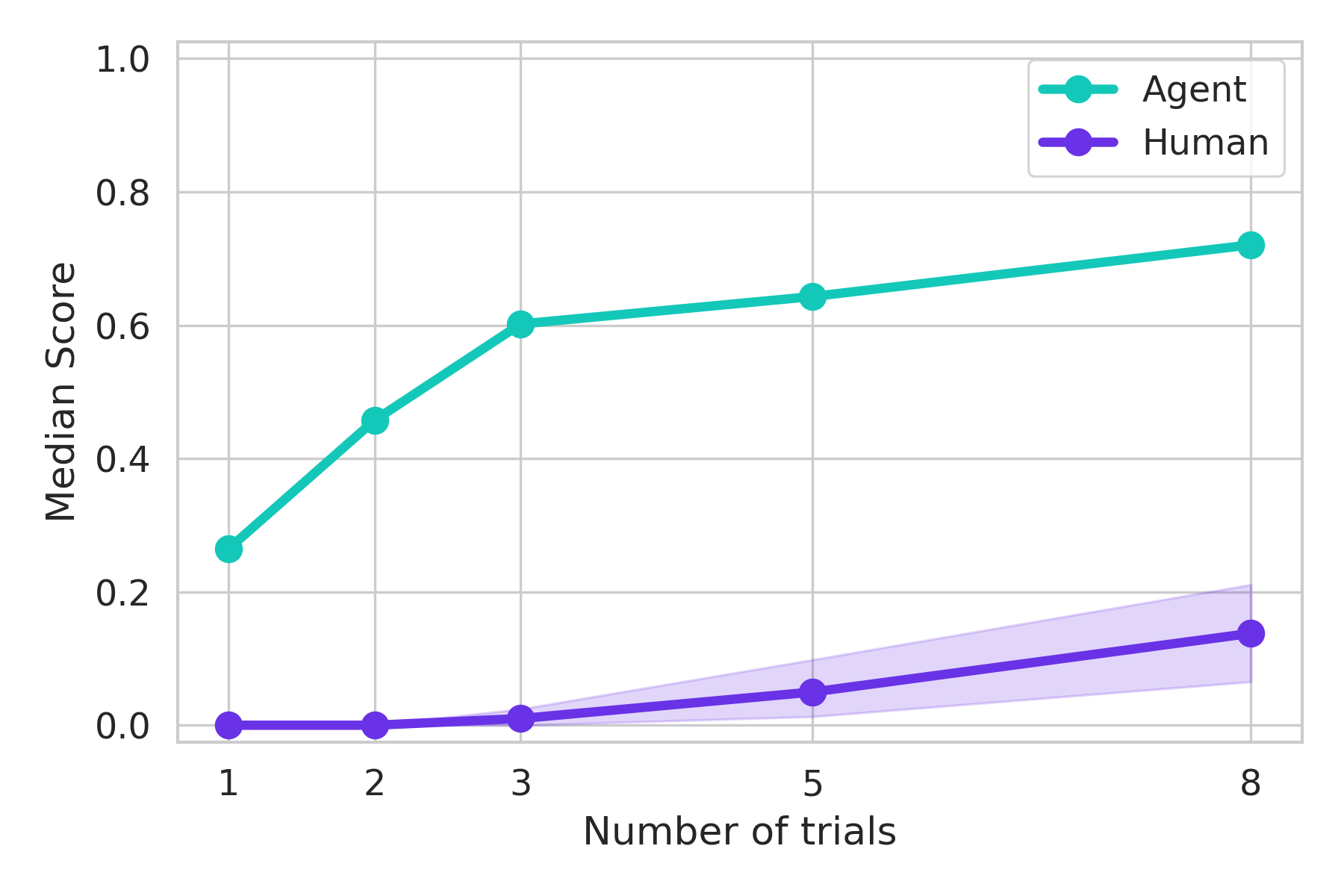}
        \vskip -3pt \caption{}
        \label{fig:human_scale_adaptation}
    \end{subfigure}
    \begin{subfigure}[b]{0.47\textwidth}
        \centering
        \includegraphics[width=\linewidth]{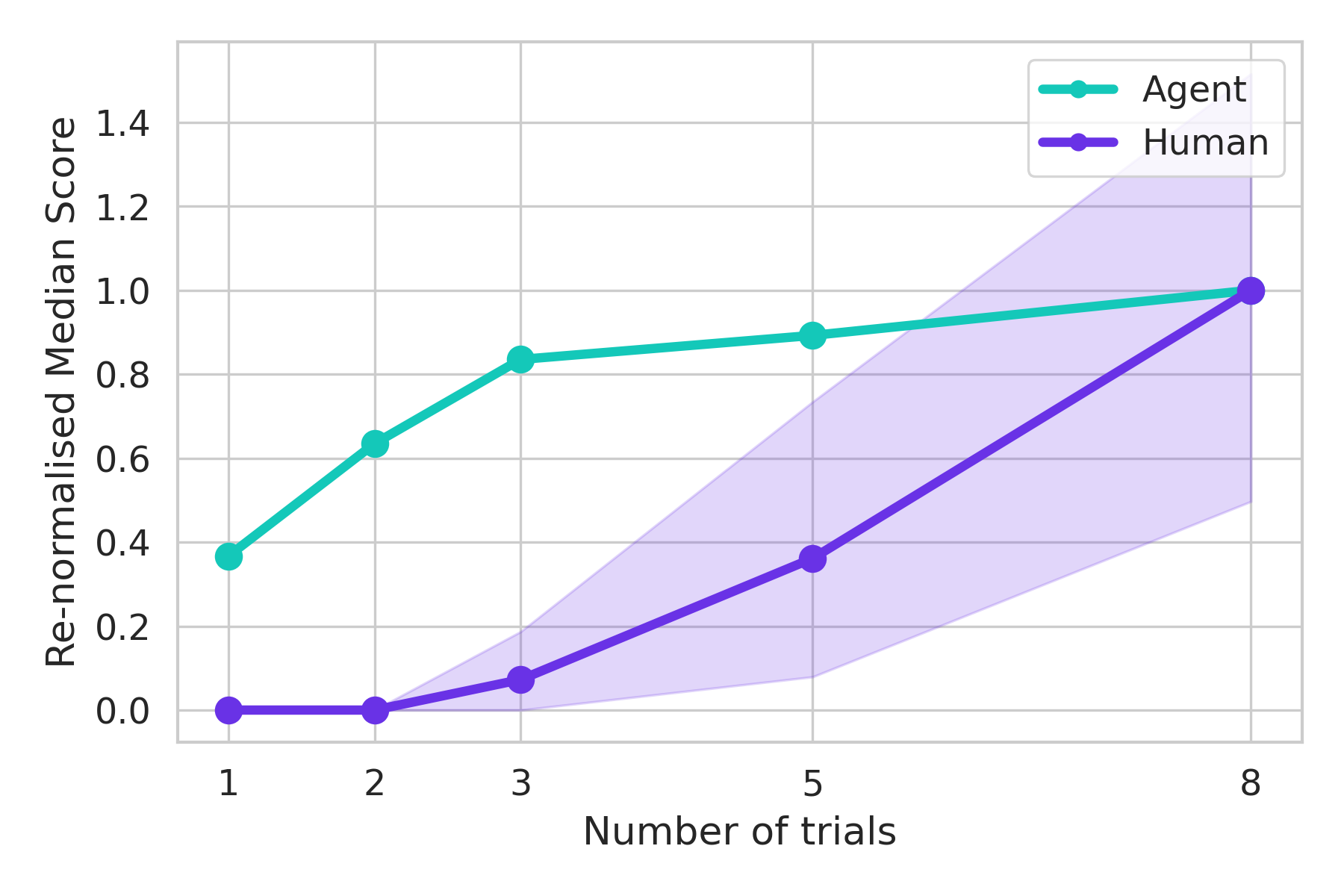}
        \vskip -3pt \caption{}
        \label{fig:human_scale_adaptation_renorm}
    \end{subfigure}

    \caption{\textbf{Human-timescale adaptation.} We report median normalised last-trial score across 30 hand-authored tasks as a function of number of trials for AdA and human players. Both AdA and the human players improve their performance with increasing number of trials, indicating that AdA is capable of human-timescale adaptation. \textbf{(a)} shows the results using our standard per-task normalisation scheme. \textbf{(b)} re-normalises the results by the maximum score per player-type to account for systematic differences between the agent and human players. In particular, human players reported lag while playing which may have resulted in lower scores.}
\end{figure}

\begin{figure}[htb]
    \centering
    \includegraphics[width=1.0\linewidth]{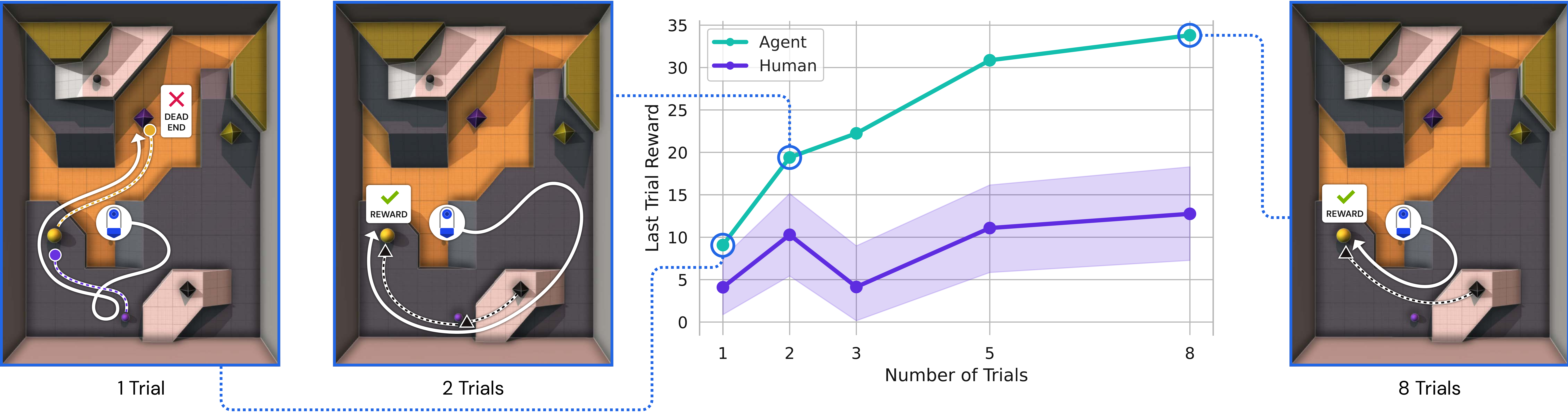}
    \caption{\textbf{Experimentation, success and refinement.} We report average performance and representative behaviour of AdA on the probe task \texttt{Wrong Pair Disappears} when evaluated with various numbers of trials. AdA's performance increases when given more trials, showing test-time adaptation. The top-down view images show representative last-trial trajectories when given different numbers of total trials. A corresponding \href{https://youtu.be/FCDLu4iTBGE}{video} for the case $k = 3$ shows the behaviour across all trials within one episode.}
    \label{fig:probe_task_performance}
\end{figure}

We compare the performance of AdA to that of a set of human players on $30$ held-out hand-authored probe tasks, seeking to assess whether AdA adapts on the same timescale as humans. Figure \ref{fig:human_scale_adaptation} shows the median scores for AdA and for human players as a function of number of trials. Both AdA and human players were able to improve their score as they experienced more trials of the tasks, indicating that AdA exhibits human-timescale adaptation on this set of probe tasks. We provide more details of the scores obtained on each task in Figure \ref{fig:human_scale_adaptation_all_tasks}. This reveals a small set of tasks which humans can solve but AdA can't, such as the \texttt{Spacer Tool} task: in this task one object must be used as a tool to move another, a situation which is extremely rare in XLand. There are also tasks like \texttt{Small Workstation, All Rules Visible} which can be solved by AdA but not by humans, likely due to complex control requirements. The majority of tasks, however, show adaptation from both humans and AdA, with the slopes of AdA's score being as steep as, if not steeper than, those of the human players, especially for lower numbers of trials. For full details of our human experiment design, see Appendix \ref{app:human-data-collection}.

Figure~\ref{fig:probe_task_performance} analyses the behaviour of AdA in more detail on a specific held-out task. The increase in score with a larger number of trials indicates that the task is solved more consistently and more quickly when given a larger number of trials. Examining the trajectories for different numbers of trials, we can explain this effect in terms of the behaviour of AdA. When given $1$ or $2$ trials AdA's behaviour shows structured hypothesis-driven exploration: trying out different combinations of objects and coming across the solution or a dead end. Once the solution is found, AdA refines its strategy on subsequent trials, gathering the correct objects with more efficiency and combining them in the right way. Thus AdA is able to generate a higher last-trial score when provided with more trials for refinement. When given $8$ trials, the last-trial performance is close to that of the fine-tuned agent. We observe this pattern of behaviour consistently across many of our held-out probe tasks; see videos on our \href{http://sites.google.com/view/adaptive-agent/}{microsite}.

\paragraph{Multi-agent.} We train a separate agent on a mixture of fully-cooperative multi-agent and single-agent tasks to explore adaptation in the multi-agent setting. In fully-cooperative multi-agent tasks, both players have the same goal. Such tasks typically have multiple Nash equilibria \citep{DBLP:journals/corr/abs-2012-08630}. When faced with a new problem, agents must adapt on-the-fly to agree on a single equilibrium of maximal mutual benefit \citep{stone2010ad, hu2020other, https://doi.org/10.48550/arxiv.2209.14344}. This gives rise to a variety of interesting strategic novelties that are absent in the purely single-agent setting, including emergent division-of-labour and physical coordination. Both of these behaviours have received extensive study in the multi-agent RL literature (e.g. \cite{wang2020roma, https://doi.org/10.48550/arxiv.2006.06051, strouse2021collaborating, gronauer2022multi}); here for the first time to our knowledge, we demonstrate that these behaviours can emerge at test time in few-shot on held-out tasks. Co-players for our training tasks are generated using fictitious self-play \citep{heinrich2015fictitious} and then curated using PLR, as in \citet{samvelyan2022maestro}. For more details, see Table \ref{tab:human-timescale-settings} and Appendix~\ref{app:multi_agent_training}.

\begin{figure}[t!]
    \centering
    \begin{subfigure}[b]{0.48\textwidth}
        \centering
        \includegraphics[width=\linewidth]{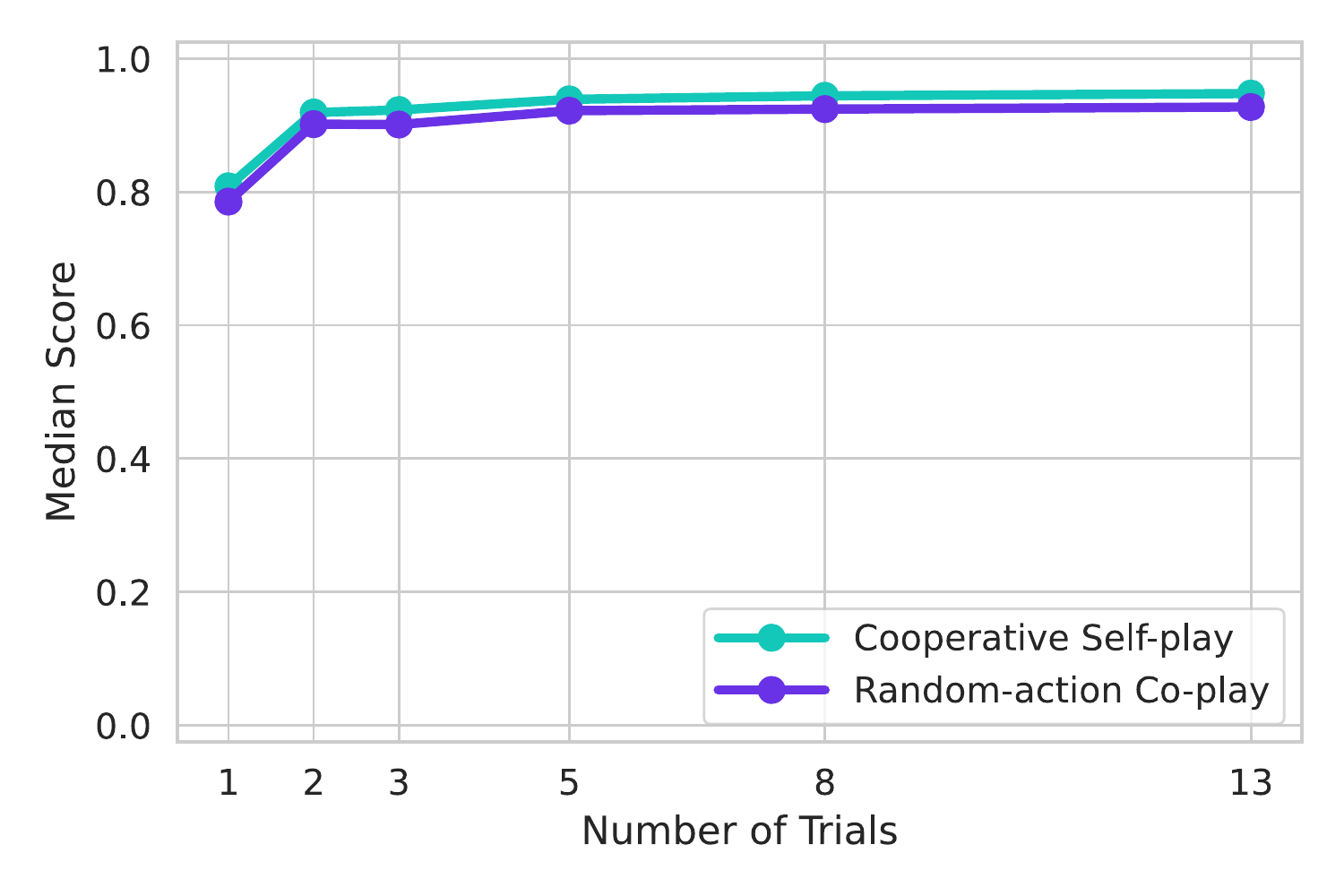}
        \vskip -3pt \caption{}
    \end{subfigure}
    \begin{subfigure}[b]{0.48\textwidth}
        \centering
        \includegraphics[width=\linewidth]{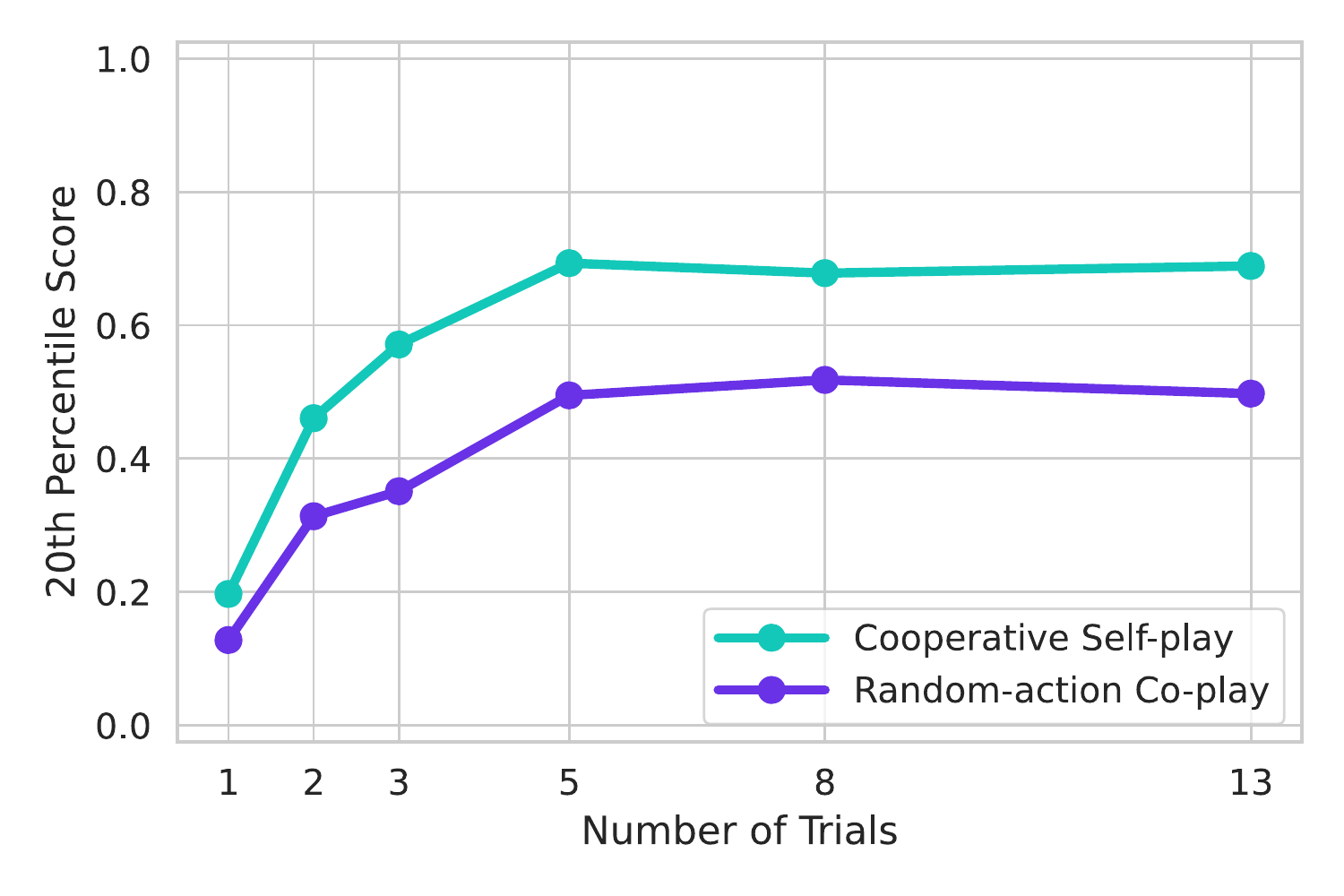}
        \vskip -3pt \caption{}
    \end{subfigure}
    \caption{
        \textbf{Two heads are better than one.} Cooperative self-play outperforms single-agent performance on the test set of two-player cooperative held-out tasks. For this evaluation we restrict ourselves to tasks whose goals and production rules do not refer to players and which are solvable by a single player (216/1000 test tasks). To produce the purple curve, we evaluate AdA twice per task when playing with a random-action policy co-player, once playing as the first and once as the second player, and take the maximum score over both evaluations before cross-task aggregation. This accounts for possible advantages playing as one player might have over playing as the other in a task.
        \textbf{(a)} Median score.
        \textbf{(b)} 20\textsuperscript{th} percentile score.
    }
\label{fig:self_play_vs_single_agent}
\end{figure}

\begin{figure}[t!]
    \centering
    \includegraphics[width=1.0\linewidth]{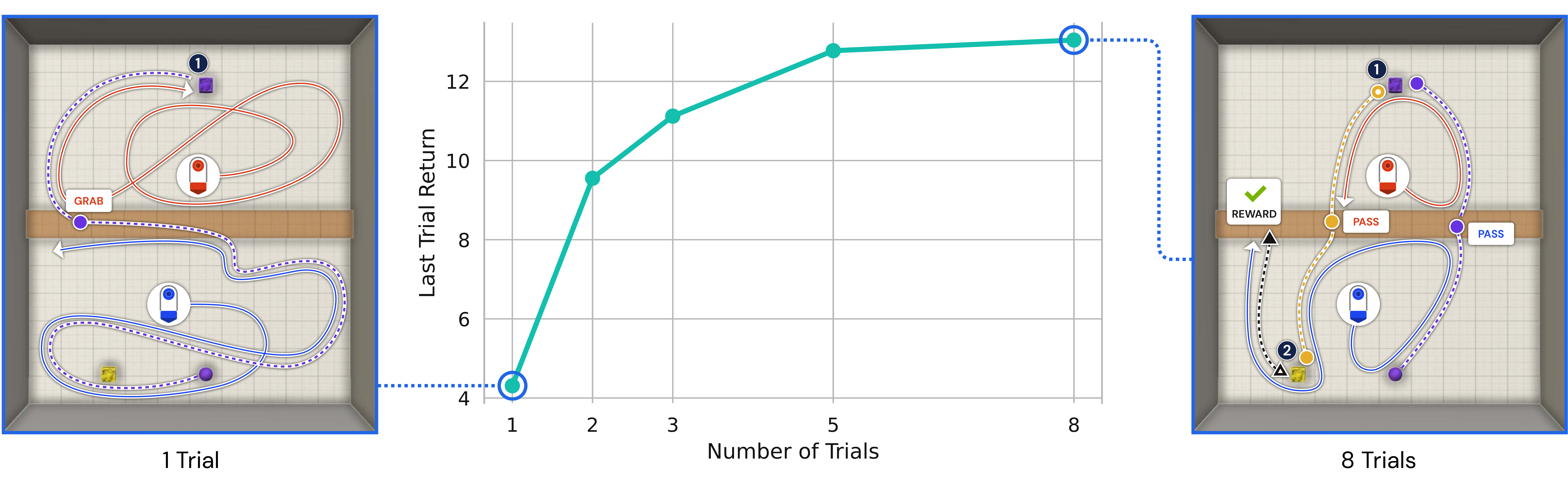}
    \caption{\textbf{Multi-agent coordination.} We report average performance and representative behaviour of AdA on the probe task \texttt{Pass Over the Wall Repeatedly} when evaluated in self-play with various numbers of trials. AdA's performance increases when given more trials, showing test-time adaptation. The top-down view images show representative last-trial trajectories when given different numbers of total trials. A corresponding \href{https://youtu.be/Rnz2MFgeicc}{video} for the case $k = 5$ shows the behaviour across all trials within one episode.}
    \label{fig:multi_agent_probe_forced_coop}
\end{figure}

Analogously with the single-agent setting, we find strong evidence of adaptation across almost $90 \%$ of the space of held-out test tasks (Figure \ref{fig:multi_agent_percentiles}). Futhermore, we evaluate the resulting agent on a held-out test set of cooperative multi-agent tasks in two ways: in self-play and in co-play with a random-action policy. As shown in Figure~\ref{fig:self_play_vs_single_agent}, self-play outperforms co-play with a random-action policy by a large margin both in a zero-shot and in a few-shot setting. This indicates that the agents are dividing the labour required to solve the tasks, thereby solving the task more quickly (or at all) and improving their shared performance.

Examples of emergent social behaviour in self-play are shown in Figures~\ref{fig:multi_agent_probe_forced_coop} and \ref{fig:multi_agent_probe_emergent_coop}. When given only a few trials, the agents explore the space of possible solutions, sometimes operating independently and sometimes together. Given more trials, once the agents find a solution, they optimise their paths by coordinating physically and dividing labour to solve the task efficiently. This behaviour emerges from adaptation at test time and was not explicitly incentivised during training, other than through the high-level fully cooperative reward function. Videos of such behavior in a variety of tasks are available on our \href{http://sites.google.com/view/adaptive-agent/}{microsite}.

\begin{figure}[t]
    \centering
    \begin{subfigure}[b]{0.47\textwidth}
        \centering
        \includegraphics[width=\linewidth]{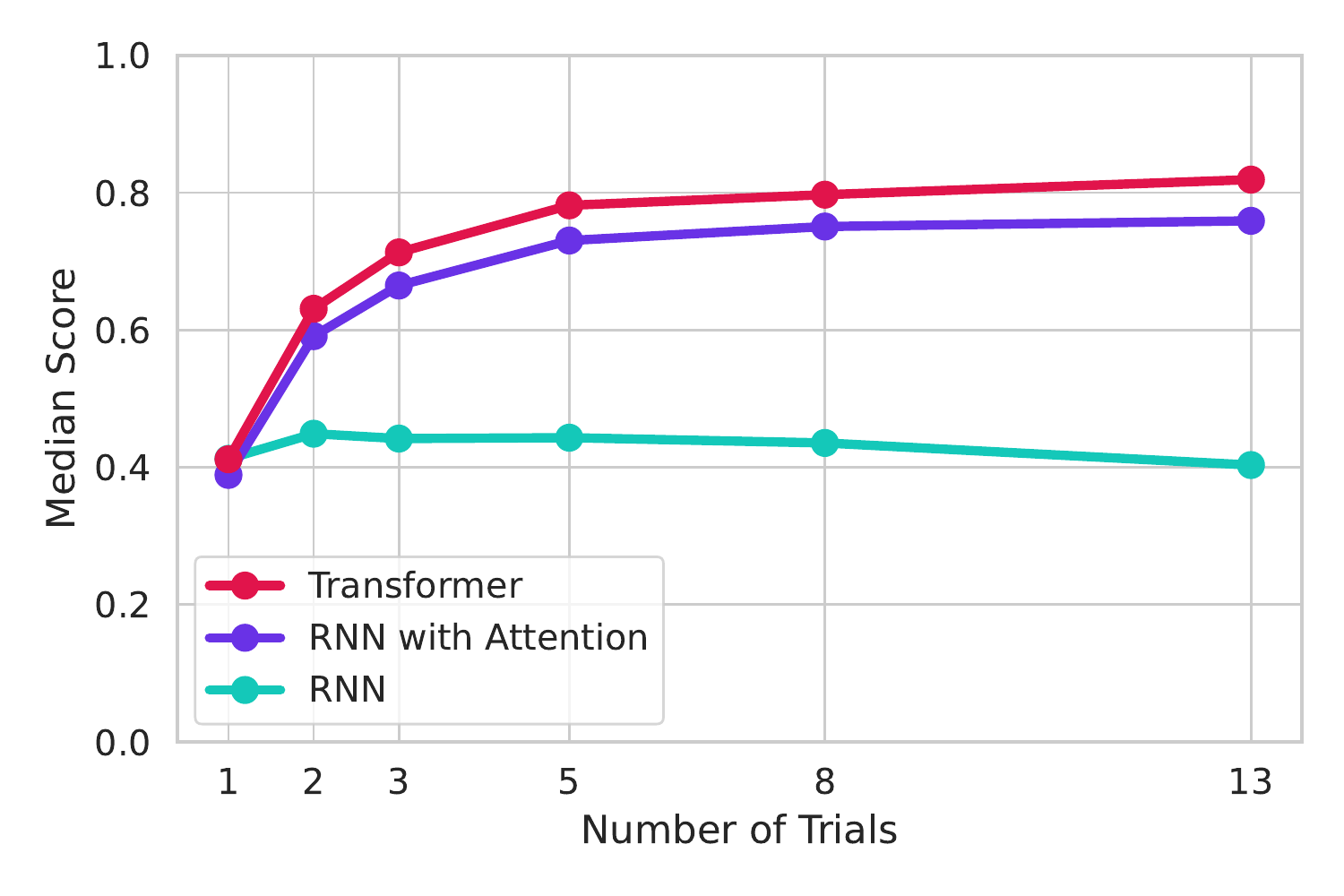}
        \vskip -3pt  \caption{Impact of architecture}
        \label{fig:architecture_comparison}
    \end{subfigure}
    \begin{subfigure}[b]{0.47\textwidth}
         \centering
         \includegraphics[width=\linewidth]{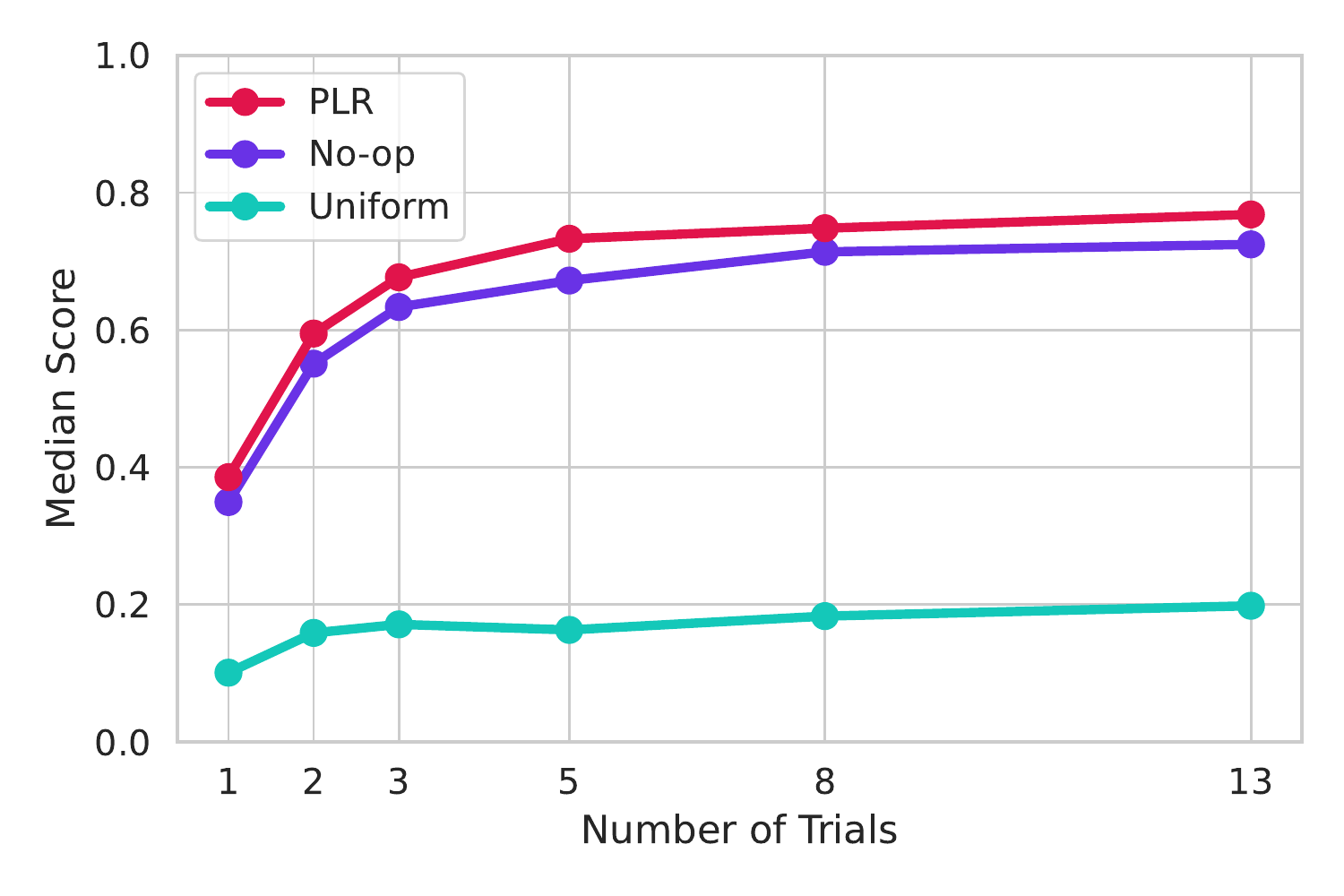}
         \vskip -3pt \caption{Impact of curriculum}
         \label{fig:curricula_performance}
     \end{subfigure}
    
    \caption{
         \textbf{(a)} Adaptation over increasing numbers of trials for different choices of architectures. Incorporating attention modules is essential to achieve adaptation, with Transformer-XL architectures performing best. \textbf{(b)} Adaptation over increasing numbers of trials for different choices of curricula. No-op filtering and PLR greatly improve both zero-shot generalisation and few-shot adaptation over the uniform sampling baseline.
    }
\end{figure}

\subsection{Architecture influences performance} \label{sec:results_architecture}

We now dive deeper into understanding which components of our method are critical, via a series of ablation studies. In these studies we use a single initialisation seed, because we see low variance across seeds when training AdA (see Appendix \ref{app:seed-variation}). All ablations are in the single-agent setting, unless stated otherwise.

First, we empirically contrast different choices of architectures: Transformer-XL, RNN, and RNN with Attention. To implement the RNN, we use a GRU \citep{cho2014properties}. To facilitate comparison, we match the total network size for all architectures. Table \ref{tab:arch_exp} shows details on the experimental setup. Figure \ref{fig:architecture_comparison} shows that while the Transformer-XL is the best performing architecture in this comparison, incorporating a multi-head attention module into an RNN recovers most of the performance of the Transformer, highlighting the effectiveness of attention modules.

\subsection{Auto-curriculum learning improves performance} \label{sec:results_curriculum}

To establish the importance of automated curriculum learning, we compare adaptation when training with the curricula methods outlined in Section~\ref{sec:methods_auto_curriculum}: no-op filtering and PLR. Figure~\ref{fig:curricula_performance} shows the median last-trial score of agents trained with different curricula. Both no-op filtering and PLR curricula strongly outperform a baseline trained with uniformly sampled tasks. Moreover, PLR outperforms No-op filtering, particularly at a higher number of trials, indicating that a regret-based curriculum is especially helpful for learning longer-term adaptation. In Appendix \ref{app:autocurriculum} we detail training configuration, and also compare the sample efficiency of our methods, where we see that both auto-curriculum approaches are more sample-efficient than uniform sampling, in terms of both learning steps and FLOPs.

\begin{figure}[htb!]
    \centering
    \includegraphics[width=0.99\linewidth]{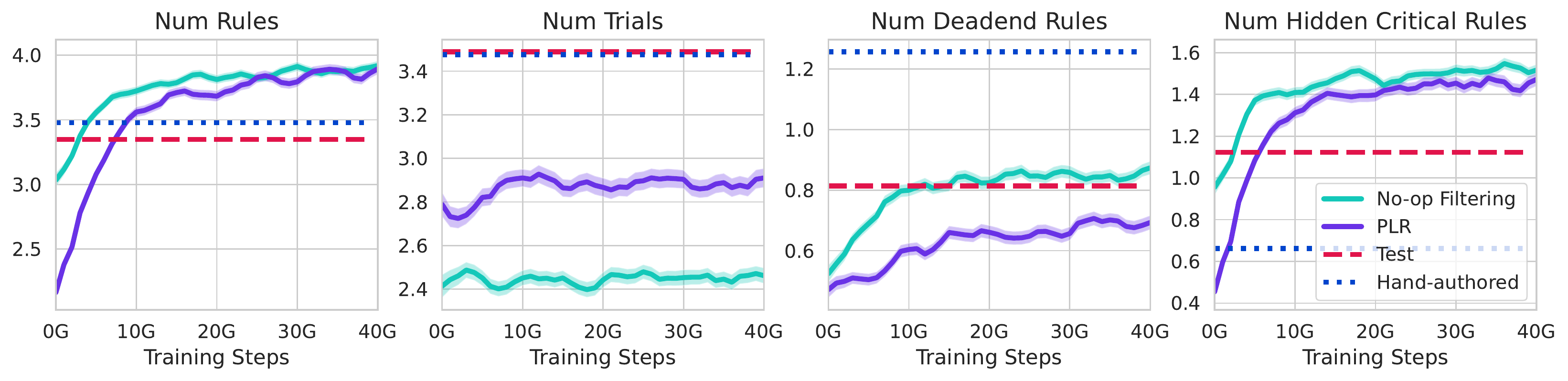}
    \caption{
        Emergent curricula for no-op filtering and PLR. Plots show a selection of task metrics for the dynamic training set, averaged over all tasks in the set, with standard error shaded. In all plots, a higher metric value corresponds to greater task difficulty. For example, tasks with a higher number of rules require more trial-and-error to find the correct rules to trigger. Horizontal lines show the same metric values averaged over the test (dashed) and hand-authored (dotted) evaluation task sets.}
    \label{fig:emergent_curricula}
\end{figure}

In Figure~\ref{fig:emergent_curricula} we show the evolution of task complexity for both methods. In both cases, simpler tasks are initially prioritised, with a clear curriculum emerging. Neither method explicitly optimises to increase these metrics, yet the task complexity increases as a result of the agent's improving capabilities. See Figure~\ref{fig:emergent_curricula_extended} for additional metrics of task complexity.

\subsection{Scaling the agent increases performance}
\label{sec:scaling_results}

\begin{figure}[t!]
    \centering
    \begin{subfigure}[b]{0.49\textwidth}
        \centering
        \includegraphics[width=\linewidth]{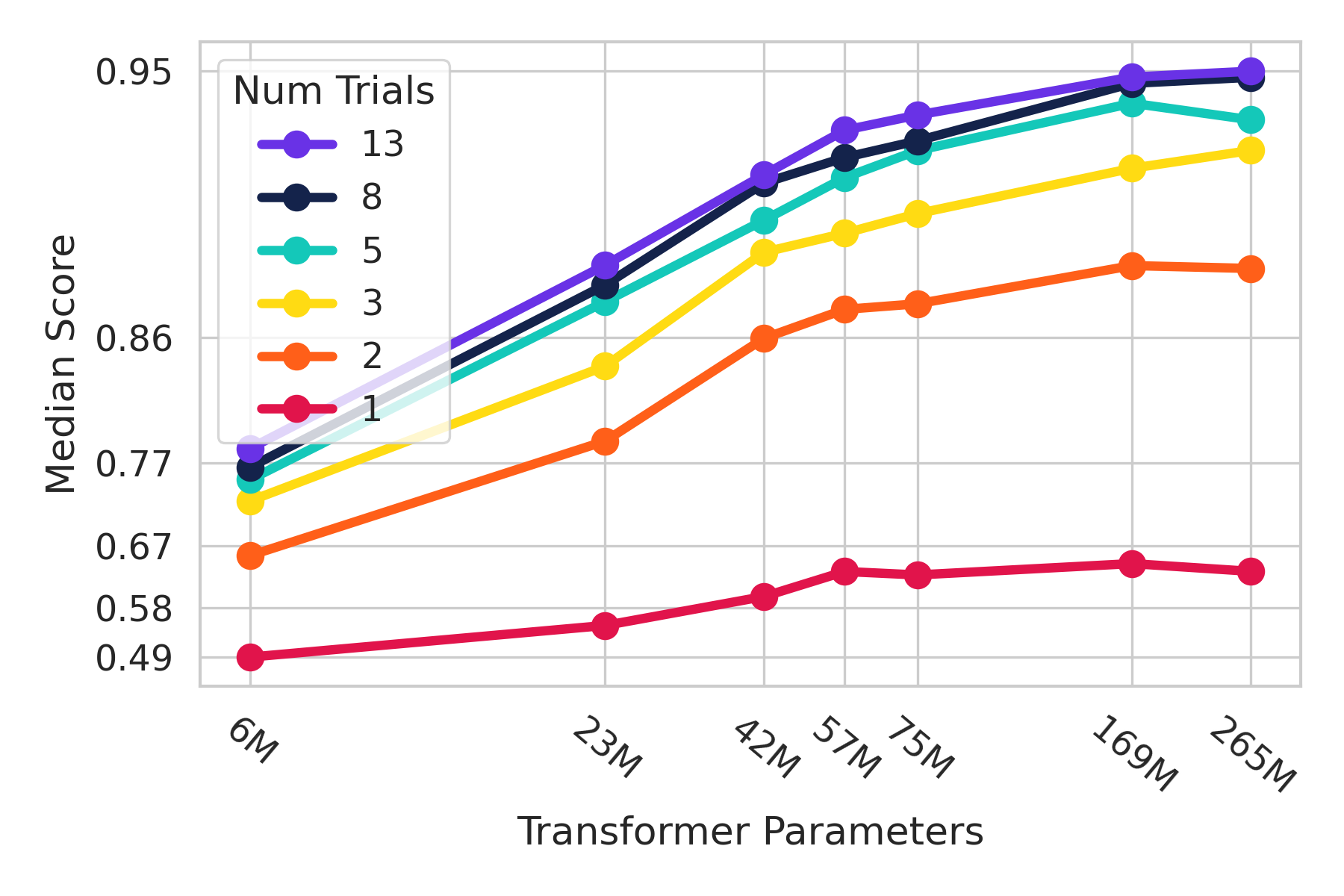}
        \vskip -3pt \caption{}
        \label{figure:scaling_network_50th}
    \end{subfigure}
    \begin{subfigure}[b]{0.49\textwidth}
        \centering
        \includegraphics[width=\linewidth]{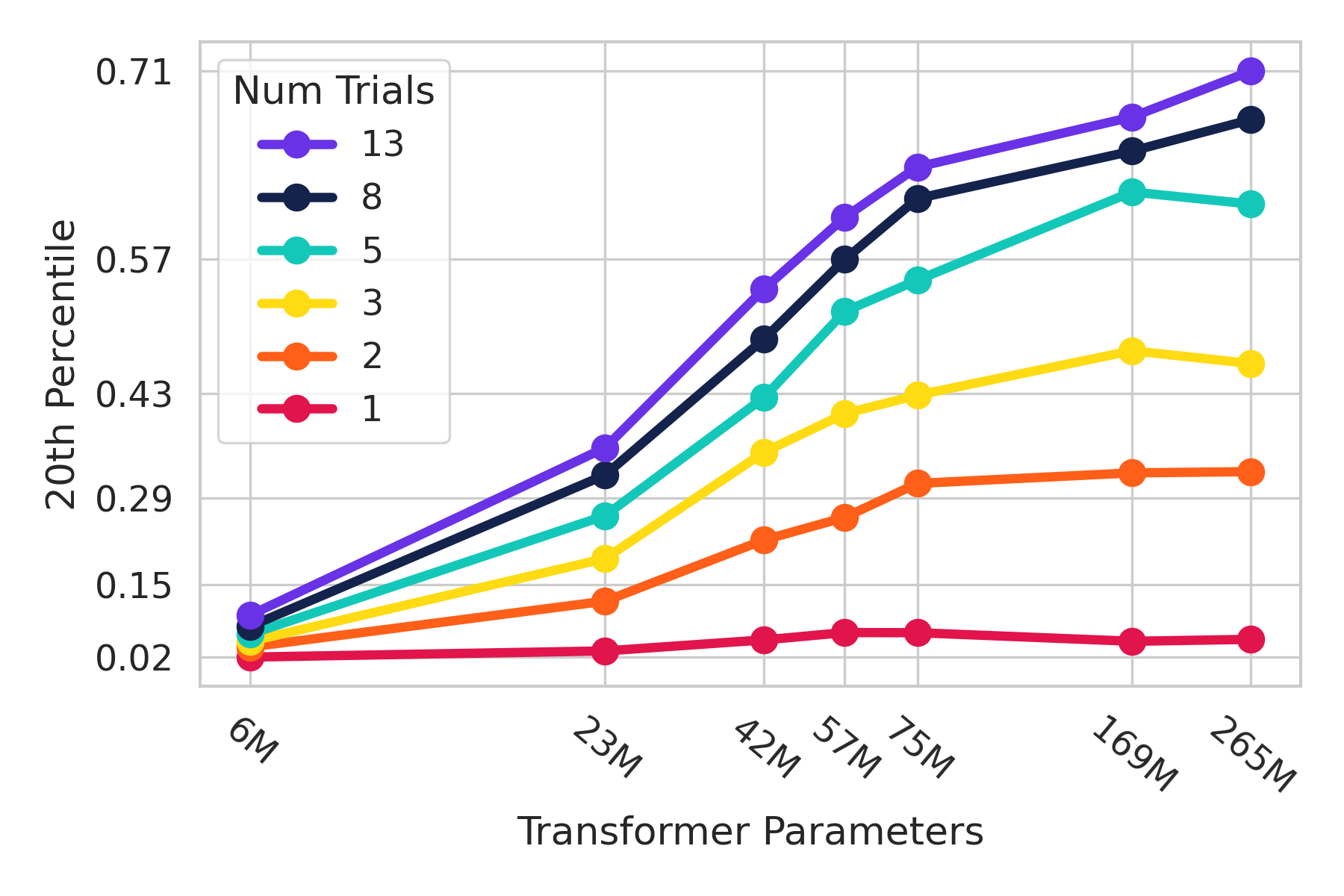}
        \vskip -3pt \caption{}
        \label{figure:scaling_network_20th}
    \end{subfigure}
    \caption{Scaling Transformer parameters increases 
    both median \textbf{(a)} and 20th percentile \textbf{(b)} test score. Both axes are log-scaled, according to the functions $\log(x)$ and $-\log(1-y)$, respectively, and the relationship between model size and performance appears roughly linear on this scale. The slope is steeper when evaluating higher numbers of trials, showing that scaling the model is particularly effective at encouraging stronger adaptation, as opposed to stronger zero-shot generalisation.}
\label{figure:scaling_network}
\end{figure}

Methods that scale well are critical for continued progress in machine learning, and understanding how methods scale is important for deciding where to spend time and compute in the future. Scaling laws have been determined for many foundation models (see Section \ref{sec:related_work}), where performance is related to model size and other factors as a power law, which can be seen as a linear relationship on a log-log plot. Inspired by such analyses, we investigate how adaptation scales with Transformer model size and memory length.

\paragraph{Scaling network size.}

We show how performance scales with the size of AdA's Transformer model, experimenting with the model sizes shown in Table \ref{tab:model_scaling}. When investigating scaling laws for model size, we follow \citet{kaplan2020scaling} in measuring only Transformer (i.e. non-embedding) parameters, which range across $3$ orders of magnitude, from $6$M to $265$M Transformer parameters (i.e. from $41$M to $533$M total parameters). A complete list of hyperparameters is shown in Table \ref{tab:model_hparams}.

Figure \ref{figure:scaling_network} shows that larger networks increase performance, especially when given more test-time trials to adapt. Though larger models seem to help in the median test-set score (Figure \ref{figure:scaling_network_50th}), model scale particularly has impact on the lower percentiles of the test set (Figure \ref{figure:scaling_network_20th}). This indicates that larger models allow the agent to generalise its adaptation to a broader range of tasks. 
The roughly linear relationship between model size and performance on the log-log plot is indicative of a power law scaling relationship, albeit only shown across two to three orders of magnitude. That the curves are not exactly linear may be due to several factors: that we haven't trained to convergence (though performance increases had slowed for all models), and that we use a 23M parameter distillation teacher across experiments for all model sizes. 

\begin{figure}[t!]
    \centering
    \begin{subfigure}[b]{0.49\textwidth}
        \centering
        \includegraphics[width=\linewidth]{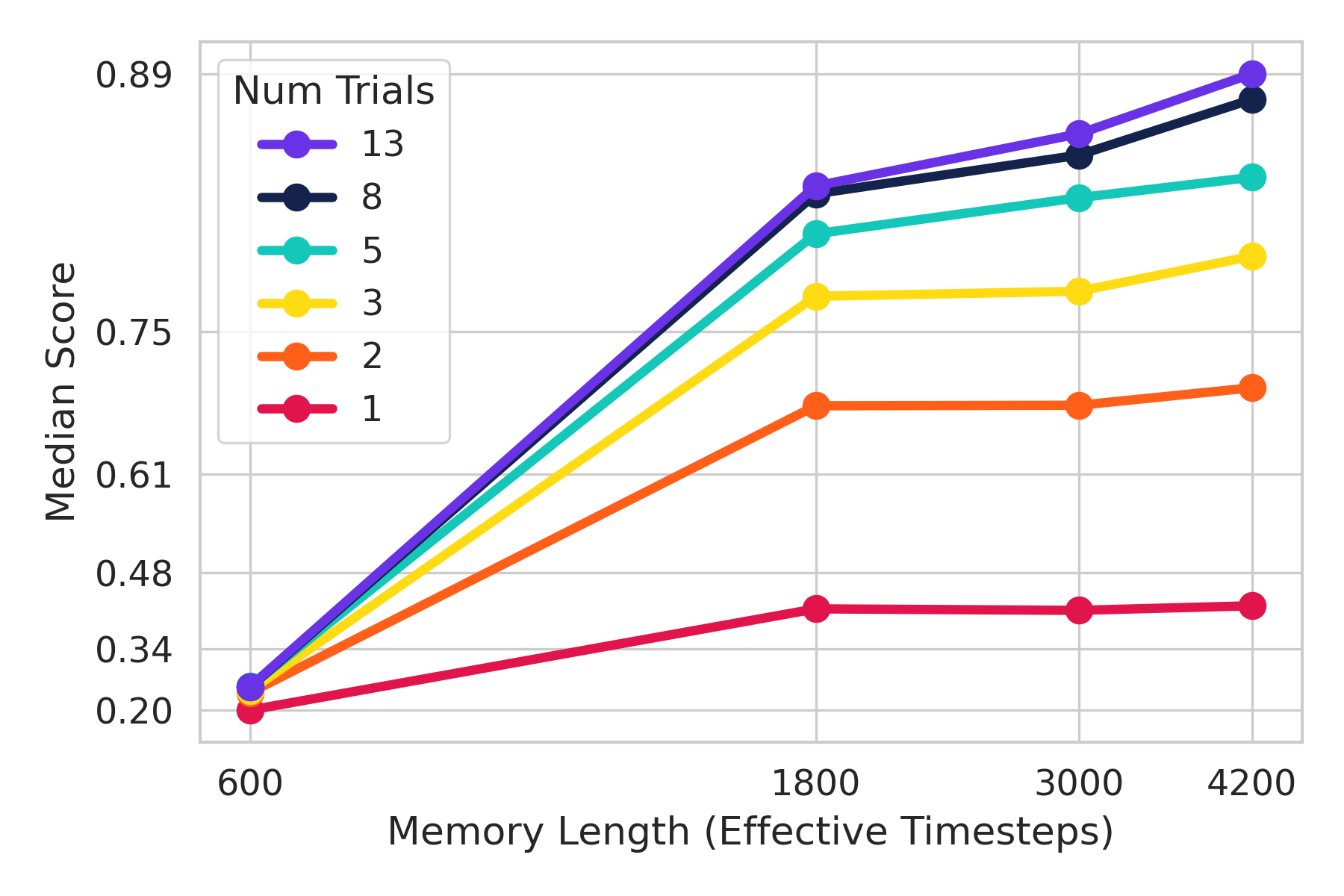}
        \vskip -3pt \caption{}
        \label{figure:scaling_memory_50th}
    \end{subfigure}
    \begin{subfigure}[b]{0.49\textwidth}
        \centering
        \includegraphics[width=\linewidth]{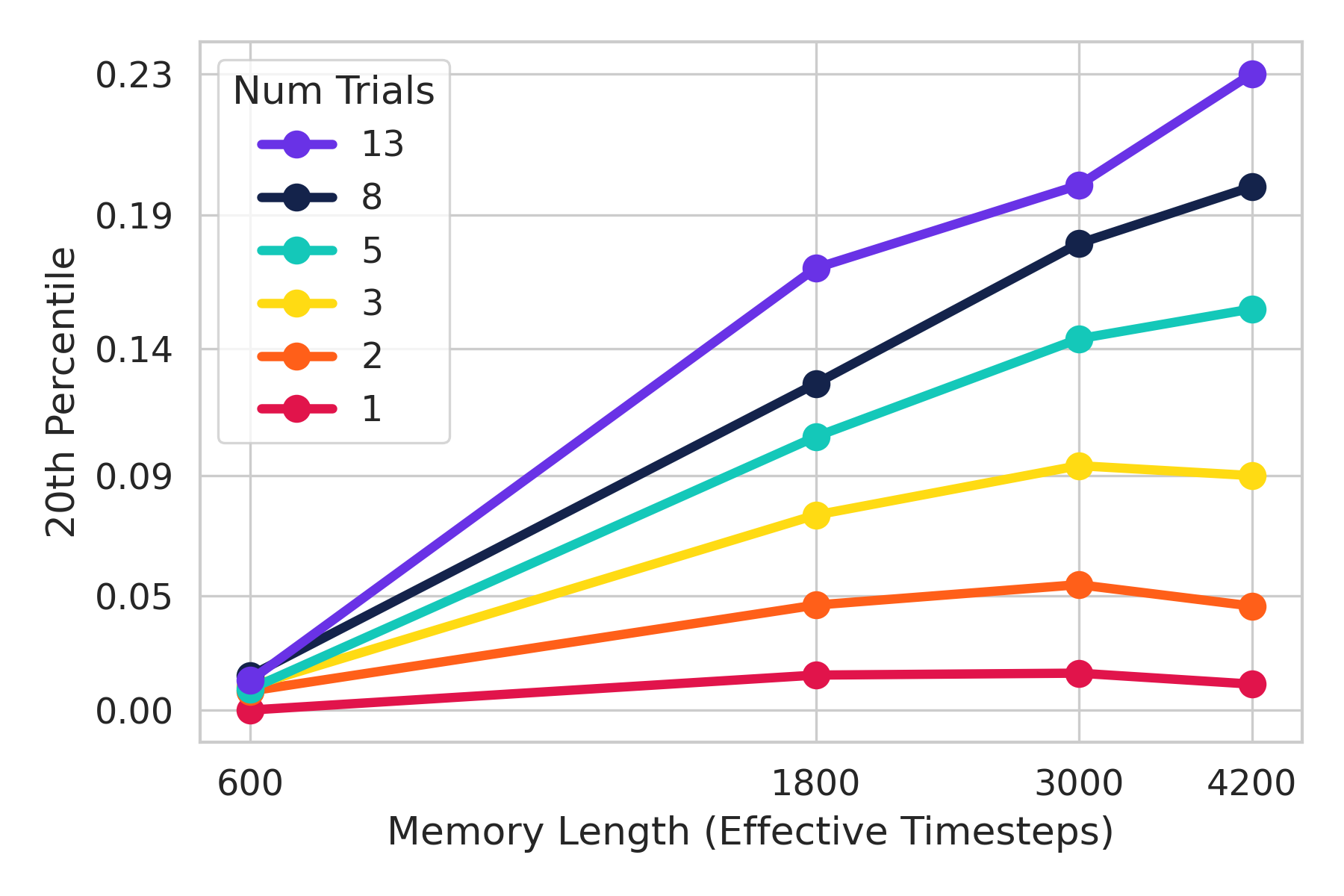}
        \vskip -3pt \caption{}
        \label{figure:scaling_memory_20th}
    \end{subfigure}
    \caption{Scaling Transformer-XL memory length increases 
    both median \textbf{(a)} and 20th percentile \textbf{(b)} test score. Both axes are log-scaled, according to the functions $\log(x)$ and $-\log(1-y)$, respectively, and the relationship between memory length and performance appears roughly linear on this scale. The slope is steeper when evaluating higher numbers of trials, showing that scaling the memory is particularly effective at encouraging stronger few-shot adaptation.}
\label{figure:scaling_memory}
\end{figure}

Appendix \ref{app:network-scaling} details the computational costs of the various model sizes, and shows FLOPs adjusted results. While larger models do indeed have better zero-shot score and adaptation than smaller ones for the same number of training steps, and are more sample efficient, the biggest model may not always be the best choice when compute cost is taken into account.

\paragraph{Scaling memory length.}

Performance also scales with the length of AdA's memory. The experimental setting is shown in Table \ref{tab:memory-scaling}, where we examine the number of previous network activations we cache, investigating values from $100$ to $700$, which, with $6$ Transformer-XL blocks, yields an effective timestep range of $600$ to $4200$ timesteps.\footnote{Transformer-XL enables the use of longer, variable-length context windows by concatenating a cached memory of previous attention layer inputs to the keys and values during each forward pass. Since inputs to intermediate layers are activations from the previous layer, which in themselves contain information about the past, caching $M$ activations theoretically allows for an effective memory horizon of $M \times L$, where $L$ is the number of attention layers in the network. } 

Figure \ref{figure:scaling_memory} shows that, as with model size, scaling memory length helps performance, especially in the lower test percentiles, pushing performance on the tails of the distribution. For any of our tasks, the maximum trial duration is 300 timesteps, so it is interesting that performance on, for example, $5$ trials ($1500$ timesteps) continues to increase for ``effective memory lengths'' between 1800 and 4200. This indicates that it is easier for the Transformer-XL to make use of explicitly given memory activations rather than relying on theoretically longer-range information implicit in those activations.

\subsection{Scaling the task pool increases performance} \label{sec:results_task_scaling}
Another important factor to scale is the amount of data a model is trained on. For example, \citet{hoffmann2022training} showed that in order to get the most out of scaling a language model, one must scale the amount of training data at the same rate as the number of parameters. In our case, relevant data come from interaction with different tasks, so we examine the effect of scaling the number and complexity of different tasks in the XLand pool.

\paragraph{Scaling size of task pool.}

Here we examine the effect of varying the number of training tasks from which the auto-curriculum can sample. Recall that in XLand, a task is the combination of a world (the physical layout of terrain and objects) and a game (specifying the goal and production rules). We investigate the effects of training on tasks sampled from a small pool of 200M distinct tasks (4,000 worlds $\times$ 50,000 games) compared with a large pool of 25B distinct tasks (50,000 worlds $\times$ 500,000 games). Table \ref{tab:task-scaling} shows the full experimental setup for these comparisons.

\begin{figure}[t!]
    \centering
    \begin{subfigure}[b]{0.49\textwidth}
        \centering
        \includegraphics[width=\linewidth]{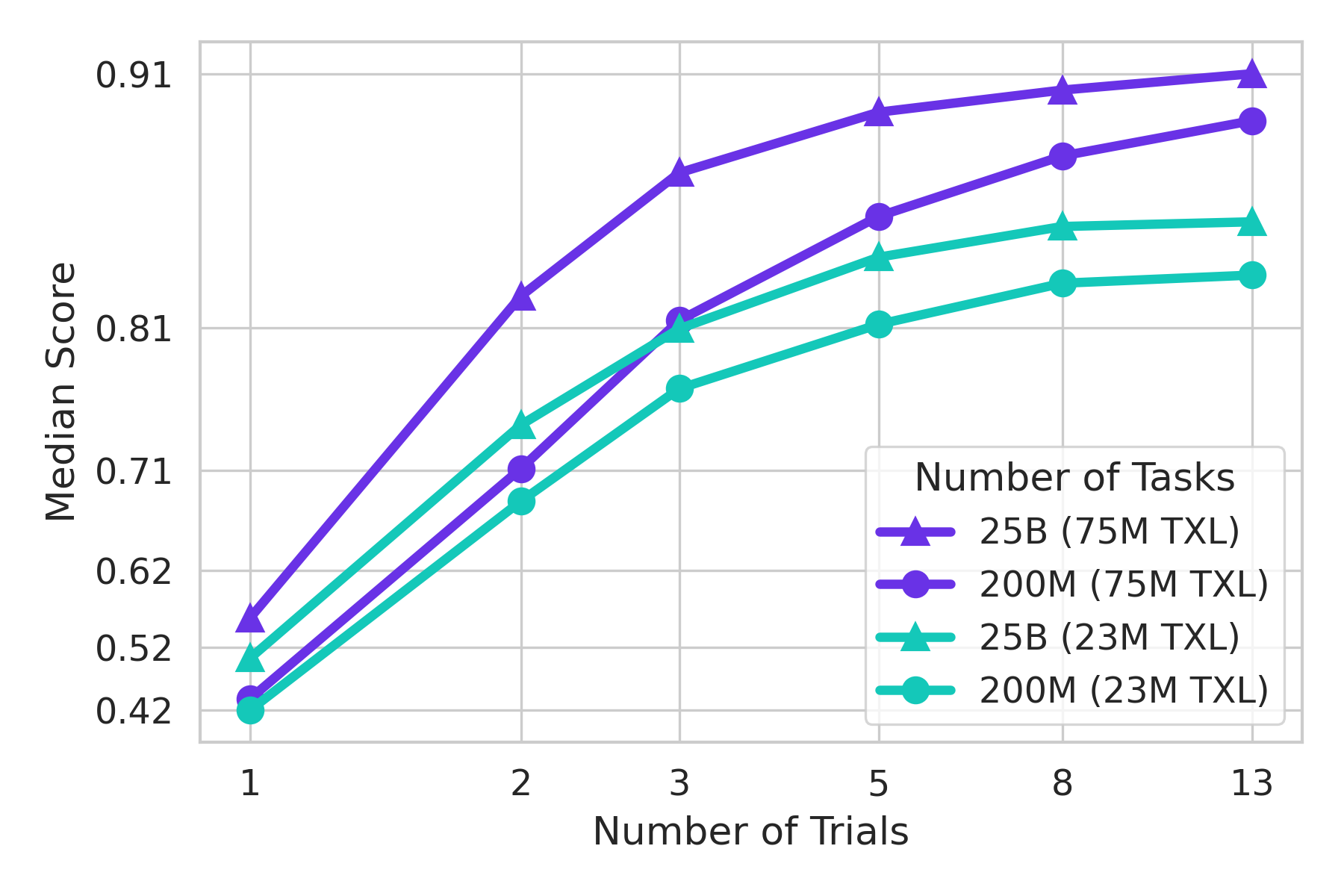}
        \vskip -3pt \caption{}
    \label{figure:scaling_tasks_50th}
    \end{subfigure}
    \begin{subfigure}[b]{0.49\textwidth}
        \centering
        \includegraphics[width=\linewidth]{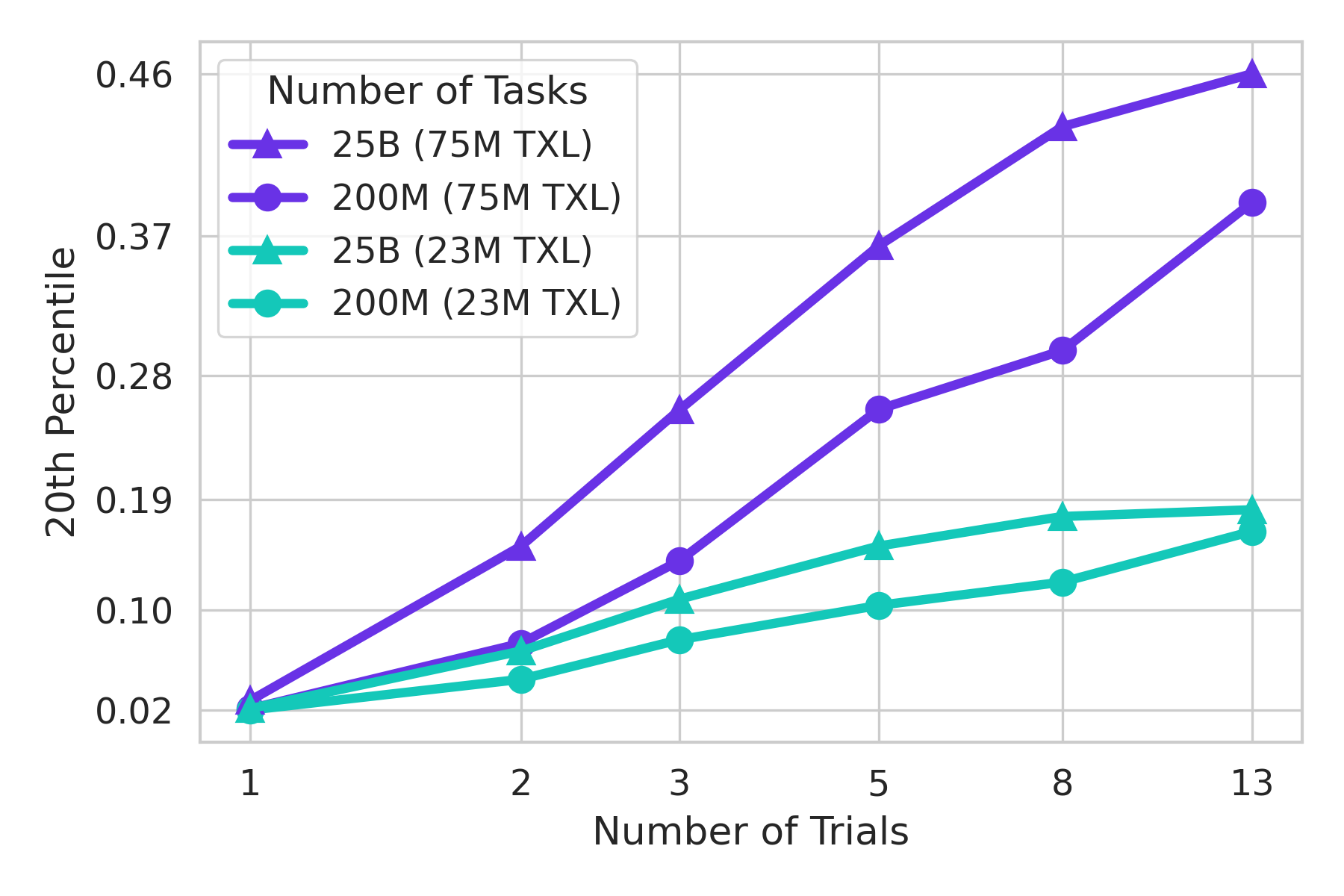}
        \vskip -3pt \caption{}
    \label{figure:scaling_tasks_20th}
    \end{subfigure}
    \caption{
         Median \textbf{(a)} and 20\textsuperscript{th} percentile \textbf{(b)} adaptation scales with the size of the task pool. The effect is especially prominent for larger models. We show the $y$-axis on a logarithmic scale as in the other scaling experiments. Here, we plot number of trials on the $x$-axis and examine the gaps between the curves for the two task distributions (triangle markers vs. circular markers).
    \label{figure:scaling_tasks}
    }
\end{figure}

Figure \ref{figure:scaling_tasks} shows higher test score for identically sized models on the larger task pool. As in the other scaling experiments, we especially see improved performance on the $20$\textsuperscript{th} percentile. The results are shown for two different sizes of models, with the larger Transformer yielding a larger gap when scaling the size of the task pool. This suggests that the large models are especially prone to overfitting to a smaller task pool.

\paragraph{Scaling complexity of task pool.}
One final axis along which it is possible to scale our method is the overall complexity of the task distribution. For example, tasks with a flat terrain will be, on average, less complex to solve than tasks with terrain variation. In Figure \ref{app:scaling_complexity}, we show that low environment complexity can be a bottleneck to scaling, by comparing the effectiveness of model scaling between agents trained on two distributions of the same size but different complexity and evaluated on their respective test sets. Open-ended settings with unbounded environment complexity, such as multi-agent systems, may therefore be particularly important for scaling up adaptive agents.

\subsection{Distillation improves performance and enables scaling agents} \label{sec:results_kickstarting}

\begin{figure}[t!]
    \centering
    \begin{subfigure}[b]{0.49\textwidth}
        \centering
        \includegraphics[width=\linewidth]{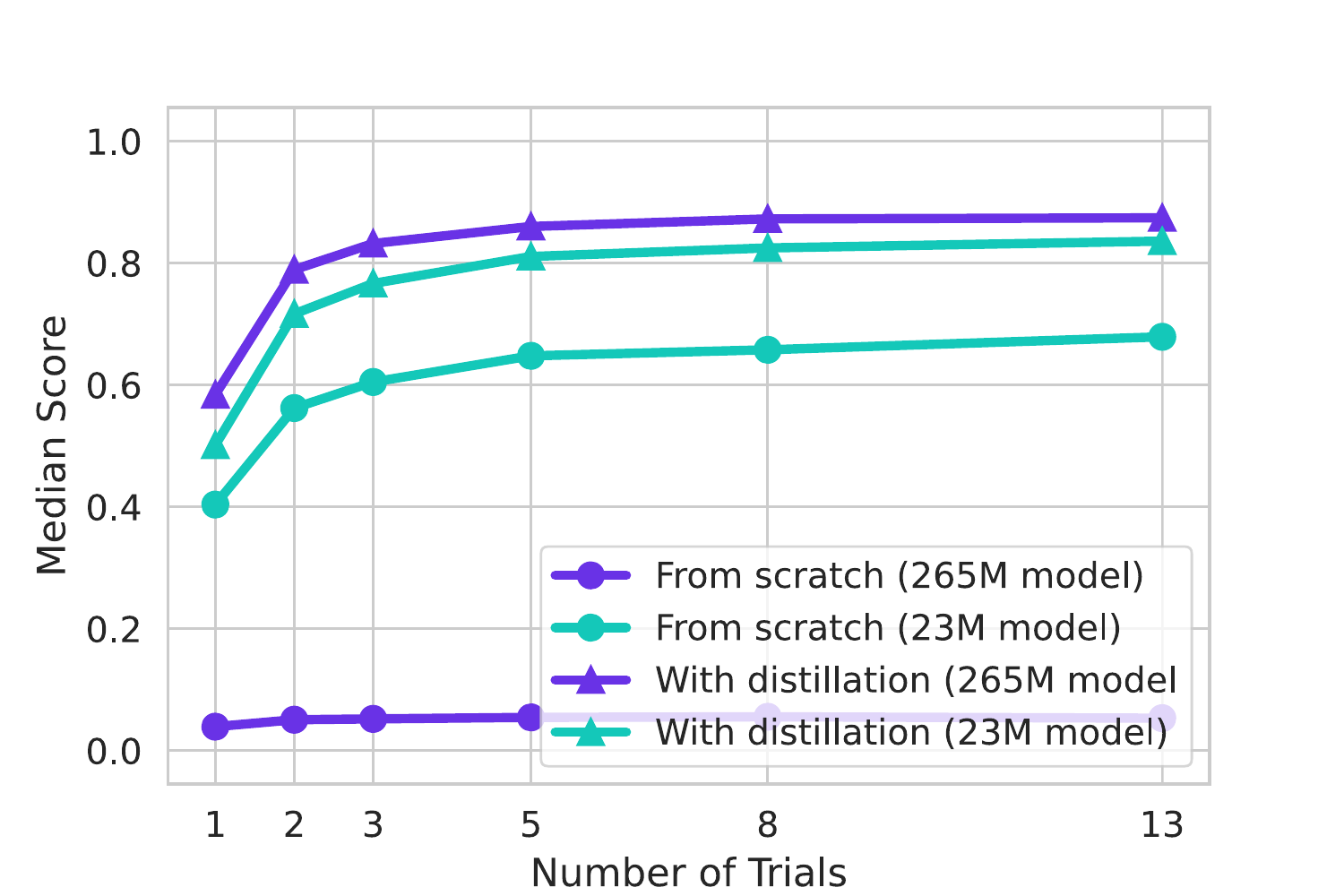}
        \vskip -3pt \caption{}
    \end{subfigure}
    \begin{subfigure}[b]{0.49\textwidth}
        \centering
        \includegraphics[width=\linewidth]{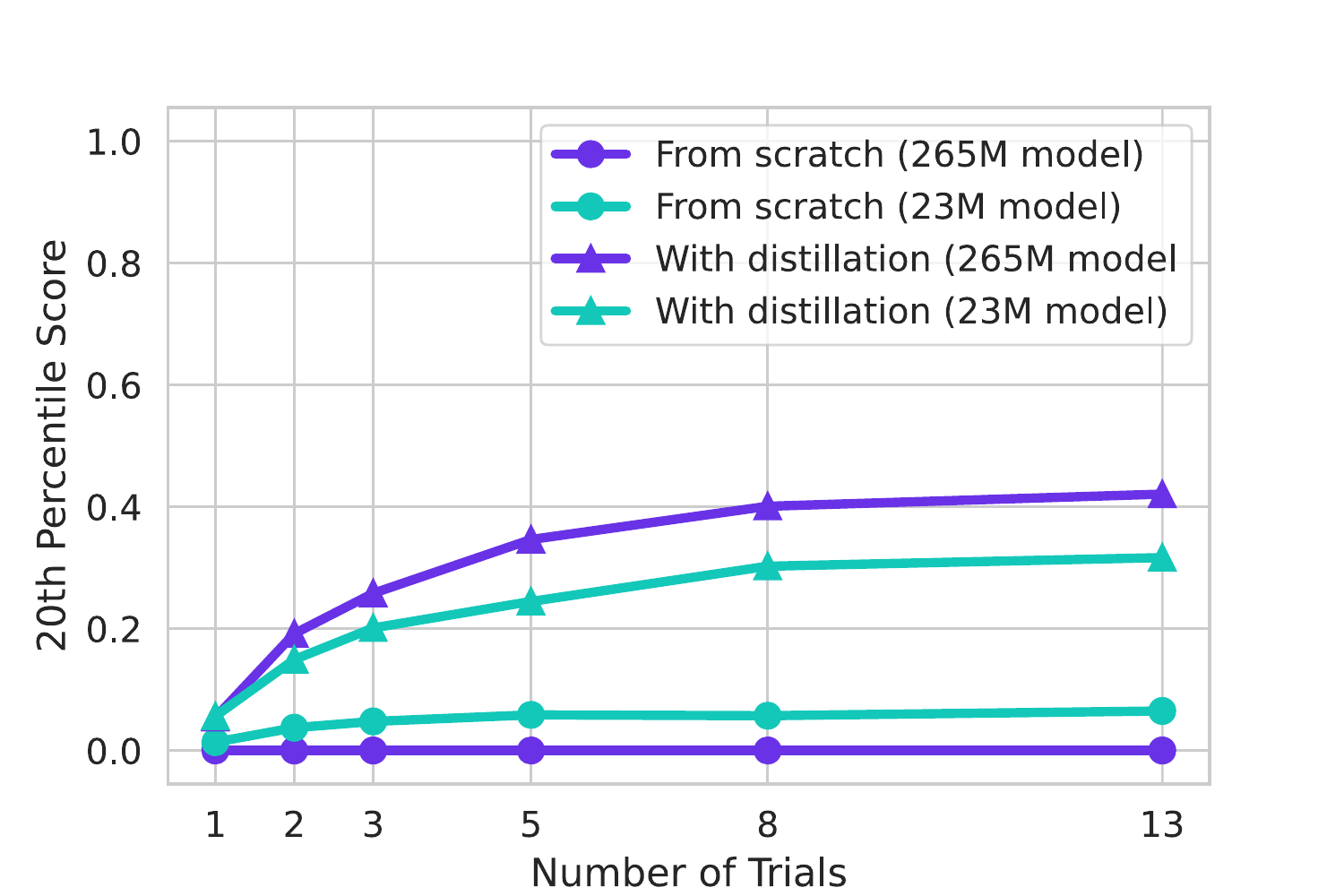}
        \vskip -3pt \caption{}
    \end{subfigure}
    \caption{
        Adaptation over increasing numbers of trials when training from scratch or when kickstarting with distillation, for models with 23M and 265M Transformer parameters. Circle markers show training from scratch while triangle markers show training kickstarted with 4 billion frames of distillation. For this ablation, agents were trained in the multi-agent setup described in Section~\ref{sec:results_human_scale_adaptation} and evaluated on the multi-agent test set after 22 billion total training frames.
        \textbf{(a)} Median score.
        \textbf{(b)} 20\textsuperscript{th} percentile score.
    }
\label{fig:scaling_from_scratch_vs_distillation}
\end{figure}

\begin{figure}[t!]
    \centering
    \begin{subfigure}[b]{0.49\textwidth}
        \centering
        \includegraphics[width=\linewidth]{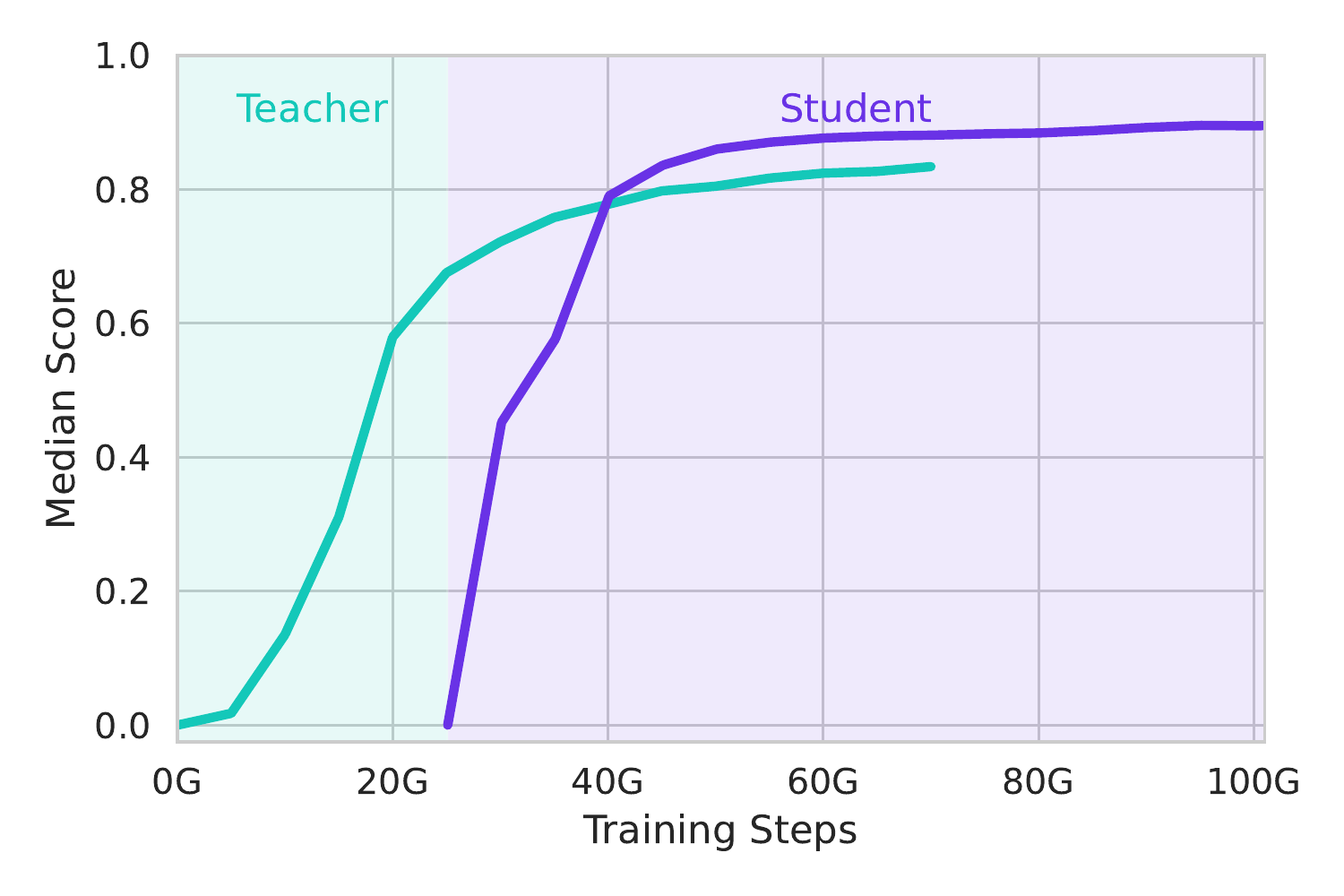} 
        \vskip -3pt \caption{}
    \end{subfigure}
    \begin{subfigure}[b]{0.49\textwidth}
        \centering
        \includegraphics[width=\linewidth]{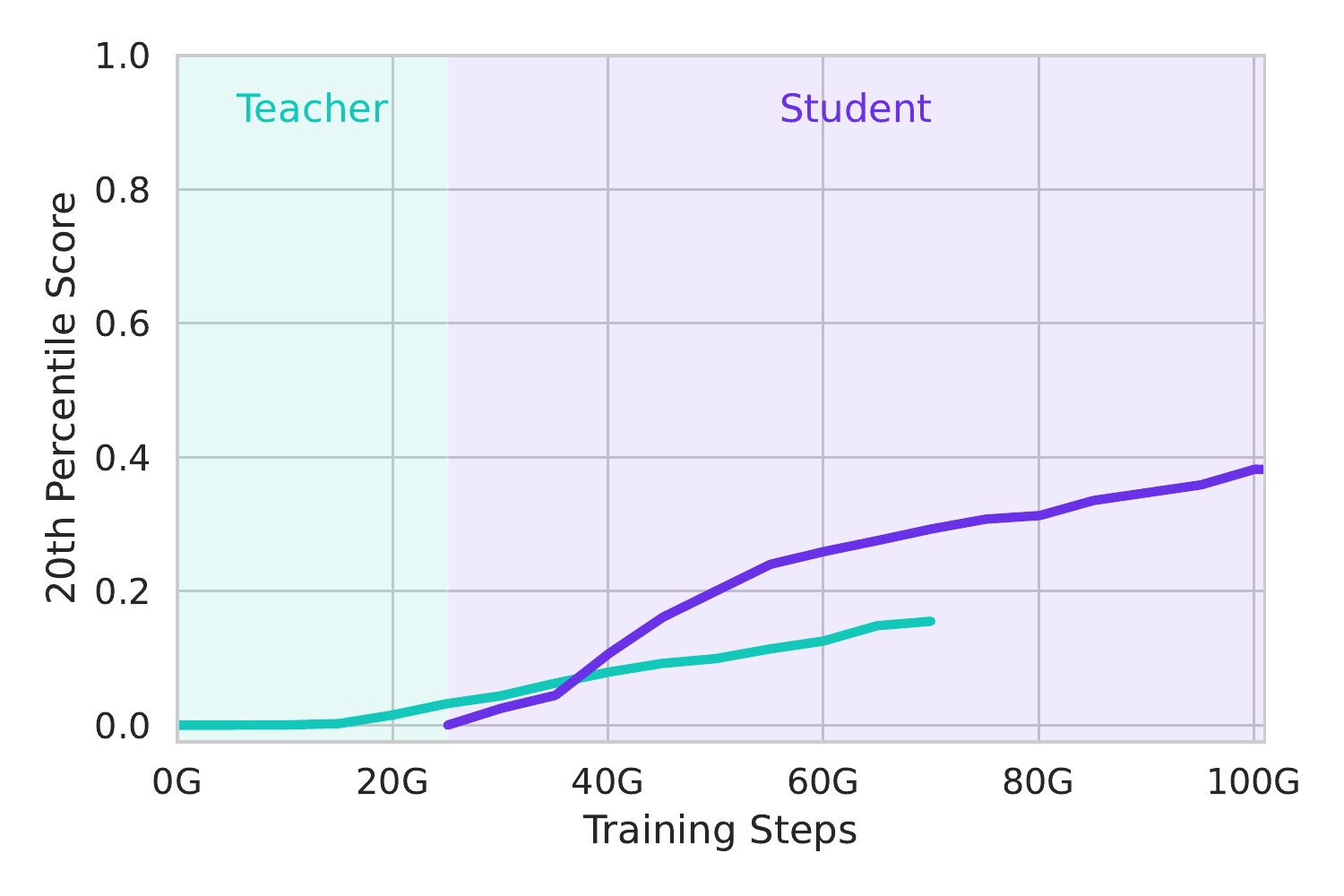} 
        \vskip -3pt \caption{}
    \end{subfigure}
    \caption{
        Normalised last-trial score for $k=13$ using the 23M parameter Transformer-XL. The teacher is trained from scratch, while the otherwise identical student is distilled from a snapshot of the teacher, taken after 25 billion steps of training. The $x$-axis counts the combined amount of experience, including experience used to train the teacher. The comparison shows that distillation can greatly increase the performance of the student, even if the combined amount of experience and updates are equivalent. This is true for median score~\textbf{(a)}, but even more so for the 20\textsuperscript{th} percentile~\textbf{(b)}. 
    }
    \label{fig:distillation}
\end{figure}

All of the scaling comparison experiments shown in the previous section use an identical distillation teacher for the first frames of training, as detailed in Appendix \ref{app:scaling_distillation_teacher}. Now, we look at the role distillation plays in scaling. In short, we find that kickstarting training with a distillation period is crucial when scaling up model size. As shown in Figure~\ref{fig:scaling_from_scratch_vs_distillation}, training a $265$M parameter Transformer model without distillation results in poor performance compared to a much smaller $23$M parameter Transformer trained in the same way. However, when training with distillation from a $23$M parameter teacher for the first 4 billion training frames, the $265$M model clearly outperforms the $23$M variant. See experiment details in Appendix \ref{app:distillation-exp}.

Additionally, we find that even when the model size is the same for both student and teacher, we observe large gains from distillation, for a constant total frame budget (Figure \ref{fig:distillation}). We speculate that this is due to bad representations learned early on by the student agent~\citep{nikishin2022primacy, cetin22stabilizing}, which can be avoided by using distillation. This is also consistent with findings in offline RL, where additional data is often required to effectively scale the model \citep{reid2022wikipedia}. The effect is largest for the first round of distillation, with diminishing returns in subsequent rounds of distillation (Figure \ref{app:repeated_distillation}).

\subsection{Training on more trials with skip memory enables many-shot adaptation} \label{sec:results_few_to_many_shot}

So far, we have considered the few-shot regime in which we train on 1 to 6 trials and evaluate up to 13 trials. In this section, we evaluate AdA's ability to adapt over longer time horizons. We find that when trained with $k \in \{1, 2, \dots 6\}$, agents do not continue to adapt past 13 trials; however, this long-term adaptation capability is greatly improved by increasing the maximum number of trials during training to 24 and increasing the effective length of the memory accordingly. These results show that our method naturally extends to many-shot timescales, with episodes lasting in excess of $30$ minutes.\footnote{$48$ trials of a $40$s task lasts for 32 minutes. By contrast, the average length of a Starcraft 2 game is between $10$ and $15$ minutes, and AlphaStar acted less frequently per-second than AdA does \citep{alphastar}.} In this section, we ablate both factors separately, and show that both are important for long-range adaptation. The training configuration (which is identical to that of the memory scaling experiments save for the number of training steps) is detailed in Table \ref{tab:skip-memory-scaling}.

As we noted in Section \ref{sec:scaling_results}, increasing the memory length leads to increased capacity that benefits the agent even when the entire episode fits in memory, but also comes at the cost of increased computation. To disentangle these factors, we propose a simple change to the memory architecture described in Section \ref{sec:memory-architectures} which increases effective memory length without increasing computational cost. We use a GRU to encode trajectories before feeding them to the Transformer-XL. This allows us to sub-sample timesteps from the encoded trajectories, enabling the agent to attend over 4 times as many trials without additional computation. We show that the additional GRU on its own does not affect the performance of the agent greatly.

\begin{figure}[t]
    \centering
    \begin{subfigure}{0.49\textwidth}
    \includegraphics[width=\linewidth]{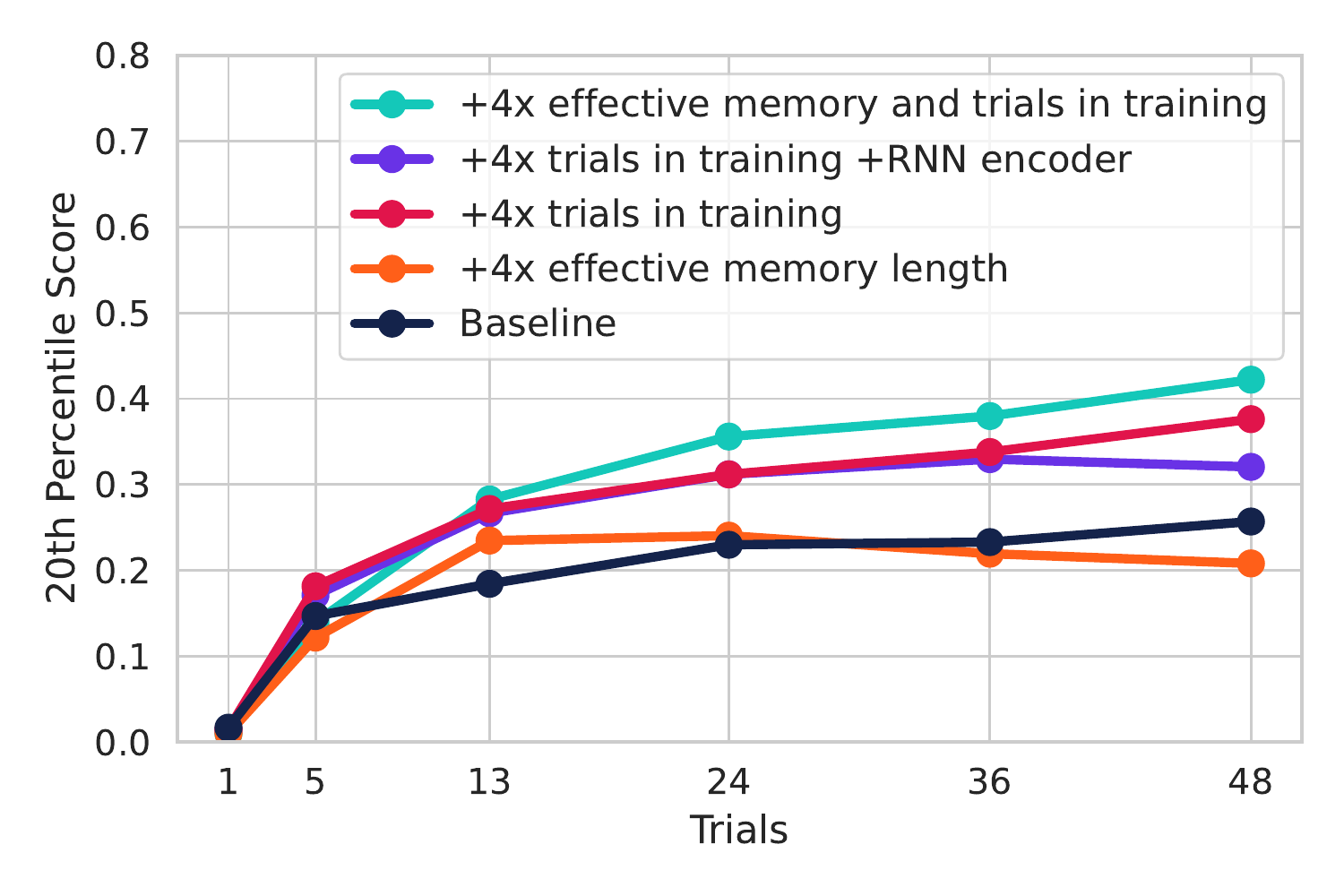} 
     \vskip -3pt \caption{}
    \label{fig:manyshot_ablation}
    \end{subfigure}
    \begin{subfigure}{0.49\textwidth}
    \includegraphics[width=\linewidth]{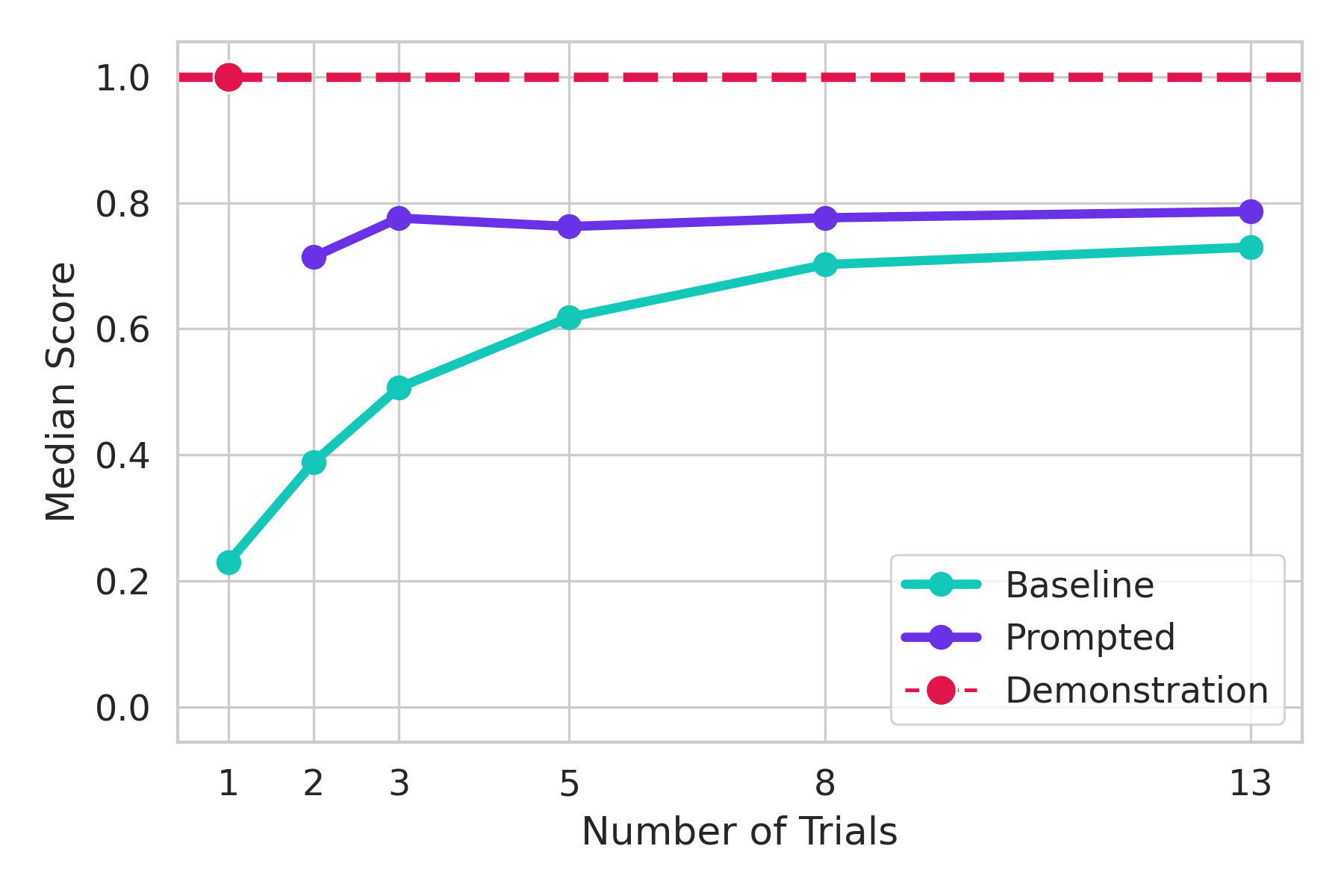}
     \vskip -3pt \caption{}
    \label{fig:first-person-demonstration}
    \end{subfigure}
    \caption{\textbf{(a)} Ablation showing the 20\textsuperscript{th} percentile of test scores as we vary the maximum number of training trials (from a $k=6$ baseline to $k=24$) and increase the effective memory size via sub-sampling (from 1800 steps to 7200 steps). Together, these factors enable the agent to adapt over a larger number of trials (lasting over 30 minutes). Increasing the number of training trials has the biggest effect and is a prerequisite for sub-sampling to be effective. This figure furthermore shows that adding an RNN encoder to facilitate sub-sampling does not by itself greatly affect performance. \textbf{(b)} Median hand-authored task score of AdA prompted with a first-person demonstration in the first trial of each episode, compared with an unprompted baseline. The prompted score lies strictly above the baseline which indicates that AdA is able to use information from a demonstration prompt to improve its performance. However, the score lies below that of the demonstration which suggests that it is not able to make perfect use of the demonstration.}
\end{figure}

As can be seen in Figure~\ref{fig:manyshot_ablation}, increasing the number of trials in the training distribution significantly boosts performance in later trials, especially when the memory length is scaled accordingly. In other words, the adaptation strategy learned by AdA benefits from experiencing a large number of trials, rather than just very recent ones. Therefore we can conclude that AdA is capable of adaptation based on long-term knowledge integrated into memory across many trials, as opposed to merely encoding a simple meta-strategy that only depends on the trajectory from the previous trial.  

Increasing the number of trials in training leads to better adaptation even in the absence of increased memory. This indicates that the agent is able to learn better exploration and refinement strategies when afforded longer training episodes consisting of more trials. Note that increasing effective memory without increasing the number of training trials does not improve performance, as the agent has not been trained to make use of the additional memory capacity.

\subsection{AdA can leverage prompting with first-person demonstrations}

To determine whether AdA can learn in zero-shot from first-person demonstrations, we prompted it with a first-person demonstration by a fine-tuned teacher, as follows. The teacher took control of the avatar in the first trial, while AdA continued to receive observations as usual, conditioning its Transformer memory. AdA was then allowed to proceed on its own for the remaining trials and its scores recorded in the usual manner. 

Figure \ref{fig:first-person-demonstration} shows the median score on our hand-authored test set of prompted AdA compared to an unprompted baseline. Prompted AdA is unable to exactly mimic the teacher's demonstration in the second trial of a median task, shown by a drop in score. It does, however, outperform an unprompted baseline across all numbers of trials, indicating that it is able to profitably incorporate information from the demonstration into its policy. This process is analogous to prompting of large language models, where the agent's memory is primed with an example of desired behaviour from which it continues. We note that AdA was never trained with such off-policy first-person demonstrations, yet its in-context learning algorithm is still able to generalise to these. 

In Figure \ref{fig:prompting-all-tasks} we provide prompting results for all single-agent hand-authored tasks and discuss the circumstances under which prompting is effective. In Appendix \ref{app:prompting} we also provide early results investigating prompting with human demonstrations on a subset of tasks. These reveal remarkable success in some cases, but also confirm that human demonstrations cannot  overcome inherent limitations of AdA's task distribution. Two videos compare the behaviour when \href{https://youtu.be/jdf4jkiJYa4}{prompted} and when \href{https://youtu.be/CuxPknjnYOI}{not prompted} on the task \texttt{Object permanence: yellow cube}.

\section{Related Work} \label{sec:related_work}

In this work, we leverage advances in attention-based models for meta-learning in an open-ended task space. Our agent learns a form of in-context RL algorithm, while also automatically curating the training task distribution; thus we combine two pillars of an \emph{AI generating algorithm} (AI-GA,~\citet{clune2019aiga}). The most similar work to ours is \citet{xland}, which also considers training in a vast multi-agent task space with auto-curricula and generational learning. A key difference in our work is that we focus on \emph{adaptation} (vs. zero-shot performance), and make use of large Transformer models. \citet{akkaya2019solving} also demonstrated the effectiveness of adaptive curricula while meta-learning a policy to control a robot hand, however they focused on a specific sim-to-real setting rather than a more generally capable agent. We now summarise literature related to each component of our work in turn. 

\paragraph{Procedural environment generation.} We make use of procedural content generation (PCG) to generate a vast, diverse task distribution. PCG has been studied for many years in the games community \citep{togelius2008evolving, pcg} and more recently has been used to create testbeds for RL agents \citep{pcg_illuminating, coinrun, raileanu2020ride}. Indeed, in the past few years a series of challenging PCG environments have been proposed \citep{obstacletower, cobbe2020procgen, kuettler2020nethack, samvelyan2021minihack, gym_minigrid, hafner2022benchmarking, https://doi.org/10.48550/arxiv.2206.06994}, mostly focusing on testing and improving generalisation in RL \citep{rl_generalization_survey, bhatt2022dsage, fontaine2021overcooked}. More recently there has been increased emphasis on open-ended worlds: \citet{albrecht2022avalon} proposed Avalon, a 3D world supporting complex tasks, while Minecraft \citep{johnson2016malmo} has been proposed as a challenge for Open-Endedness and RL \citep{minerl2021, fan2022minedojo, evocraft2021}, but unlike XLand it does not admit control of the full simulation stack, thereby limiting the smoothness of the task space. 

\paragraph{Open-ended learning.} A series of recent works have demonstrated the effectiveness of agent-environment co-adaptation with a distribution of tasks \citep{wang2019poet, wang2020enhanced_poet, accel}. Our approach bears resemblance to the unsupervised environment design (UED, \citet{dennis2020paired}) paradigm, since we seek to train a generalist agent without knowledge of the test tasks. One of the pioneering methods in this space was PAIRED~\citep{dennis2020paired}, which seeks to generate tasks with an RL-trained adversary. We build on Prioritised Level Replay \citep{jiang2020prioritized, jiang2021robustplr}, a method which instead curates randomly sampled environments which have high regret. Our work also relates to curriculum learning \citep{matiisen2020tscl, portelas2019teacher, sukhbaatar2018asp, openai2021asymmetric, campero2020amigo, fang2021adaptive,mu2022improving}, with the key difference that these methods typically have a specific downstream goal or task in mind. There have also been works training agents with auto-curricula over co-players, although these typically focus on singleton environments \citep{alphastar, dota} or uniformly sampled tasks \citep{Baker2020Emergent, Jaderberg859, liu2018emergent, https://doi.org/10.48550/arxiv.2203.00715}. Similar to XLand 2.0's production rule system, \citet{DBLP:conf/iclr/ZhongRG20} train agents to generalise to unobserved environment dynamics. However, they investigate zero-shot generalisation where the agent has to infer underlying environment dynamics from language descriptions, whereas AdA agents discovery these rules at test time via on-the-fly hypothesis-driven exploration over multiple trials.

\paragraph{Adaptation.} 
This work focuses on few-shot adaptation in control problems, commonly framed as \emph{meta-RL}. We focus on \emph{memory-based} meta-RL, and build upon the work of \citet{duan2017rl} and \citet{wang2016learning}, who showed that if an agent observes rewards and terminations, and the memory does not reset, a memory-based policy can implement a learning algorithm. This has proven to be an effective approach that can learn Bayes-optimal strategies~\citep{ortega2019metalearning,mikulik2020metatrained} and may have neurological analogues~\citep{wang2018prefrontal}. Indeed, our agents learn conceptual exploration strategies, something that would require the outer learner of a meta-gradient approach to estimate the return of the inner learner~\citep{stadie2018some}. Solutions in this space either rely on high-variance Monte Carlo returns~\citep{stadie2018some,garcia2019meta,vuorio2021no} or history-dependent estimators~\citep{zheng2020can}. Our work is also inspired by Alchemy~\citep{wang2102alchemy}, a meta-RL benchmark domain whose mechanics have inspired the production rules in our work. The authors use memory-based meta-RL with a small Transformer, but find that the agent's performance is only marginally better than that of a random heuristic. Transformers have also been shown to be effective for meta-RL on simple domains \citep{melo2022transformers} and for learning RL algorithms \citep{laskin2022algorithmdistillation} from offline data. Other approaches for meta-RL include meta-gradients~\citep{andrychowicz2016learning,maml_finn_icml17,xu2018meta, flennerhag2022bootstrapped}, which can be efficient but often suffer from instability and myopia~\citep{metz2021gradients,vuorio2021no,flennerhag2022bootstrapped}, and latent-variable based approaches~\citep{finn2018probabilistic,humplik2019meta,rakelly2019efficient,zintgraf2019varibad}. Adaptation also plays a critical role in robotics, with agents trained to adapt to varying terrain \citep{clavera2018learning, kumar2021rma} or damaged joints \citep{Cully2015RobotsTC}.

\paragraph{Transformers in RL and beyond.} Transformer architectures have recently shown to be highly effective for  \emph{offline} RL \citep{chen2020decision, janner2021sequence, reed2022a}, yet successes in the \emph{online} setting remain limited. One of the few works to successfully train Transformer-based policies was \citet{parisotto2020stabilizing}, who introduced several heuristics to stabilise training in a simpler, smaller-scale setting. Indeed, while we make use of a similar Transformer-XL architecture \citep{vaswani2017attention, dai2019transformer}, we demonstrate scaling laws for online meta-RL that resemble those seen in other communities, such as language \citep{DevlinCLT19, kaplan2020scaling, brown2020gpt3, rae2021scaling}. Similarly, \citet{melo2022transformers} use Transformers for fast adaptation in a smaller-scale meta-RL setting, interpreting the self-attention mechanism as a means of building an episodic memory from timestep embeddings, through the recursive application of Transformer layers. Transformer architectures have also been used in meta-learning outside of RL, for example learning general-purpose algorithms \citep{kirsch2022general} or hyperparameter optimisers \citep{optformer}. Transformers are also ubiquitous in modern large language models, which have been shown to be few-shot learners \citep{brown2020gpt3}. 

\section{Conclusion}  

Adaptation to new information across a range of timescales is a crucial ability for generally intelligent agents. Foundation models in particular have demonstrated an ability to acquire a large knowledge-base of information, and apply this rapidly to new scenarios. Thus far, they have relied mainly on supervised and self-supervised learning. As such, they require access to large datasets. An alternative to collecting datasets is to have an agent learn from its own experience via reinforcement learning, provided that sufficiently rich physical worlds or open-ended simulations are available. This raises the question: can large-scale, generally adaptive models be trained with RL?

In this paper, we demonstrate, for the first time to our knowledge, an agent trained with RL that is capable of rapid in-context adaptation across a vast, open-ended task space, at a timescale that is similar to that of human players. This \textit{Adaptive Agent} (AdA) explores held-out tasks in a structured way, refining its policy towards optimal behaviour given only a few interactions with the task. Further, AdA is amenable to contextual first-person prompting, strengthening its few-shot performance, analogous to prompting in large language models. AdA shows scaleable performance as a function of number of parameters, context length and richness of the training task distribution. 

Our training method is based on black-box meta-RL, previously thought of as hard to scale. We show that state-of-the-art automatic curriculum techniques can shape the data distribution to provide sufficient signal for learning to learn in an open-ended task space. Moreover, we demonstrate that attention-based architectures can take advantage of this signal much more effectively than purely recurrent networks, illustrating the importance of co-adapting data-distribution and agent architecture for facilitating rapid adaptation. Finally, distillation enables us to realise the potential of large-scale Transformer architectures. 

The future of AI research will inevitably involve training increasingly large models with increasingly general and adaptive capabilities. In this direction, we have provided a recipe for training a 500M parameter model, which we hope can pave the way for further advances at the intersection of RL and foundation models. AdA shows rapid and scalable adaptation of myriad kinds, from tool use to experimentation, from division of labour to navigation. Given scaling law trends, such models may in future become the default foundations for few-shot adaptation and fine-tuning on useful control problems in the real world.

\section{Authors and Contributions}
\label{sec:authors}

We list authors alphabetically by last name. Please direct all correspondence to Feryal Behbahani (\href{mailto:feryal@deepmind.com}{feryal@deepmind.com}) and Edward Hughes (\href{mailto:edwardhughes@deepmind.com}{edwardhughes@deepmind.com}). 

\subsection{Core contributors}

\begin{itemize}
    \item \textbf{Jakob Bauer}: technical leadership, curriculum research, infrastructure engineering, task authoring, paper writing
    \item \textbf{Kate Baumli}: agent research, scaling, agent analysis, task authoring, paper writing
    \item \textbf{Feryal Behbahani}: research vision, team leadership, agent research, paper writing
    \item \textbf{Avishkar Bhoopchand}: technical leadership, evaluation research, infrastructure engineering, task authoring, paper writing
    \item \textbf{Michael Chang}: visualisation, agent analysis, human experiments
    \item \textbf{Adrian Collister}: XLand development, human experiments
    \item \textbf{Edward Hughes}: research vision, team leadership, evaluation research, paper writing
    \item \textbf{Sheleem Kashem}: infrastructure engineering, curriculum research, human experiments 
    \item \textbf{Jack Parker-Holder}: curriculum research, paper writing
    \item \textbf{Yannick Schroecker}: agent research, scaling, task authoring, agent analysis, paper writing
    \item \textbf{Jakub Sygnowski}: infrastructure engineering, curriculum research, agent analysis, paper writing
    \item \textbf{Alexander Zacherl}: design leadership, agent analysis, task authoring, visualisation, human experiments
    \item \textbf{Lei Zhang}: curriculum research, agent analysis, paper writing
\end{itemize}

\subsection{Partial contributors}

\begin{itemize}
    \item \textbf{Nathalie Bradley-Schmieg}: project management
    \item \textbf{Natalie Clay}: QA testing, human experiments
    \item \textbf{Vibhavari Dasagi}: evaluation research
    \item \textbf{Lucy Gonzalez}: project management
    \item \textbf{Karol Gregor}: agent research
    \item \textbf{Maria Loks-Thompson}: XLand development, human experiments
    \item \textbf{Hannah Openshaw}: project management
    \item \textbf{Shreya Pathak}: agent analysis
    \item \textbf{Nicolas Perez-Nieves}: agent analysis, task authoring
    \item \textbf{Nemanja Rakicevic}: curriculum research, agent analysis
    \item \textbf{Tim Rockt\"aschel}: strategic advice, paper writing
    \item \textbf{Sarah York}: QA testing, human experiments
\end{itemize}

\subsection{Sponsors}

\begin{itemize}
    \item \textbf{Satinder Baveja}: strategic advice
    \item \textbf{Karl Tuyls}: strategic advice
\end{itemize}

\section{Acknowledgements}

We thank Max Jaderberg for early guidance on the project vision. We are grateful to Wojciech Marian Czarnecki for an early version of the production rules formalism and Catarina Barros for a prototype implementation. We thank Dawid G\'orny for support on implementing visualisation tools. We are grateful to Alex Platonov for artistic rendering of the figures and accompanying videos. We thank Nathaniel Wong, Tom Hudson and the Worlds Team for their engineering support. Further, we thank Andrew Bolt, Max Cant, Valentin Dalibard, Richard Everett, Nik Hemmings, Shaobo Hou, Jony Hudson, Errol King, George-Cristian Muraru, Alexander Neitz, Valeria Oliveira, Doina Precup, Drew Purves, Daniel Tanis, Roma Patel, and Marcus Wainwright for useful discussions and support. We are grateful to Sebastian Flennerhag and Raia Hadsell for reviewing a draft of the paper.

\bibliography{main}

\newpage
\begin{appendix}

\counterwithin{figure}{section}
\counterwithin{table}{section}

\section*{Appendix}

\section{Environment Details}
\subsection{XLand 2.0}
\label{app:xland_changes}

In this section we describe the differences between XLand 2.0 and the original XLand environment of \cite{xland}. We modify the configuration space as follows:

\begin{itemize}
    \item We introduce a new relation \texttt{touching(a,b)}. It is satisfied if objects \texttt{a} and \texttt{b} are in contact, as determined by Unity's collision detection with a distance threshold of 1 millimeter.
    \item We exclude all relations that refer to floors. As a more flexible alternative we introduce the option of spawning objects in a permanently frozen state, rendering them immobile. Frozen objects can be used as anchor points in the environment which require players to navigate to them by including them in goals or production rules.
    \item We only use predicates consisting of a single relation or its negation, excluding conjunctions or disjunctions of multiple relations. Note that production rules (Section~\ref{sec:methods_xverse}) allow us to specify tasks which require the player to sequentially satisfy predicates, or to give them multiple ways to reach a desired state.
\end{itemize}

For the reader's convenience, Tables \ref{tab:objects} and \ref{tab:colors} respectively list all shapes and colours we used for objects in XLand. Table \ref{tab:predicates} lists all predicates we used for goals and production rules.

\begin{longtable}{|p{0.25\linewidth}|p{0.1\linewidth}p{0.52\linewidth}}
    \caption{Shapes used for objects.}
    \label{tab:objects}
    \\ \hline
      \textbf{Shape name}\\
      \hline
      Wall \\
      \hline
      Cube \\
      \hline
      Sphere \\
      \hline
      Pyramid \\
      \hline
\end{longtable}

\begin{longtable}{|p{0.25\linewidth}|p{0.1\linewidth}|p{0.52\linewidth}}
    \caption{Colours used for objects.}
    \label{tab:colors}
    \\ \hline
      \textbf{Colour name} \\
      \hline
      Black \\
      \hline
      Purple \\
      \hline
      Yellow \\
      \hline
\end{longtable}

\begin{longtable}{>{\raggedright}p{0.25\linewidth}|p{0.62\linewidth}}
    \caption{Predicates used for goals and production rules.}
    \label{tab:predicates}
    \\ \hline
      \textbf{Predicate name} & \textbf{Meaning} \\
      \hline
      \texttt{touching(a,b)} & Whether \texttt{a} and \texttt{b} are in contact. \\
      \hline
      \texttt{near(a,b)} & Whether \texttt{a} and \texttt{b} are at most 1m apart.\\
      \hline
      \texttt{hold(a,b)} & Whether player \texttt{a} holds \texttt{b}.\\
      \hline
      \texttt{see(a,b)} & If \texttt{a} is a player, whether it can see \texttt{b}. If not, whether the line connecting the centres of mass of \texttt{a} and \texttt{b} is not obstructed by another object.\\
      \hline
      \texttt{not(p)} & Whether predicate \texttt{p} is not satisfied. \\
      \hline
\end{longtable}

\subsection{Pre-sampling tasks for training} \label{app:presample_tasks}

The space of goals and production rules can generate at least $10^{40}$ distinct tasks, even given quite restrictive bounds on the number of objects and rules.\footnote{This is an order of magnitude lower-bound estimate, assuming $4$ shapes, $3$, colours, $7$ predicates, a maximum of $5$ production rules with a maximum of $3$ objects on the right-hand side, $7$ blanking options, and a maximum of $20$ objects in the scene.} For convenience, we pre-sample a subset of this space using the method described below. In Section \ref{sec:results_task_scaling} we evaluate the effect of the size of the sampled set. For each task we sample a world using the procedure outlined in \citet{xland} and combine it with a game and production rules as follows.

\paragraph{Single-player tasks.} We start by uniformly sampling a player's goal, consisting of a predicate with optional negation and two objects. Then, for a fixed number of steps (which we sample uniformly between 1 and 4), we add new production rules, such that they need to be triggered in sequence to get the objects present in the goal. We initialise the world to contain the objects present in the condition of the first production rule, together with up to $10$ \textit{distractor} objects, not present in any production rule from this chain (nor in the goal).

Next, we introduce {\it dead-end} production rules. We sample them such that their condition contains either distractor objects or ones that are ultimately necessary for satisfaction of the goal, yet the spawns are always distractors. As such, triggering a dead end may put the game in an unsolvable state. Including them in the training games creates pressure on the agent to avoid indiscriminately triggering all production rules. Finally, we sample a hiding mask, specifying which part of the production rules will be visible or hidden from the player. The sampling of both the mechanism for hiding (described in Section~\ref{sec:methods_xverse}) and the actual parts to hide is uniform random.

\paragraph{Multi-player tasks.} For this work we restrict our multi-player games to fully cooperative two-player games. Such games are known to be particularly challenging, as they have multiple equilibria, and thus feature an equilibrium selection problem \citep{DBLP:journals/corr/abs-2012-08630}. To sample such a game, we start by sampling a single-player game as outlined above and randomly replace references to the first player in the goal or production rules with references to player two. We copy the goal and sample a new production rule hiding mask for player two, resulting in a fully cooperative two-player game with potentially asymmetric information provided about the task's production rules.
\section{Evaluation} \label{sec:methods_evaluation}\label{app:evaluation}

\subsection{Test scores} We evaluate our agents on a suite of 1000 held out test tasks  sampled from the same distribution as the training games, using held-out world topologies. The procedure for pre-generating the XLand task pool is detailed in Appendix \ref{app:presample_tasks}. Rejection sampling ensures that no game (goal and production rules) in the test set is contained in the training set.

In XLand, rewards are obtained on every frame in which a goal is satisfied, making total rewards incomparable between tasks. To account for this, we fine-tune AdA on the test-task set. We compute the fine-tuned agent's maximum total last-trial reward
(over any number of trials up to 13)
and use this as an estimate of the maximum possible reward obtainable in a single trial of each test task. We call this quantity the normaliser. We define the \textit{test score} $S^i_m$ of an agent $i$ on task $m$ with $k$ trials to be the total reward obtained in trial $k$ divided by the normaliser. This normalises rewards roughly to the interval $[0, 1]$. Note that it is possible for an agent under evaluation to obtain a score greater than 1, both due to noise in the evaluation process and the fact that the agent under evaluation may be better than the one used for creating the normaliser.

When reporting the scores of our agents on a game with $k$ trials, we always use the total reward of the last trial. This is a good measure of whether the agent has successfully navigated the exploration-exploitation tradeoff in a novel MDP, given knowledge of the number of trials $k$. If an agent is capable of adaptation, we expect to see the performance in the last ($k$\textsuperscript{th}) trial increase as a function of $k$: that is to say, the agent is able to make use of additional experience on-the-fly to perform better. We evaluate on $k \in \{1, 2, 3, 5, 8, 13\}$, where $8$ and $13$ are held out values of $k$ that were not seen during training.

To aggregate the scores of an agent across games, we use a fixed (usually 20\textsuperscript{th}) percentile score \citep{DBLP:journals/corr/abs-2108-13264}. This gives us a lower bound guarantee on the performance of the agent across most of the tasks: for example, if the 20\textsuperscript{th} percentile score of an agent is $0.5$, then the agent gets the score of at least $0.5$ on $80\%$ of the games. Using an aggregation method like this (as opposed to an average) allows us to concentrate on the coverage of many tasks, as opposed to focusing the effort on improving the performance on outlier tasks. Empirically, we find that our results are robust across a range of different percentiles. 

\subsection{Hand-authored probe tasks} 
\begin{figure}
  \begin{minipage}[c]{0.65\textwidth}
        \includegraphics[width=1\textwidth]{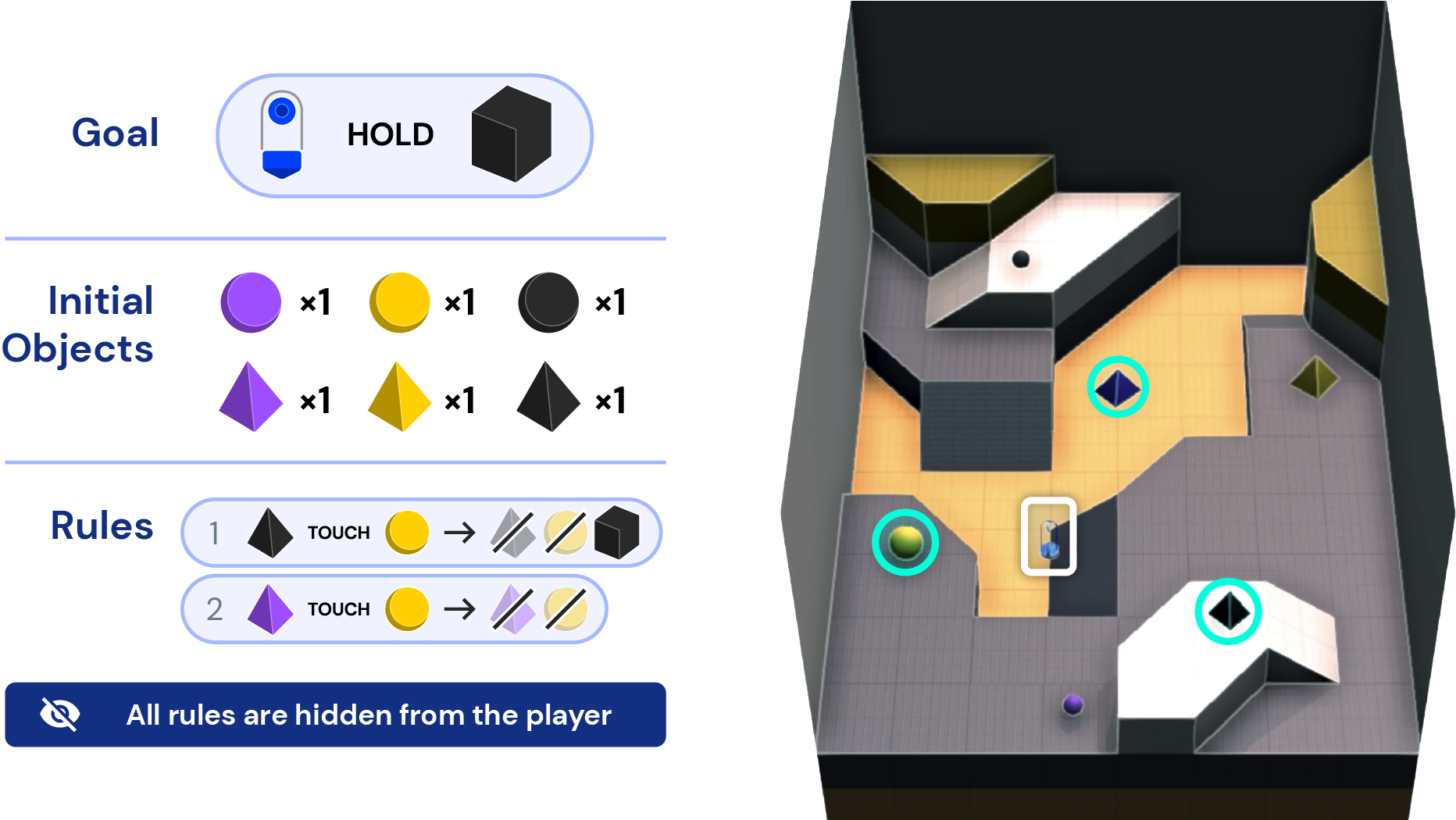}
  \end{minipage}\hfill
  \begin{minipage}[c]{0.3\textwidth}
    \caption{\textbf{Wrong Pair Disappears}: The player's goal is to hold a black cube, which does not exist among the initial objects. But there are two (hidden) production rules. The player needs to identify the correct world state which triggers the rule that creates the cube and not the one which destroys the necessary inputs. All this is embedded in a challenging world layout with one-way drops and limited visibility.} \label{fig:wrong-pair-disappears-explained}
  \end{minipage}
\end{figure}

\begin{figure}
  \begin{minipage}[c]{0.65\textwidth}
        \includegraphics[width=1\textwidth]{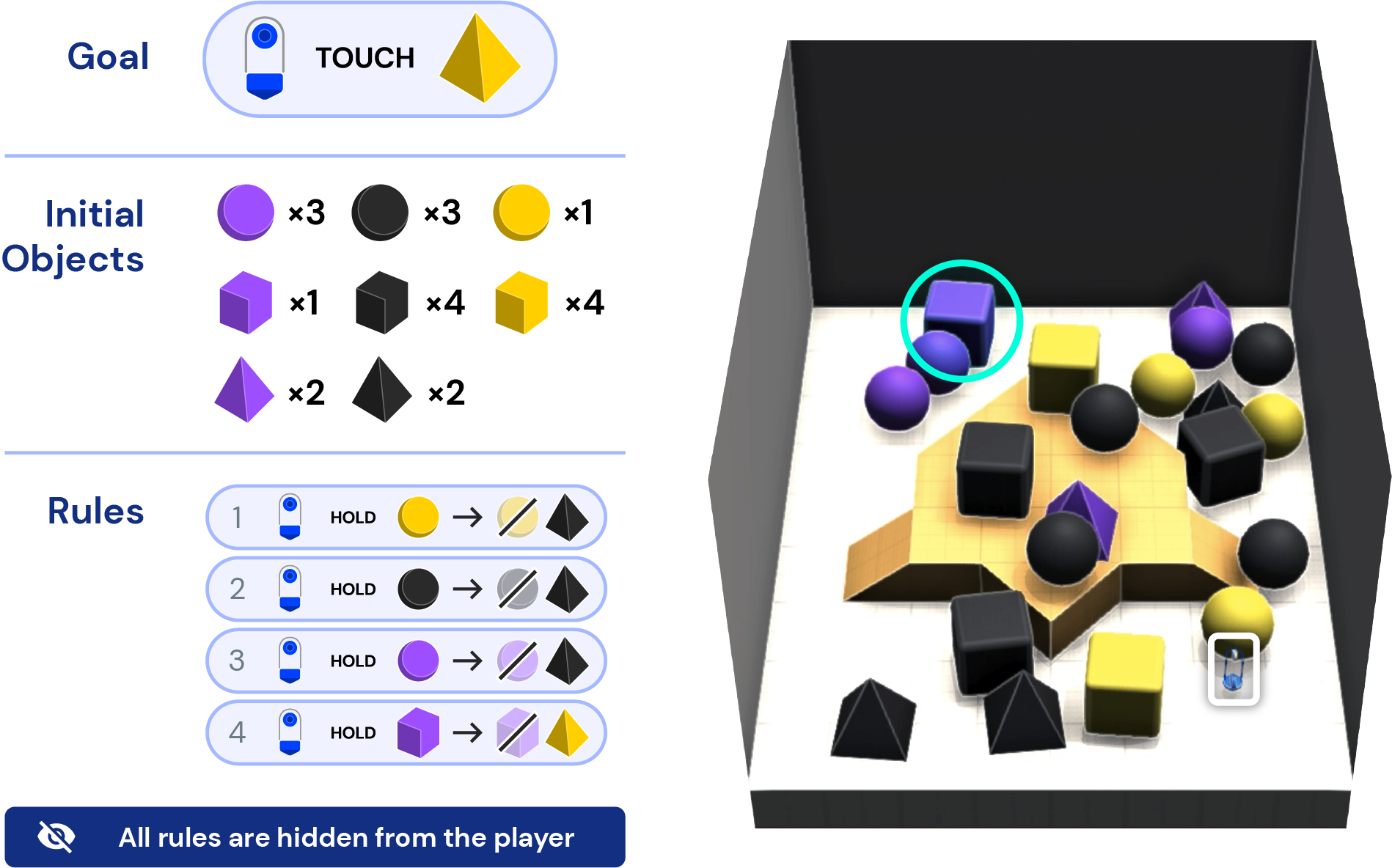}
  \end{minipage}\hfill
  \begin{minipage}[c]{0.3\textwidth}
    \caption{\textbf{Pyramid in a Haystack}: To create the necessary yellow pyramid, the player needs to find and hold the purple cube. There are several distractor objects and distractor rules in this world, requiring the player to deal not just with a hard exploration challenge but also a very noisy environment.} \label{fig:pyramid-in-a-haystack-explained}
  \end{minipage}
\end{figure}

\begin{figure}
  \begin{minipage}[c]{0.65\textwidth}
        \includegraphics[width=1\textwidth]{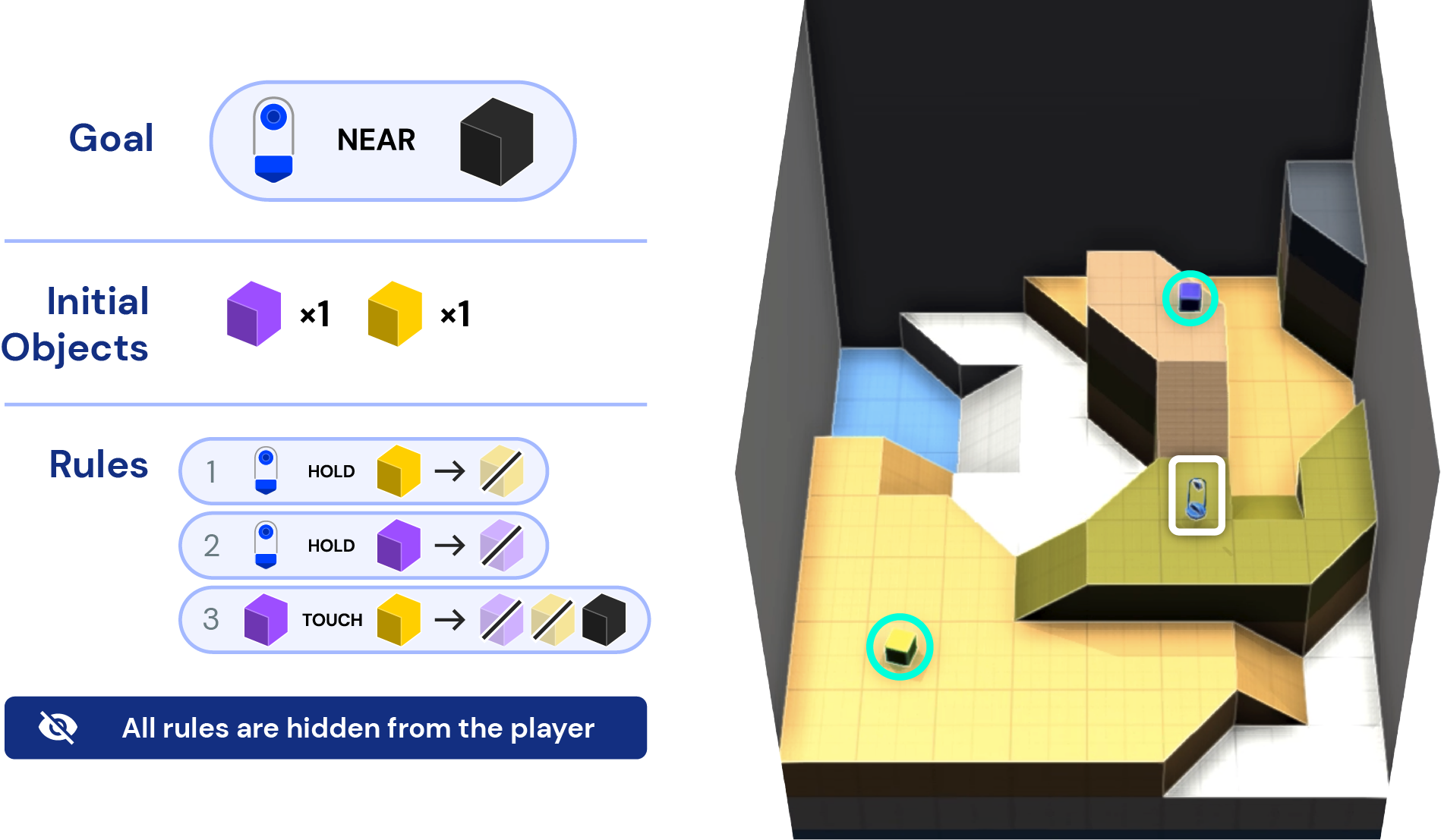}
  \end{minipage}\hfill
  \begin{minipage}[c]{0.3\textwidth}
    \caption{\textbf{Push, don't lift}: The vast majority of training and evaluation tasks require lifting objects. Here two hidden rules destroy any object when lifted. In order to create the goal state, some ``lateral thinking'' is necessary: the player needs to identify that pushing the cubes with their body is possible.} \label{fig:no-touchy-explained}
  \end{minipage}
\end{figure}

Evaluation using test tasks can only give us information with respect to the pre-sampled distribution in Appendix \ref{app:presample_tasks}. While this is certainly vast and diverse, an arbitrary task sampled from the test set is not necessarily easily understandable for humans. So in addition to quantitative evaluation on $1000$ test tasks we also investigate specific human-level capabilities of our agent on two sets of $30$ single-agent and $28$ multi-agent probe tasks. These are based on situations that are intuitive to humans and which require qualitatively different behaviours, which we can inspect in more detail ``with the naked eye''. A full description of all $58$ probe tasks can be found in Appendix \ref{appendix:human_scale_adaptation}. Representative single-agent and multi-agent probe tasks are described in detail in Figures \labelcref{fig:wrong-pair-disappears-explained,fig:pyramid-in-a-haystack-explained,fig:no-touchy-explained}.

\subsection{Adaptation metric} 

We introduce an \textit{adaptation metric} to rank our agents based on their few-shot adaptation capability across the hand-authored probe task set. We collect last-trial total reward for all $(m, k)$ pairs where $m$ is a probe task and $k \in \{1, 2, 3, 5, 8, 13\}$. We normalise the per-task scores just as for the test set above. We then aggregate over tasks using the $50$\textsuperscript{th} percentile (median). Finally, we aggregate over $k$ by saying that agent $A$ ranks higher than agent $B$ if and only if $A$'s task-aggregated scores are a Pareto improvement over $B$'s. That is to say, we would like agents that are both capable of high-quality zero-shot generalisation where possible ($k = 1$), and also that can use additional trial information to efficiently improve their policy in few-shots $k > 1$; we don't ``trade-off'' between these. 

A convenient way of using the Pareto improvement criterion to compute a scalar metric is the Nash average method of \citet{balduzzi2018re}. We construct a competitive meta-game of ``agents vs. $k$'', and compute the maximum entropy Nash equilibrium for the game. The Nash payoff is then used as the adaptation metric, and agents are ranked by this metric. As desired, this metric has the property that if neither agent $A$ nor agent $B$ Pareto-dominate the other, the $A$ and $B$ receive the same Nash payoff and are therefore ranked equally. This adaptation metric was used as the means of selecting hyperparameters for training our best performing agent in Section \ref{sec:results_human_scale_adaptation}.

\begin{figure}
  \begin{minipage}[c]{0.65\textwidth}
        \includegraphics[width=1\textwidth]{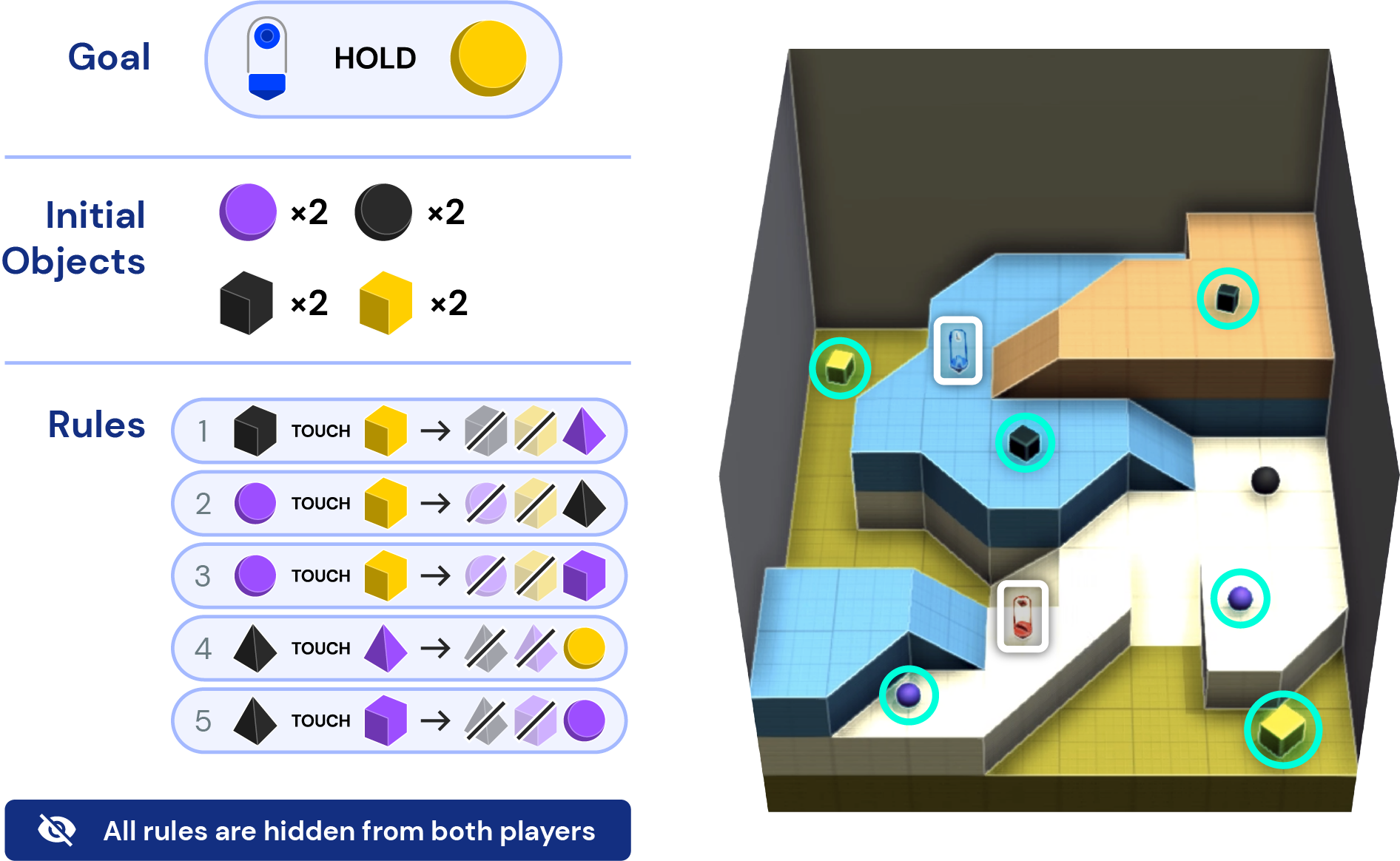}
  \end{minipage}\hfill
  \begin{minipage}[c]{0.3\textwidth}
    \caption{\textbf{Irreversible production for two}: Both players score when the first player holds a yellow sphere. There is no yellow sphere initially, but it can be produced from executing the first, second and fourth production rules in order. The other two rules are dead ends, destroying key input objects. Note that some input objects exist multiple times in the initial state, so there are multiple solution paths.} \label{fig:ma-irreversible-production-explained}
  \end{minipage}
\end{figure}

\subsection{Human data collection}\label{app:human-data-collection}

\begin{figure}
    \centering
    \begin{subfigure}[b]{0.45\textwidth}
        \centering
        \includegraphics[width=\linewidth]{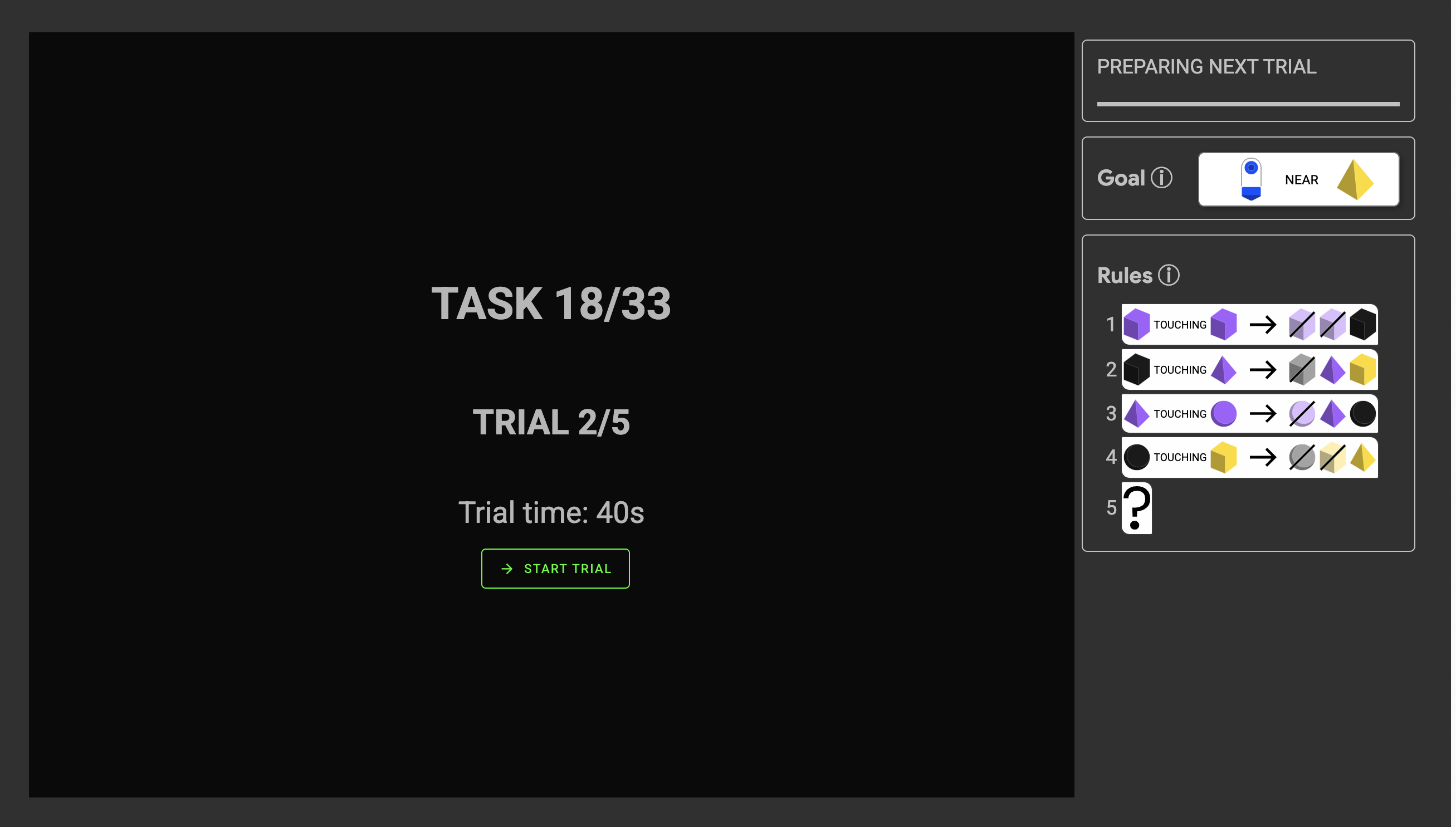}
        \caption{}
    \end{subfigure}
    \begin{subfigure}[b]{0.46\textwidth}
        \centering
        \includegraphics[width=\linewidth]{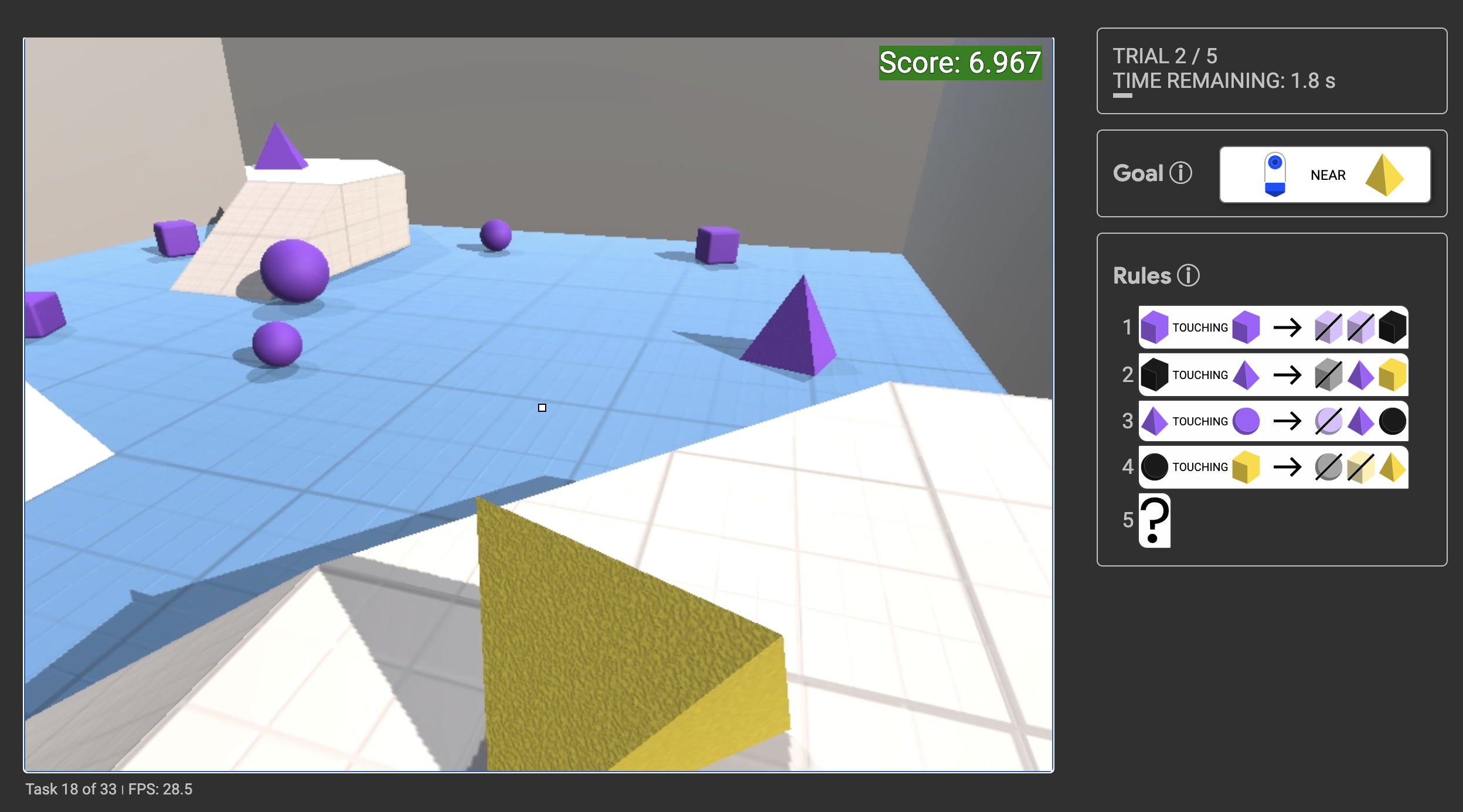}
        \caption{}
    \end{subfigure}
\caption{The human player interface. \textbf{(a)} Prior to each trial players are told how many trials they have remaining on the current task, and are given unlimited time to read the goal and production rules (subject to any hiding). \textbf{(b)} During the trial players observe the same first-person camera view as agents, but at a higher $800 \times 600$ pixel resolution. The goal, production rules, current score, and time remaining in the trial are displayed via UI elements.}
    \label{fig:human_ui}
\end{figure}

To provide a benchmark for human-timescale adaptation, we collected score data from a pool of $100$ human players on the $30$ single-agent probe tasks. Before attempting the probe tasks, each player completed a graded training curriculum of $23$ tasks to acquire familiarity with the mouse-and-keyboard control scheme and user interface (Figure \ref{fig:human_ui}), and the particular game mechanics of XLand. Since humans cannot undergo a short-term memory ``reset'' on episode boundaries, individual players attempted each of the probe tasks for a single $k$ only, with $k \in \{1, 2, 3, 5, 8\}$. All players experienced a variety of $k$ across the task set. We assigned each player a unique ordering of the $30$ probe tasks to average out any knowledge transfer from earlier tasks to later, and, within those orderings, we imposed separation between tasks with known similarity. Technical problems (e.g. internet dropout) prevented completion of 3.4\% of the $3000$ episodes, leaving an average of 19.3 samples per (task, $k$) pair, and a minimum of 17 samples for any individual (task, $k$).

\section{Agent Details}
Here we provide technical details of our agent architecture, observation and action specifications. 

\subsection{Agent Architecture}
\label{app:agent-architecture}
Here we provide an overview of the agent architecture, and provide important hyperparameters for the agent, noting that we performed minimal hyperparameter tuning in this work.

\paragraph{Observation encoder.}
The first-person view RGB observation (Table \ref{tab:observations}) is passed through a ResNet \citep{he2016resnet} with $[16, 32, 32]$ convolutional channels, each consisting of $2\times2$ blocks and a final output size of $256$. Max-pooling is used, in addition to scalar residual multipliers. \texttt{relu} activations are used throughout.

The goal observation is passed through a goal embedder, which is the same in \citet{xland}. This maps each of the 6 goal elements (negation, predicate, shape of object 1, colour of object 1, shape of object 2, colour of object 2) in the goal representation to a dense embedding vector of size 8. These are concatenated together and passed through a 3-layer MLP of size $[8, 8, 8]$, resulting in a final goal embedding of size $8$.  

Production rules are encoded in the same manner as the goal, but mapped through a larger final MLP of shape $[512, 256]$ resulting in a final production rule embedding vector of size $256$.

The encoded RGB, goal, and production rules observations are concatenated together with all remaining scalar and vector observations (including previous reward, previous action, proprioception observations, and trial timing information) and passed though a final MLP. This results in an encoded observation vector of a size matching the hidden dimension of the Transformer memory. 

\paragraph{Transformer memory.} We use a Transformer-XL with causal masking \citep{dai2019transformer}. For the actor step we use a context window of 1, and in the learner step we use a rollout context window of 80. The Transformer-XL memory uses 300 previous cached activations (1800 effective timesteps) of memory in all experiments unless otherwise stated. We apply layer normalisation before attention operations as in \citet{parisotto2020stabilizing}, use gating in the feedforward component as in \citet{shazeer2020glu}, and apply relative positional embeddings as in \citet{dai2019transformer}. As is common with Transformers, we use the \texttt{gelu} activation function throughout \citep{hendrycks2016GaussianEL}.

\paragraph{Muesli sequence model and prediction heads.} 
Next, we take the Transformer-XL output embedding, and use two MLPs of width $1000$ to produce the hidden and cell values for the initial state of the Muesli model LSTM. On top of the hidden value, we apply MLP heads of width $1000$ for the policy and value respectively. The policy MLP is followed by 6 linear softmaxed outputs corresponding to 6 action groups in a decomposed action space for the policy (as in \citet{xland}). The value MLP is followed by a 601-unit binned logit prediction as in \citet{hessel2021muesli}. Finally, the Muesli sequence model is unrolled for a further 4 steps, starting from the LSTM state embedding and producing a 1000-dimensional output vector on each LSTM step. This feeds into a further 1000-dimensional MLP followed by a 601-unit binned reward prediction. 
\subsection{Observations}
\label{app:observations}

We summarise all the observations received by the agent when running inference in Table \ref{tab:observations}. In the descriptions, ``legacy reasons'' refers to an observation format that was inherited from \citet{xland}.

\begin{longtable}{>{\raggedright}p{0.25\linewidth}|p{0.1\linewidth}|p{0.52\linewidth}}
    \caption{Agent observations.}
    \label{tab:observations}
    \\ \hline
      \textbf{Observation name} & \textbf{Shape} & \textbf{Meaning} \\
      \hline
      RGB & $72 \times 96 \times 3$ & RGB values of the first-person view of the agent. \\
      \hline
      IS HOLDING & $1$& Integer in $\{0, 1\}$ indicating whether the agent is holding an object.\\
      \hline
      HAND DISTANCE & $1$& Distance to the held object as a fraction of the agent's maximum reach (0 while no object is held).\\
      \hline
      HAND FORCE & $1$ & The force exerted by the agent on the held object as a fraction of its maximum grip force (0 while no object is held).\\
      \hline
      LAST ACTION & $10$ & Last action performed by the agent.\\
      \hline
      GOAL ATOMS & $6 \times 6$ & Agent-readable description of the goal. For legacy reasons, only the first row is non-zero. The first row (6 numbers) describes the goal with elements:\\&&1: whether the goal is negated or not,\\&& 2: index of the binary predicate,\\&&3-4: shape and colour of the first objects,\\&&5-6: shape and colour of the second object. \\
      \hline
      GOAL SOP MATRIX & $6 \times 6$ & Always the same, kept for legacy reasons.\\
      \hline
      ATOM REWARDS & $6$ & The reward of the agent in the previous frame (1 number), padded with zeros for legacy reasons.\\
      \hline
      OPTION REWARDS & $6$ & Same as ATOM REWARDS, kept for legacy reasons.\\
    \hline
      PRODUCTION RULES & $16 \times 26$ & A description of up to 16 production rules. A single production rule is described as: $3 \times 6 = 18$ numbers describing up to three object-predicate-object triggers and $4 \times 2 = 8$ numbers describing up to $4$ spawn objects. We only ever use a single object-predicate-object trigger for all tasks in this paper. Hiding (Section \ref{sec:methods_xverse}) is implemented by adding extra predicate and shape indices meaning hidden production rule, first hidden object, etc. \\
      \hline
      REWARD &$1$& The environment reward obtained in the previous step.\\
      \hline
      TRIALS REMAINING &$5$& One-hot encoding of the number of trials remaining in the episode, up to a maximum of 5 trials remaining. \\
      \hline
      MORE THAN 5 TRIALS &$1$& Integer in $\{0,1\}$ indicating whether there are more than 5 trials remaining in the episode. \\
      \hline
      TIME UNTIL LAST TRIAL &$1$& Time remaining (in seconds) until the final trial in the current episode. \\
      \hline
      TIME LEFT IN CURRENT TRIAL &$1$& Time remaining (in seconds) until the end of the current trial. \\
      \hline
      DURATION OF LAST TRIAL &$1$& The duration (in seconds) of the final trial in this episode. In our setting, all trials for the same task have the same duration.  \\
      \hline
      DURATION OF NEXT TRIAL &$1$& The duration (in seconds) of the next trial in the current episode. In our setting, all trials for the same task have the same duration. \\
      \hline
\end{longtable}

\section{Training Details}
\label{app:training-details}
\subsection{Meta-RL}\label{appendix:metarl}

AdA is trained with a meta-RL setup, in which episodes of experience for the agent comprise multiple trials of interaction with the task environment, where the task is reset on trial boundaries. In this setting, it is known that the agent's policy can converge to Bayes-optimal behaviour, experimenting on-the-fly to reduce its epistemic uncertainty, and reusing discovered information to achieve goals increasingly efficiently. It is important that the agent's memory is not reset at trial boundaries, but only at episode boundaries. Similarly, the agent trains with a fixed discount factor $\gamma \neq 0$ throughout the episode, including on trial boundaries. For training, we sample $(m, k)$ pairs according to a factorised distribution $(\rho_{\mathcal{M}}, \rho_K)$, the parameters of which are controlled by an automatic curriculum (Section \ref{sec:methods_auto_curriculum}). The MDPs $m$ are all drawn from a procedurally generated domain $\mathcal{M}$, called XLand (Section \ref{sec:methods_xverse}). We choose as our space of trials $K = \{1,2, \dots 6\}$, and provide our agent with $k$ as a conditioning input. After training, our agent is capable of few-shot adaptation across a wide-range of MDPs, including in held-out $m \in \mathcal{M}$ on which $\rho_{\mathcal{M}}$ puts no probability mass, and when $k > 6$.

\subsection{Single-agent training}
\label{app:single_agent_training}

Our single-agent training setup uses a task pool generated as described in Section~\ref{sec:methods_xverse}. The experimental setup for the single-agent distillation teacher is summarised in Table \ref{tab:single-agent-teacher}. Single-agent training used an earlier version of XLand 2.0 than multi-agent experiments, without the frozen objects described in Section \ref{app:xland_changes}. Frozen objects were also therefore excluded from the test and hand-authored probe task sets. AdA was implemented using JAX \citep{jax2018github} and the DeepMind JAX Ecosystem \citep{deepmind2020jax} and trained on 64 Google TPUv3 devices. The wall-clock time for training this version of AdA from scratch was approximately 5 weeks: 1 week to train the teacher, and 4 weeks to train AdA. Even after this amount of training, AdA had not reached convergence, illustrating the benefits of open-ended learning methods.

\begin{table}[h!]
  \begin{center}
    \caption{Distillation teacher for the single-agent experiments in Section \ref{sec:results_human_scale_adaptation}.}
    \begin{tabular}{c|c|c|c|c|c}
    \label{tab:single-agent-teacher}
      Model Parameters & Memory & Task pool & Curriculum & Teacher & Steps\\
      \hline
      23M TXL / 76M total & 1800 & 25B & PLR \ref{tab:plr_hyperparameters} & None & 25B \\
    \end{tabular}
  \end{center}
\end{table}

\subsection{Multi-agent training}
\label{app:multi_agent_training}

Starting from the 25B-sized task pool used for single-agent training, we generate a two-player task pool of the same size by the procedure described in Appendix~\ref{app:presample_tasks}. For all our multi-agent experiments, we use a half-half mixture of single-player and two-player tasks, as described in Section~\ref{sec:methods_xverse}. For each task we decide whether to spawn some of the initial objects permanently frozen (see Appendix~\ref{app:xland_changes}) with $50\%$ probability. For tasks with frozen objects, we iterate over the initially spawned object types and freeze all spawned instances of this type with a probability of $20\%$, while ensuring that the task remains solvable.

During training we uniformly sample a co-player policy from a pool generated using fictitious self-play. The co-player pool is initialised with a random-action policy. Every 500M training frames we add a snapshot of the learning agent to the pool, thereby adding more and more capable co-players over time. Finally we apply the PLR auto-curriculum method (see Section~\ref{sec:methods_auto_curriculum}) to curate the tasks (worlds, games and co-players) using the agent's TD-error based fitness. The experimental setup for the multi-agent distillation teacher is summarised in Table \ref{tab:multi-agent-teacher}.

\begin{table}[h!]
  \begin{center}
    \caption{Distillation teacher for the multi-agent experiments in Section \ref{sec:results_human_scale_adaptation}.}
    \begin{tabular}{c|c|c|c|c|c}
    \label{tab:multi-agent-teacher}
      Model Parameters & Memory & Task pool & Curriculum & Teacher & Steps\\
      \hline
      23M TXL / 76M total & 1800 & see Sec~\ref{app:multi_agent_training} & PLR \ref{tab:plr_hyperparameters} & None & 22B \\
    \end{tabular}
  \end{center}
\end{table}

\subsection{Architecture experiments}
\label{app:architecture-experiments}
Table \ref{tab:arch_exp} shows the experimental setup for the experiments comparing different memory architectures in Section \ref{sec:results_architecture}.
\begin{table}[h!]
  \begin{center}
    \caption{Experimental setup for comparing different memory architectures.}
    \begin{tabular}{c|c|c|c|c|c|c}
    \label{tab:arch_exp}
      Architecture & Parameters & Memory & Task pool & Curriculum & Teacher & Steps\\
      \hline
      Transformer-XL & \multirow{3}{*}{76M total} & 1800 & \multirow{3}{*}{25B} &\multirow{3}{*}{No-op}  & \multirow{3}{*}{None} & \multirow{3}{*}{50B} \\
      GRU with Attention & & - & & & & \\ 
      GRU & & - & & & & \\ 
    \end{tabular}
  \end{center}
\end{table}

\subsection{Auto-curriculum learning}
\label{app:autocurriculum}

\paragraph{No-op filtering details.} Here we provide additional details of No-op filtering. For each task from the XLand training pool, we evaluate the learning agent (without sending any experience to the learner) and no-op policy on the task for 10 independent episodes, each of length 1 trial,
producing scores $\{R_0, \dots, R_9\}$, $\{R'_0, \dots, R'_9\}$, respectively. We admit the proposal task for training if it satisfies the following criteria:

\begin{enumerate}
    \item $\max R'_i \leq \epsilon_1$ (No-op is not too good.)
    \item $|\{i : R_i \geq \epsilon_2 \}| \leq \epsilon_3$ (Agent is not too good.)
    \item $|\{i : R_i \geq \max R'_i + \epsilon_0 \}| \geq \epsilon_4$ or $|\{i : R_i \leq \min R'_i - \epsilon_0 \}| \geq \epsilon_5$ (Agent is sufficiently different from no-op.)
    \item $\max R_i - \min R_i \geq \epsilon_6$ (Agent scores have sufficient variance.)
\end{enumerate}
The $\epsilon_i$'s are thresholds and become hyperparameters in our training setup. Since different tasks have different durations in general, we use relative thresholds defined as a fraction of trial duration for $\epsilon_0$, $\epsilon_1$, $\epsilon_2$, $\epsilon_6$ and absolute thresholds for the rest. Once a task is admitted for training, it is run using the full number of trials specified by the task and for 30 episodes. All experience from these runs are sent to the learner. See Table~\ref{tab:noop_hyperparameters} for the hyperparameters used.

\begin{table}[h!]
    \centering
    \caption{No-op filtering hyperparameters.}
    \begin{tabular}{c|c|c}
         Parameter & Value & Relative to trial duration \\ \hline
         $\epsilon_0$ & 0.01 & Y \\
         $\epsilon_1$ & 1.1 & Y \\
         $\epsilon_2$ & 0.4 & Y \\
         $\epsilon_3$ & 5 & N \\
         $\epsilon_4$ & 1 & N \\
         $\epsilon_5$ & 3 & N\\
         $\epsilon_6$ & 0.01 & Y 
    \end{tabular}
    \label{tab:noop_hyperparameters}
\end{table}

\paragraph{PLR details.} 
Here we provide additional details for Prioritised Level Replay (PLR), used in training AdA. PLR uses a \emph{fitness score} that approximates the agent regret for a given task~\citep{jiang2021robustplr, jiang2020prioritized}. PLR maintains an archive $\mathcal{P}$ of tasks to replay with fixed maximum size. With probability $p$ (referred to as the \emph{replay probability}) a task is sampled from $\mathcal{P}$ while taking into account the fitness score and staleness of each task (see~\citet{jiang2020prioritized}, Section~3) to train the learning agent. The staleness represents how much time has passed since the task was last sampled, and ensures that all tasks in $\mathcal{P}$ have accurate scores. The final probabilities are computed by combining the fitness and staleness scores, with staleness weighted using the parameter $s\in[0,1]$.

Tasks are added to $\mathcal{P}$ by first sampling with probability $1-p$ a proposal task from the training task set and evaluating its fitness score. If the fitness score is greater than the minimum fitness score of all tasks in $\mathcal{P}$, the proposal task is added to $\mathcal{P}$. If the new size of $\mathcal{P}$ exceeds the fixed maximum size, the lowest fitness task is dropped. Note that in PLR a task can potentially be trained on indefinitely, if it never leaves $\mathcal{P}$. 

We found that using last-trial fitness led to better empirical performance and sample efficiency than first or average trial fitness. This is likely because in earlier trials, error-based fitness is higher as the agent is pursuing exploratory behavior, which should not be taken as a sign that the policy is sub-optimal. However, high error-based fitness in the last trial likely indicates a sub-optimal policy when solving the task after time for adaptation, analogous to the regret approximation in the original single-trial PLR. 

In order to use the last-trial fitness as our objective we need to make a number of changes to the original PLR framework, which was designed for single trials in a more homogeneous domain. We denote the per-step fitness score at the $i$\textsuperscript{th} step of trial $k$ by $f_{i, k}$. First, to avoid adding a bias towards longer trial durations we use the \emph{average} per-trial fitness score $\tilde{f}_k \triangleq \sum_i \tilde{f}_{i, k} / N_k$ where $N_k$ is the number of steps per trial. Next, to ensure we do not prioritise lower values of $k$, which tend to have a higher average last-trial fitness score, we then normalise $\tilde{f}_k$ by $f_k \triangleq (\tilde{f}_k - \mu_k) / \sigma_k$ where $\mu_k$ and $\sigma_k$ are rolling per-trial means and variances for each trial index, calculated from all evaluated tasks. Finally, we can define the fitness for PLR to be $f_k$, the normalised last-trial fitness score.

As described in~\citet{jiang2021robustplr, jiang2020prioritized}, PLR contains the following hyper-parameters: replay probability $p$, maximum replay size $N_{\textrm{max}}$, minimum replay size $N_{\textrm{min}}$ (we set $p = 0$ if $|\mathcal{P}| < N_{\textrm{min}}$), $N_{\textrm{train}}$ the total number of trials for which to run a training task before re-sampling, and $s$ the staleness coefficient. See Table~\ref{tab:plr_hyperparameters} for the hyper-parameters used. We conducted a grid search over the replay probability $p\in\{0.1,0.2,0.5\}$, size of the replay pool $N_{\textrm{max}} \in \{1000, 10000, 50000\}$ and staleness coefficient $s\in\{0.1,0.2\}$, and in all cases set $N_{\textrm{min}}$ to be $90\%$ of $N_{\textrm{max}}$. 

\begin{table}[h!]
    \centering
    \caption{PLR hyperparameters.}
    \begin{tabular}{c|c}
         Parameter & Value \\ \hline
         $p$ & 0.2 \\
         $N_{\textrm{max}}$ & 1000 \\
         $N_{\textrm{min}}$ & 900 \\
         $N_{\textrm{train}}$ & 30 \\
         $s$ & 0.2
    \end{tabular}
    \label{tab:plr_hyperparameters}
\end{table}

\paragraph{PLR fitness metric.}
It remains for us to define the per-step fitness score $f_{i,k}$. For this, we use the simplest regret-like metric, the 1-step TD-error \citep{jiang2020prioritized, schaul2015prioritized}. Concretely, we estimate the \emph{TD-error fitness} based on the immediate value-predictions of the model: $|r_{t} + \gamma \hat{v}^{t+1}_0 - \hat{v}^t_0|$. In some settings this may be undesirable, for example, TD-errors typically increase as the agent achieves higher rewards. Therefore, we also propose to compute fitness metrics based on the Muesli dynamics model. Rather than simply using the accuracy of the model prediction, we look at the impact of the prediction on the value function and action predictions. We define the \emph{value-model fitness} as $|\hat{v}^{t+1}_0 - \hat{v}^t_1|$, the difference between the value estimate at the predicted next state and the true next state. We also define a value-agnostic metric, the \emph{action-model fitness} as follows: $JS(\hat{\pi}^{t+1}_0, \hat{\pi}^t_1)$, i.e. the difference between the action predictions at the predicted next state and the actual next state, where difference is measured with the Jensen-Shannon divergence \citep{NIPS2017_fb60d411, NEURIPS2021_08f0efeb, FilosVMFBFBS0O22, pislar2022when}.

\begin{figure}[b!]
    \centering
    \includegraphics[width=0.6\linewidth]{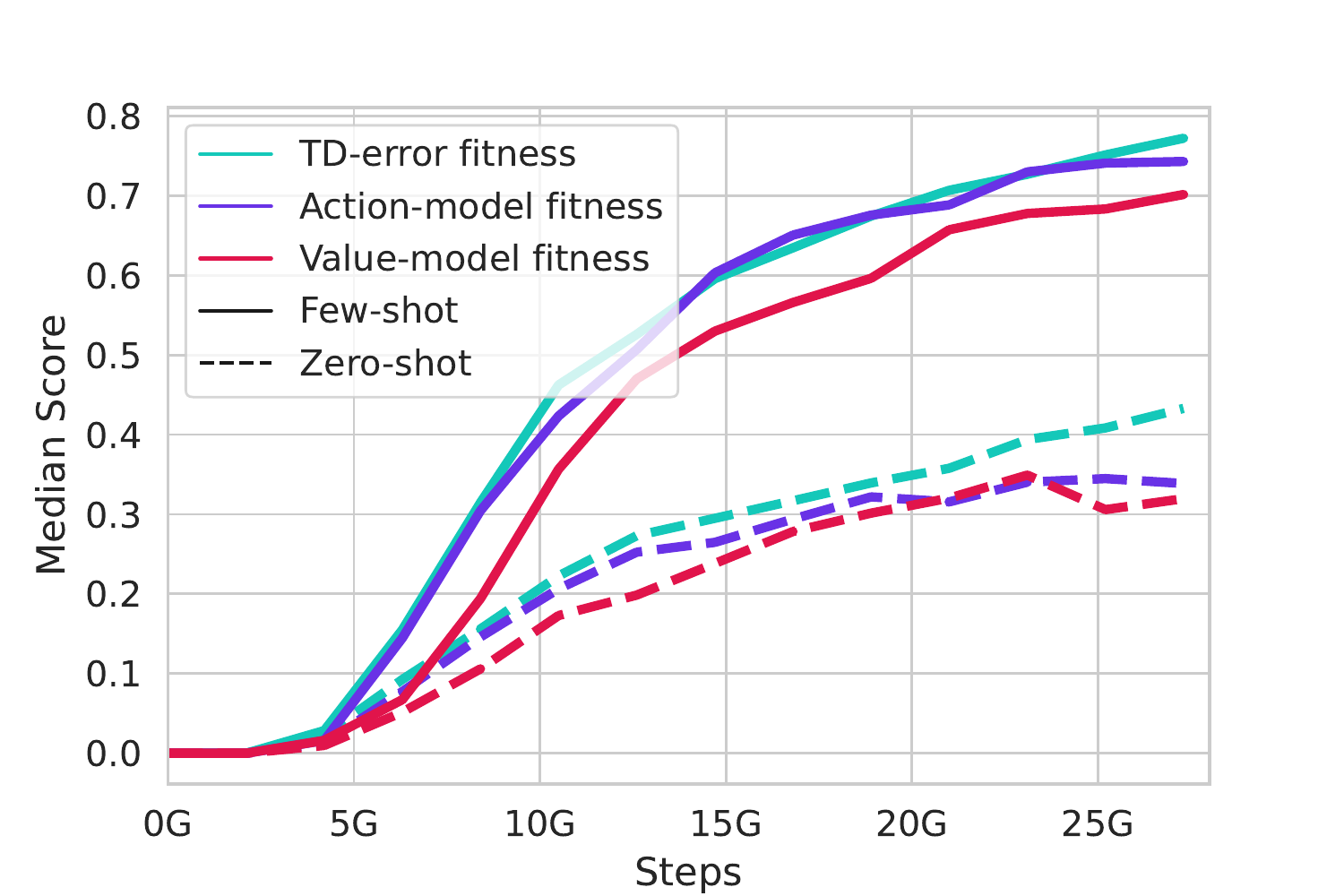}
    \caption{
         PLR fitness metric comparison for zero-shot generalisation ($k = 1$) and few-shot adaptation ($k = 13$). We compare the TD-error fitness used in our main agents against two approaches using the Muesli dynamics model. We see that action-model fitness matches TD-error fitness in few-shot performance, with weaker zero-shot performance.
    }
    \label{fig:autocurriculum_abl1}
\end{figure}

In Figure~\ref{fig:autocurriculum_abl1} we show training curves for TD-error fitness, value-model fitness, and action-model fitness. Table \ref{tab:fitness_functions} shows the experimental setup for these experiments. We see that both TD-error and action-model fitness metrics outperform the value-model fitness. We chose TD-error for our PLR training runs because it has better asymptotic performance in both the zero-shot and the few-shot setting, and because it was shown to perform well in previous work~\citep{jiang2021robustplr}.

\begin{table}[h!]
  \begin{center}
    \caption{Experimental setup for comparing different PLR fitness functions.}
    \begin{tabular}{c|c|c|c|c|c|c}
    \label{tab:fitness_functions}
      Model parameters & Memory & Task pool & Curriculum & Teacher & Steps & Fitness function \\
      \hline
      \multirow{3}{*}{23M TXL / 76M total} & 
      \multirow{3}{*}{1800} & 
      \multirow{3}{*}{25B} &
      \multirow{3}{*}{PLR} & 
      \multirow{3}{*}{None} & 
      \multirow{3}{*}{25B} & 
      TD error \\
       & & & & & & Value model \\ 
       & & & & & & Action model \\ 
    \end{tabular}
  \end{center}
\end{table}

\paragraph{Curriculum efficiency.}

\begin{figure}[htb]
    \centering
    \begin{subfigure}[b]{0.49\textwidth}
        \centering
        \includegraphics[width=\linewidth]{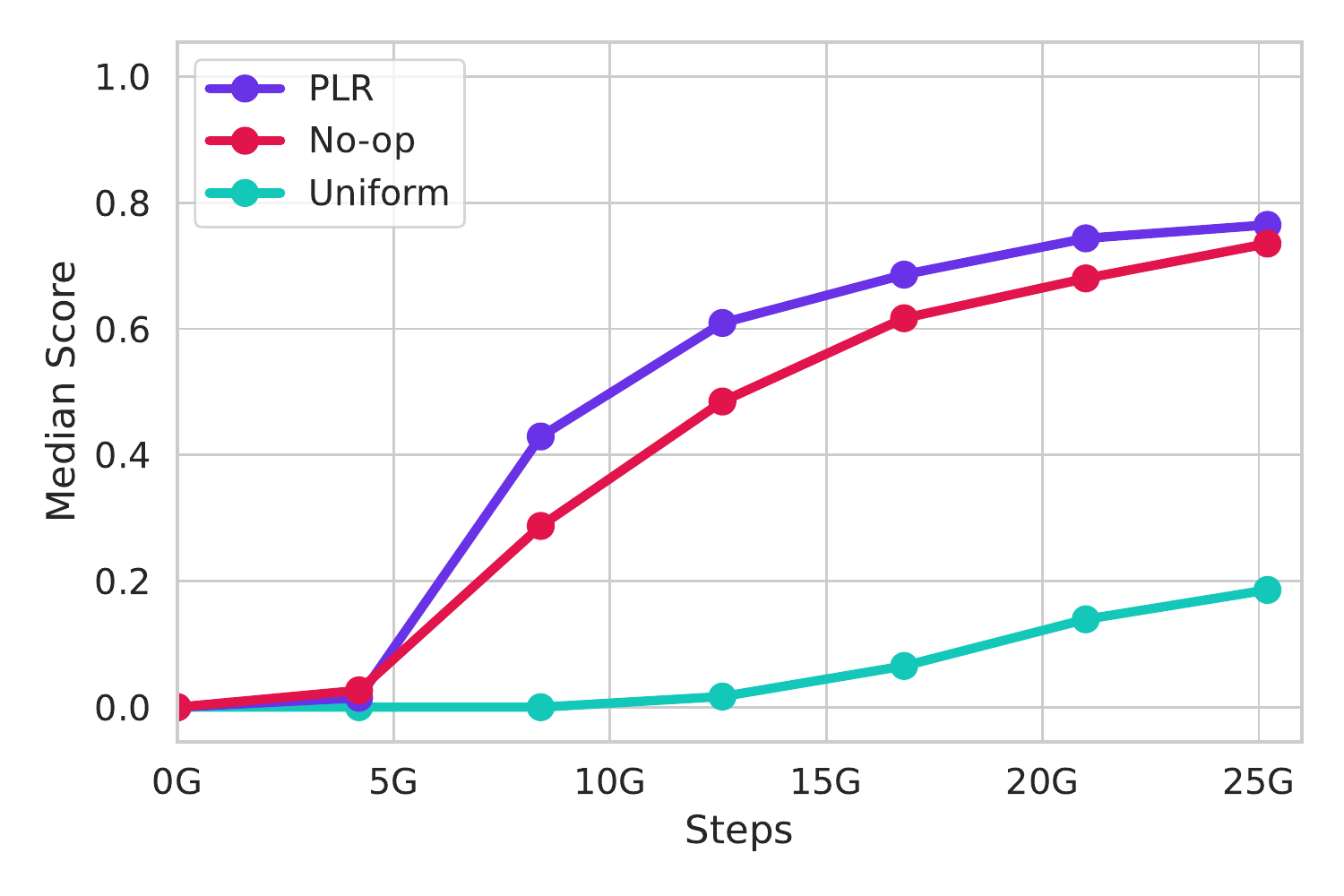}
        \vskip -3pt \caption{}
    \end{subfigure}
    \begin{subfigure}[b]{0.49\textwidth}
        \centering
        \includegraphics[width=\linewidth]{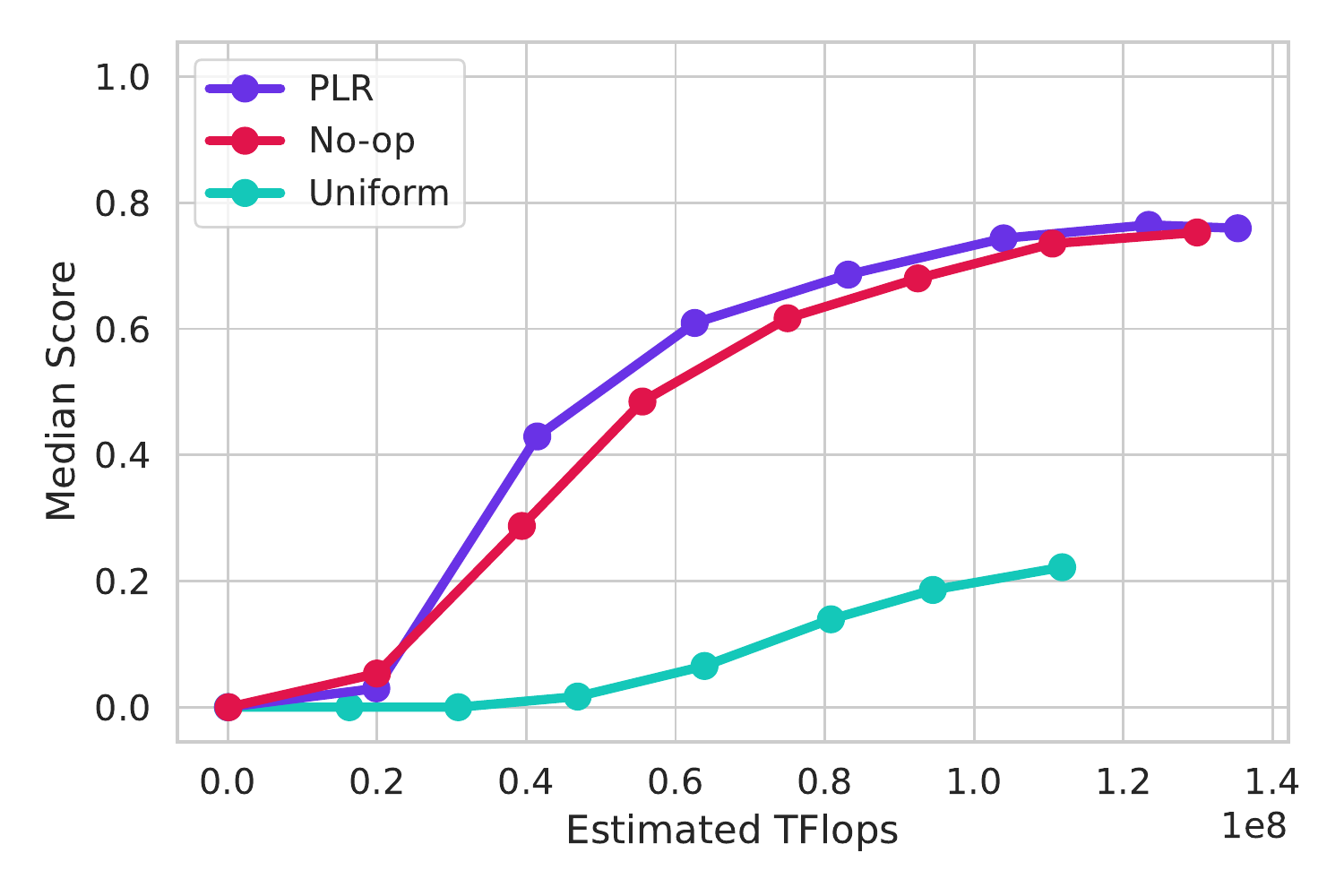}
        \vskip -3pt \caption{}
    \end{subfigure}
    \caption{
        \textbf{(a)} Sample efficiency in steps for different choices of curricula. Both No-op and PLR significantly improves sample efficiency over uniform sampling of tasks. Few-shot denotes $k=13$ score and zero-shot denotes $k=1$ score. 
        \textbf{(b)} Sample efficiency in FLOPs for different choices of curricula. No-op has an initial advantage over PLR, but PLR outperforms No-op later in training.
    }
\label{fig:curricula_efficiency}
\end{figure}

Next, we compare training sample efficiency for baseline uniform sampling and the different curriculum methods, in units of both learner steps and FLOPS. Figure~\ref{fig:curricula_efficiency} shows the median last-trial scores for few-shot ($k=13$) and zero-shot evaluation tasks as a function of learner steps. We see that both No-op filtering and PLR curricula strongly improve training sample efficiency over uniform sampling, with PLR being more efficient early in the training process. When we plot the same few-shot median performance as a function of FLOPS, we see that No-op has a slight early advantage, but PLR outperforms No-op later in training. The initial advantage for No-op may be because No-op expends more FLOPS (10 evaluations per task vs.~1 in PLR) for task evaluation, which finds higher quality training tasks at the start of training.

\begin{figure}[h!]
    \centering
    \includegraphics[width=0.99\linewidth]{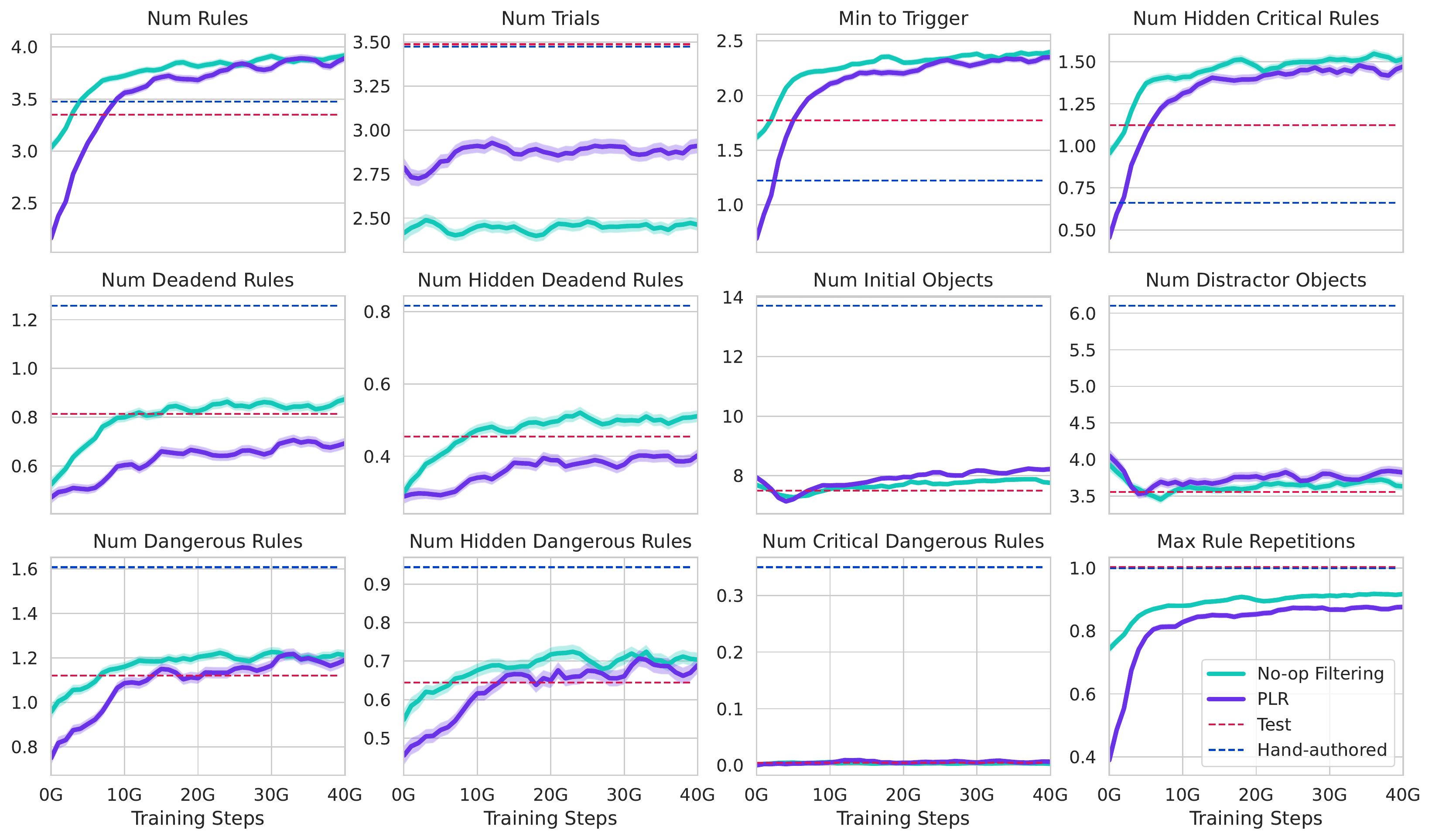}
    \caption{
        Emergent curricula for No-op filtering and PLR. Plots show the full set of task metrics for the dynamic training set, averaged over all tasks in the set, with standard error shaded. In all plots, a higher metric value corresponds to greater task difficulty. Horizontal lines show the same metric values averaged over the test (dashed) and hand-authored (dotted) evaluation task sets.}
    \label{fig:emergent_curricula_extended}
\end{figure}

\paragraph{Emergent curricula.} In Figure~\ref{fig:emergent_curricula_extended} we show task metrics analysing the tasks selected by PLR and No-op filtering, expanding upon the results shown in Figure \ref{fig:emergent_curricula}.

\subsection{Distillation teacher for scaling experiments}
\label{app:scaling_distillation_teacher}
In all of our scaling experiments (Sections \ref{sec:scaling_results} and \ref{sec:results_task_scaling}), we distill the policy from an identical teacher snapshot to ensure our experiments are comparable. Training details for the teacher are detailed in Table \ref{tab:Teacher}. This teacher is used to kickstart our agents for their first 4B frames of training. 

\begin{table}[h!]
  \begin{center}
    \caption{Distillation teacher for scaling experiments.}
    \begin{tabular}{c|c|c|c|c|c}
    \label{tab:Teacher}
      Model parameters & Memory & Task pool & Curriculum & Teacher & Steps\\ 
      \hline
      23M TXL / 76M total & 1800 & 200M & No-op & None & 23B \\ 
    \end{tabular}
  \end{center}
\end{table}

\subsection{Scaling the network}
\label{app:network-scaling}
Table \ref{tab:model_scaling} shows the experimental setup for the model size scaling experiments in Section \ref{sec:scaling_results}. More details about the number of parameters for the various model sizes can be found in Table~\ref{tab:model_hparams}.

\begin{table}[t!]
  \begin{center}
    \caption{Experimental setup for model size scaling.}
    \begin{tabular}{c|c|c|c|c|c}
    \label{tab:model_scaling}
      Model parameters & Memory & Task pool & Curriculum  & Teacher & Steps \\ 
      \hline
      6M TXL / 41M total & \multirow{7}{*}{1800} & \multirow{7}{*}{25B} & \multirow{7}{*}{No-op} & \multirow{7}{*}{Table \ref{tab:Teacher}} & \multirow{7}{*}{75B}\\ 
      23M TXL / 76M total&&&&&\\ 
      42M TXL / 112M total&&&&&\\
      57M TXL / 141M total&&&&&\\
      75M TXL / 175M total&&&&&\\
      169M TXL / 353M total&&&&&\\
      265M TXL / 533M total&&&&&\\
    \end{tabular}
  \end{center}
\end{table}

Transformer-XL memory is a cached memory of previous attention layer inputs, concatenated to the keys and values during each forward pass. Inputs to intermediate layer are activations from the previous layer, which in themselves contain information about the past. Caching $M$ activations this way theoretically allows for an effective memory horizon of $M$ x $L$, where $L$ is the number of attention layers in the network.
Therefore, to avoid implicitly scaling effective Transformer-XL memory length, in our model size scaling experiments, we fix the number of layers in the Transformer, and scale parameters only by altering the Transformer embedding size ($d_\textrm{model}$), with the feed-forward size fixed at $4 d_\textrm{model}$, as is standard in Transformer architectures \citep{vaswani2017attention}.

\begin{table}[h!]
  \begin{center}
    \caption{Transformer hyperparameters for different model sizes.}
    \begin{tabular}{c|c|c|c|c|c|c}
    \label{tab:model_hparams}
      Model parameters & Embedding size & Blocks & Key size & Value size & Heads & FFW size\\
      \hline
      6M TXL / 41M total & 288 & 6 & 48 & 48 & 6 & 1152\\
      23M TXL / 76M total & 576 & 6 & 48 & 48 & 12 & 2304\\
      42M TXL / 112M total & 768 & 6 & 32 & 32 & 24 & 3072\\
      57M TXL / 141M total & 896 & 6 & 32 &32 & 28 & 3584\\
      75M TXL / 175M total & 1024 & 6 & 32 & 32 & 32 & 4096\\
      169M TXL / 353M total & 1536 & 6 & 48 & 48 & 32 & 6144\\
      265M TXL / 533M total & 1920 & 6 & 48 & 48 & 40 & 7680\\
    \end{tabular}
  \end{center}
\end{table}

\subsection{Scaling the memory length}
\label{app:memory-scaling}
Table \ref{tab:memory-scaling} shows the details of the experimental setup for the memory length scaling experiments in Section \ref{sec:scaling_results}. We show the effective memory timesteps for each experiment, computed as the number of cached network activations times the number of transformer blocks (6).

\begin{table}[h!]
  \begin{center}
    \caption{Experimental setup for scaling the memory length.}
    \begin{tabular}{c|c|c|c|c|c}
    \label{tab:memory-scaling}
      Model Parameters & Memory & Training task pool & Curriculum & Teacher & Training steps \\
      \hline
      \multirow{4}{*}{23M TXL / 76M total} & 600 & \multirow{4}{*}{200M} & \multirow{4}{*}{No-op} & \multirow{4}{*}{Table \ref{tab:Teacher}} & \multirow{4}{*}{25B}\\
      & 1800 &&&&\\
      & 3000 &&&&\\
      & 4200 &&&&\\
    \end{tabular}
  \end{center}
\end{table}

\subsection{Scaling the size of the task pool}
\label{app:task-scaling}
Table \ref{tab:task-scaling} shows the details of the experimental setup for scaling the size of the task pool in Section~\ref{sec:results_task_scaling}. 

\begin{table}[h!]
  \begin{center}
    \caption{Experimental setup for scaling the task pool size.}
    \begin{tabular}{c|c|c|c|c|c}
    \label{tab:task-scaling}
      Model parameters & Memory & Training task pool & Curriculum & Teacher & Training steps \\
      \hline
      \multirow{2}{*}{23M TXL / 76M total} & \multirow{4}{*}{1800} & 200M & \multirow{4}{*}{No-op} & \multirow{4}{*}{Table \ref{tab:Teacher}} & \multirow{4}{*}{25B}\\
      && 25B &&&\\
      \multirow{2}{*}{75M TXL / 175M total} &  & 200M & & & \\
      && 25B &&&\\
    \end{tabular}
  \end{center}
\end{table}

\subsection{Scaling the complexity of the task pool}
\label{app:task-complexity-scaling}
Table \ref{tab:task_complexity_scaling} shows the details of the experimental setup for scaling the complexity of the task pool in Appendix \ref{app:scaling_complexity}. In this experiment, the distillation teachers are different for the two agents we compare. Therefore we cannot disentangle the effects of distillation and task complexity. Nevertheless, the results remain indicative of the importance of task complexity. The teacher for the task distribution across multiple world topologies is trained as in Table \ref{tab:Teacher}. The teacher for the task distribution in a single room comes from a long lineage (> 6 generations) of distillation teachers starting with agents trained on XLand 1.0 \citep{xland}.

\begin{table}[h!]
  \begin{center}
    \caption{Experimental setup for scaling the complexity of the task distribution.}
    \begin{tabular}{c|c|c|c|c}
    \label{tab:task_complexity_scaling}
      Model parameters & Memory & Task pool & Curriculum  & Steps \\ 
      \hline
      6M TXL / 41M total & \multirow{5}{*}{1800} & \multirow{5}{*}{4k worlds $\times$ 50k games} & \multirow{5}{*}{No-op} & \multirow{5}{*}{23B}\\ 
      23M TXL / 76M total&&&&\\ 
      42M TXL / 112M total&&&&\\
      57M TXL / 141M total&&&&\\
      75M TXL / 175M total&&&&\\
      \hline
      6M TXL / 41M total &  \multirow{4}{*}{1800} & \multirow{4}{*}{1 world $\times$ 5k inits $\times$ 50k games} & \multirow{4}{*}{No-op}&\multirow{4}{*}{23B}\\ 
      23M TXL / 76M total&&&&\\ 
      42M TXL / 112M total&&&&\\
      57M TXL / 141M total&&&&\\
    \end{tabular}
  \end{center}
\end{table}

\subsection{Distillation enables scaling agents}
\label{app:distillation-exp}
Table \ref{tab:distillation-exp} shows the experimental setup for the distillation experiments in Section \ref{sec:results_kickstarting}.

\begin{table}[h!]
  \begin{center}
    \caption{Experimental setup for distillation experiments.}
    \begin{tabular}{c|c|c|c|c|c}
    \label{tab:distillation-exp}
      Model Parameters & Memory & Training task pool & Curriculum & Teacher & Steps\\
      \hline
      \multirow{2}{*}{TXL 23M TXL / 76M total} & \multirow{4}{*}{1800} & \multirow{4}{*}{See Sec~\ref{app:multi_agent_training}} & \multirow{4}{*}{PLR (\ref{tab:plr_hyperparameters})} &Table \ref{tab:multi-agent-teacher} & \multirow{4}{*}{22B}\\
      &&&&None&\\
      \multirow{2}{*}{TXL 265M TXL / 533M total}&&&& Table \ref{tab:multi-agent-teacher}&\\
      &&&&None&\\
    \end{tabular}
  \end{center}
\end{table}

\subsection{Training on more trials with skip memory}
\label{app:skip-memory-scaling}
Table \ref{tab:skip-memory-scaling} shows the details of the experimental setup for the memory scaling experiments in Section \ref{sec:results_few_to_many_shot}. This is the same setup as in Table \ref{tab:skip-memory-scaling}, except for the number of training steps and the variation of memory architecture and training trials discussed in the main text.

\begin{table}[h!]
  \begin{center}
    \caption{Experimental setup for experiments training on more trials with skip memory.}
    \begin{tabular}{c|c|c|c|c|c}
    \label{tab:skip-memory-scaling}
      Model Parameters & Memory & Training task pool & Curriculum & Teacher & Steps \\
      \hline
      \multirow{2}{*}{23M TXL / 76M total} & 1800 & \multirow{2}{*}{200M} & 
      \multirow{2}{*}{No-op} & 
      \multirow{2}{*}{Table \ref{tab:Teacher}} & \multirow{2}{*}{50B}\\
      & $1800 \times 4=7200$ &&&&\\
    \end{tabular}
  \end{center}
\end{table}

\newpage
\section{Additional Experiments}

\subsection{Multi-agent adaptation}\label{sec:ma_adaptation}

\begin{figure}[h!]
    \centering
    \includegraphics[width=0.6\linewidth]{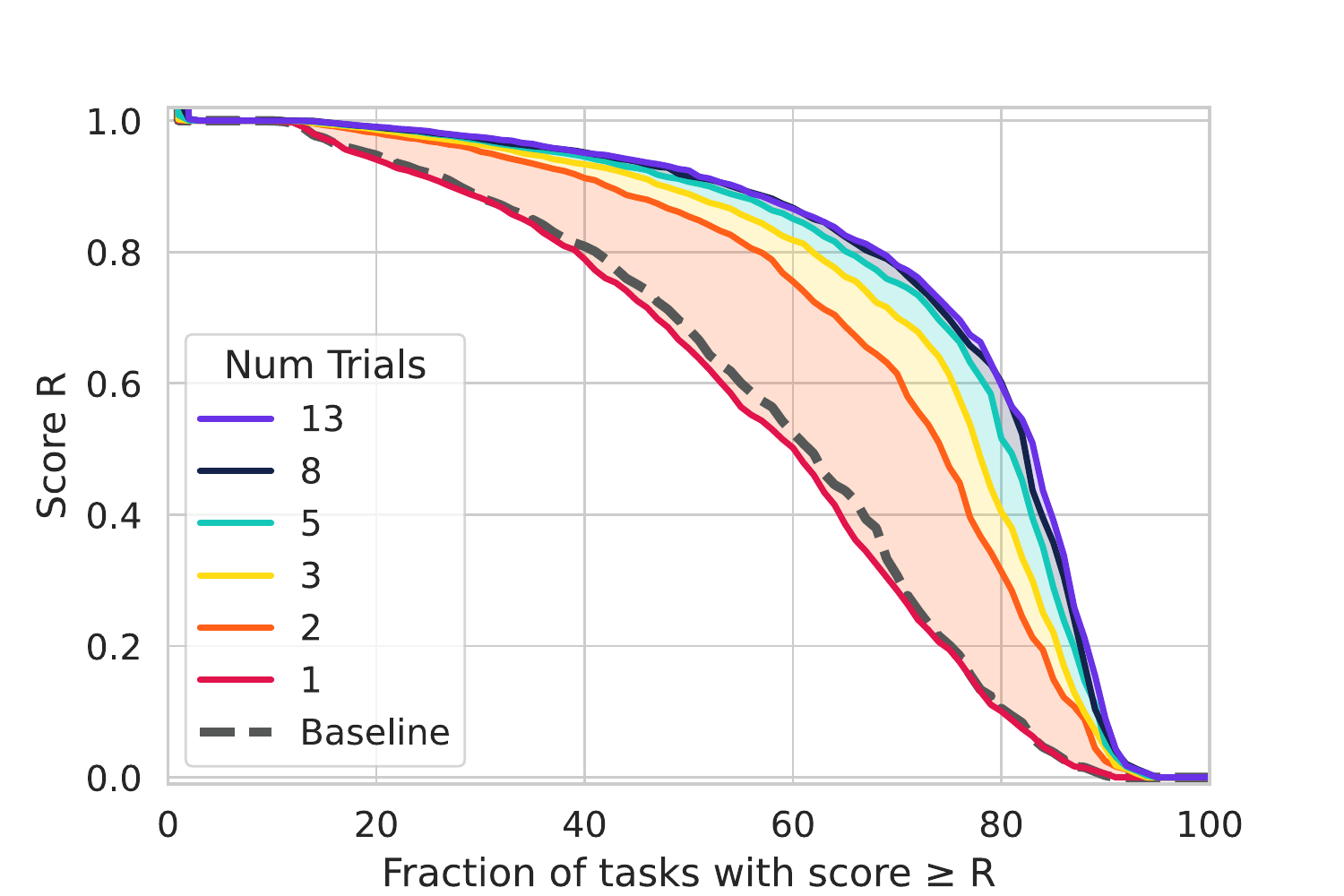}
    \caption{We report the distribution of normalised task scores over the multi-agent task test set when evaluated with various numbers of trials. All tasks are evaluated in cooperative self-play. On the $y$-axis is the total last-trial reward relative to that of an agent fine-tuned on the test tasks (approximating ``infinite trials'' performance). Curves moving further towards the top right corner indicate better performance. When given more trials, the agent achieves higher scores in the last trial, showing test-time adaptation across most of the task distribution (shaded regions). 
    }
    \label{fig:multi_agent_percentiles}
\end{figure}

\begin{figure}[h!]
    \centering
    \includegraphics[width=1.0\linewidth]{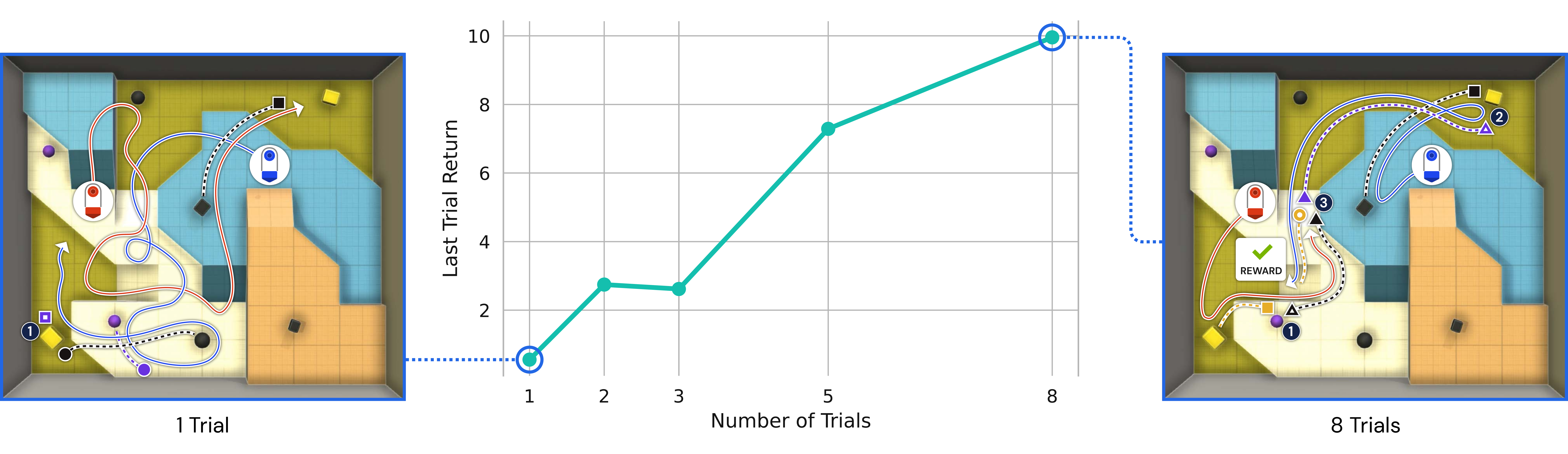}
    \caption{Average performance and representative behaviour of AdA on the probe task \texttt{Irreversible Production for Two} when evaluated in self-play with various numbers of trials. AdA's performance increases when given more trials, showing test-time adaptation. The top-down view images show representative last-trial trajectories when given different numbers of total trials. A corresponding \href{https://youtu.be/hBc-7YMVo0U}{video} for the case $k = 8$ shows the behaviour across all trials within one episode.}
    \label{fig:multi_agent_probe_emergent_coop}
\end{figure}

Figure \ref{fig:multi_agent_percentiles} demonstrates adaptation across a wide range of percentiles on a test-task set of multi-agent tasks. Figure \ref{fig:multi_agent_probe_emergent_coop} demonstrates last-trial performance of AdA in one particular probe task. To generate these plots, AdA was trained as described in Section \ref{sec:results_human_scale_adaptation} (Multi-agent). 
\subsection{Conditioning on number of shots doesn't affect agents' performance}  \label{sec:results_conditioning}

\begin{figure}[t!]
    \centering
    \includegraphics[width=0.5\linewidth]{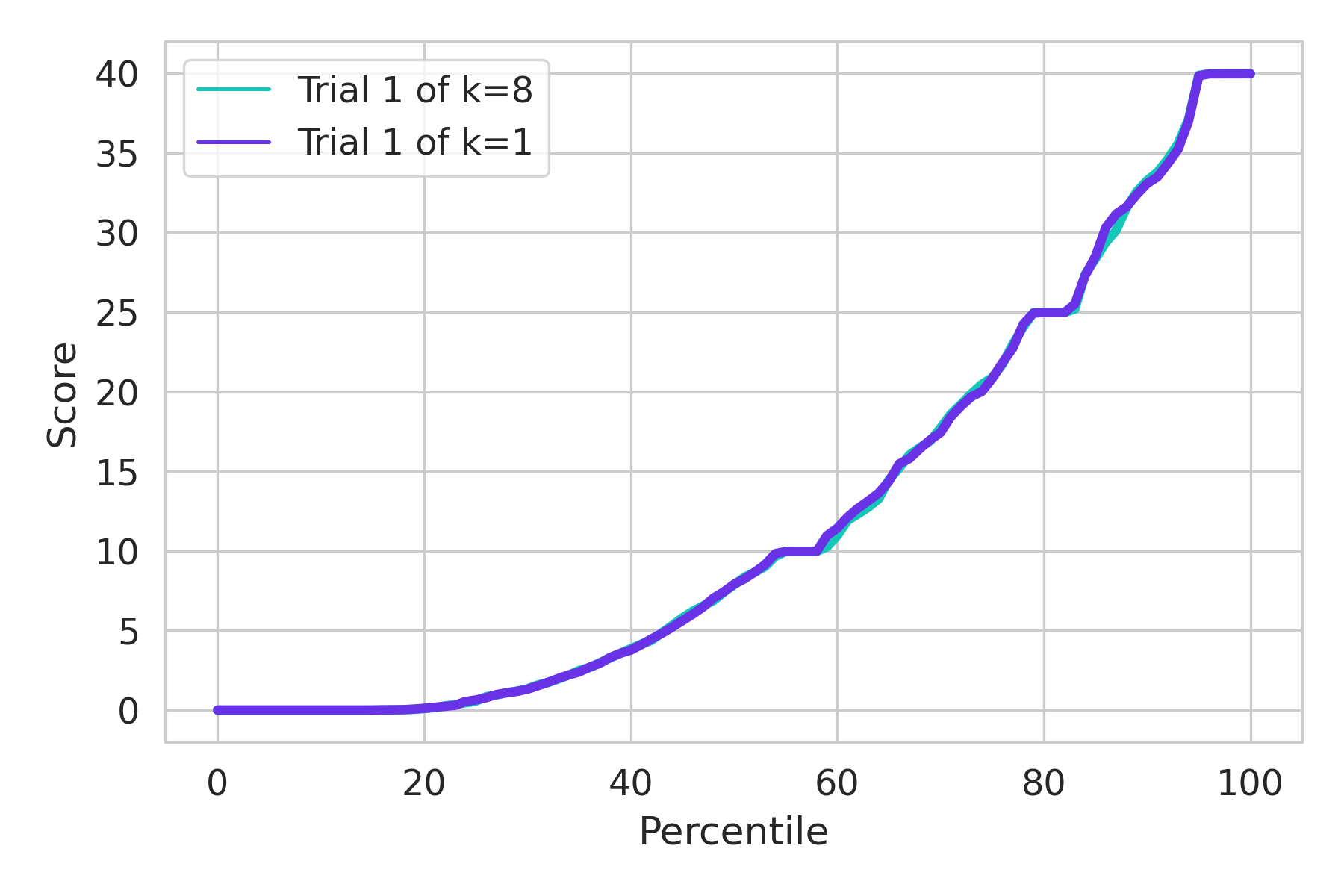}

    \caption{A comparison of the first-trial score in episodes with 1 trial and episodes with 8 trials. The lines are almost perfectly overlapping, which indicates that our agent does not leverage number-of-trials conditioning information to adjust its policy to a more exploratory one in early trials when more trials are available.}
    \label{fig:k-conditioning}
\end{figure}

Figure \ref{fig:k-conditioning} shows the score obtained by AdA for each percentile, in trial 1 of episodes with only 1 trial ($k=1$) and in trial 1 of episodes with 8 trials ($k=8$) in our held-out test set. The overlap of these lines indicates that AdA does not use the trial conditioning information it observes to adjust its behaviour in any way that affects its score. If the agent were to follow a more exploratory policy when it has more trials, we might expect the scores of trial 1 with $k=8$ to be lower than the score of trial 1 with $k=1$. 

This may be the optimal policy for our XLand 2.0 tasks, or it may reveal a limitation of our training procedure. One can imagine a scenario in which, knowing that there are $8$ trials in total, a Bayes-optimal policy chooses to display a less rewarding and more exploratory behaviour in trial $1$, compared to how it would behave if told that there was only a single trial in which to collect reward. For instance, an agent may be able to guarantee a deterministic reward later, having discovered some key information, at the cost of foregoing an stochastic reward early on. We did not directly incentivise this behaviour in our training process. In fact, we may have discouraged it, since AdA learns from all rewards in an episode (not just in the last trial), and with a discount factor smaller than 1, which could lead to myopic behaviour.  

\subsection{Scaling complexity of the task pool}
\label{app:scaling_complexity}

One final axis along which it is possible to scale our method is the overall complexity of the task distribution. We compare our main task distribution, as described in Section \ref{sec:methods_xverse} and Appendix \ref{app:presample_tasks} to a subset which maintains the use of the same goals, production rules, and objects, but eliminates any navigational complexity by having a single world topology: an empty room. Recall that we count the number of tasks as the product of number of worlds and the number of games. To disentangle the effects of scaling complexity versus scaling the sheer number of tasks, we add 5,000 unique object initialisation points for the empty room. These serve as the proxy ``4,000 worlds'' and are, by design, much less diverse and complex than the 4,000 worlds in the main training pool.\footnote{Note that the distribution over world topologies we use here is smaller than the distribution used in the model scaling experiments in Section \ref{sec:scaling_results}, and results are therefore not comparable across these sections.} For more details of the experimental setup, see Appendix \ref{app:task-complexity-scaling} and Table ~\ref{tab:task_complexity_scaling}.

\begin{figure}[t!]
    \centering
    \begin{subfigure}[b]{0.49\textwidth}
        \includegraphics[width=\textwidth]{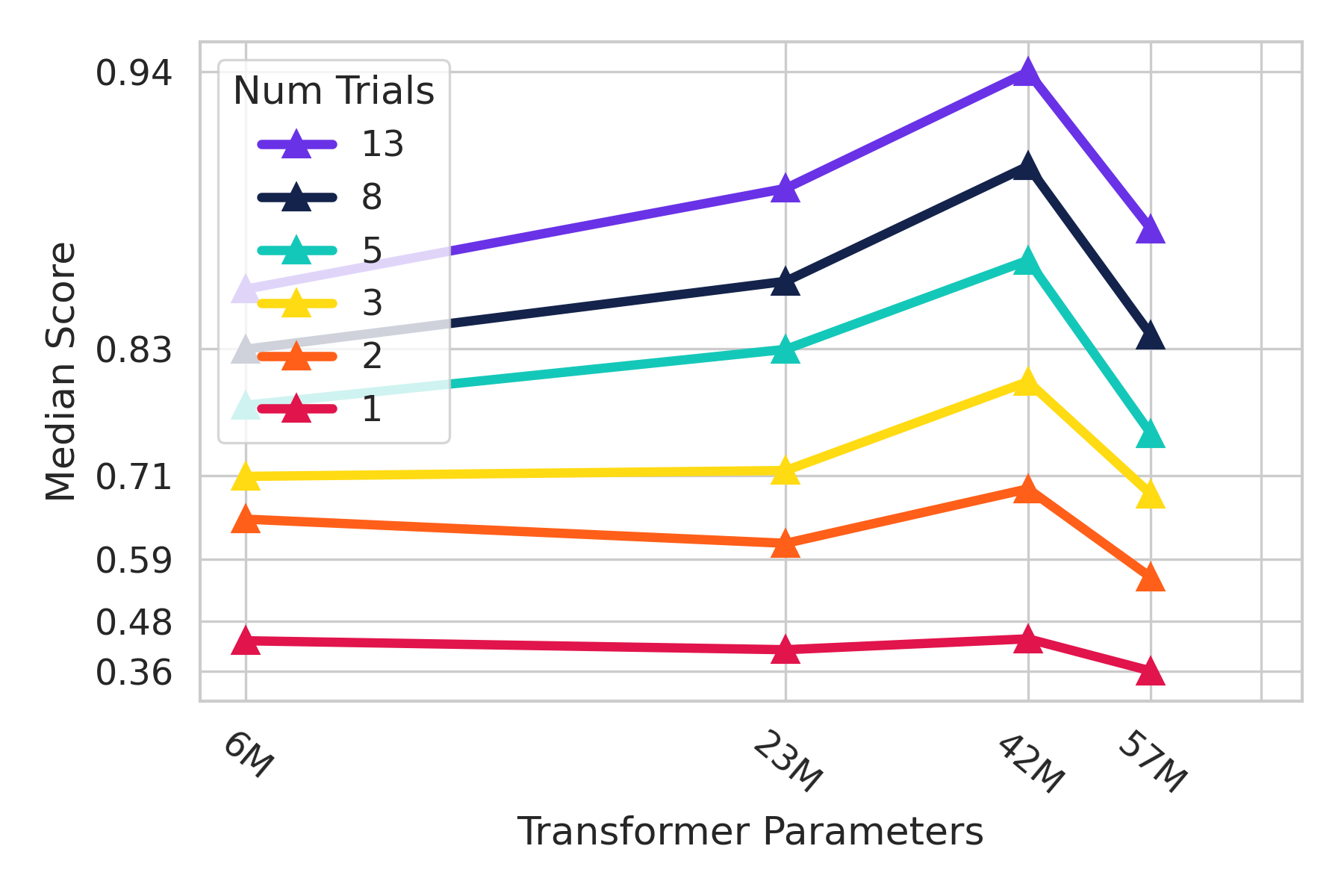}
        \caption{\centering Task distribution with only empty room inhibits scaling (median).}
        \label{subfig:complexity-scaling-empty-room-50th}
    \end{subfigure}
    \hfill
    \begin{subfigure}[b]{0.49\textwidth}
        \centering
        \includegraphics[width=\textwidth]{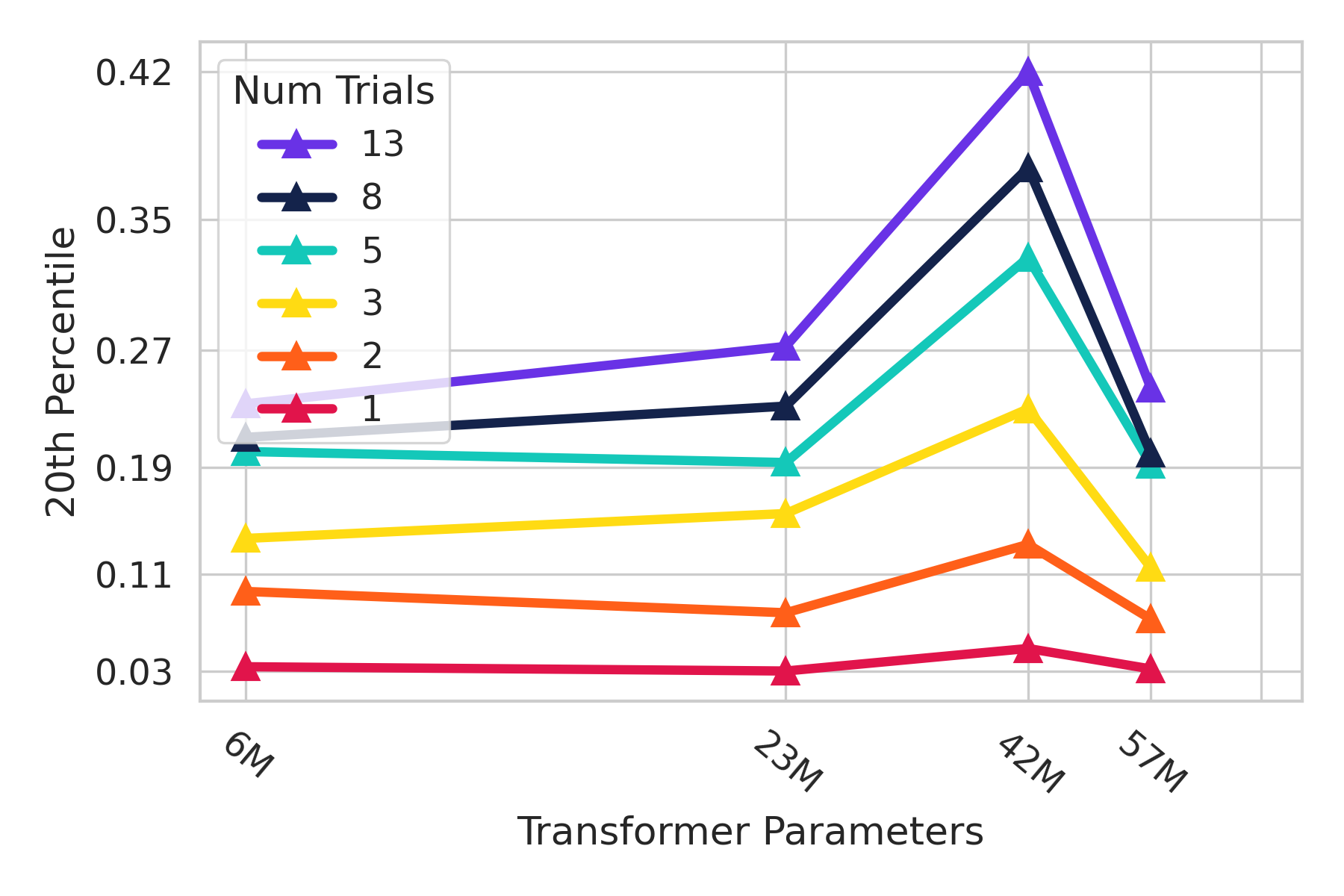}
        \caption{\centering Task distribution with only empty room inhibits scaling (20th percentile).}
        \label{subfig:complexity-scaling-empty-room-20th}
    \end{subfigure}
    
    \vspace{2mm}
    
    \begin{subfigure}[b]{0.49\textwidth}
        \centering
        \includegraphics[width=\textwidth]{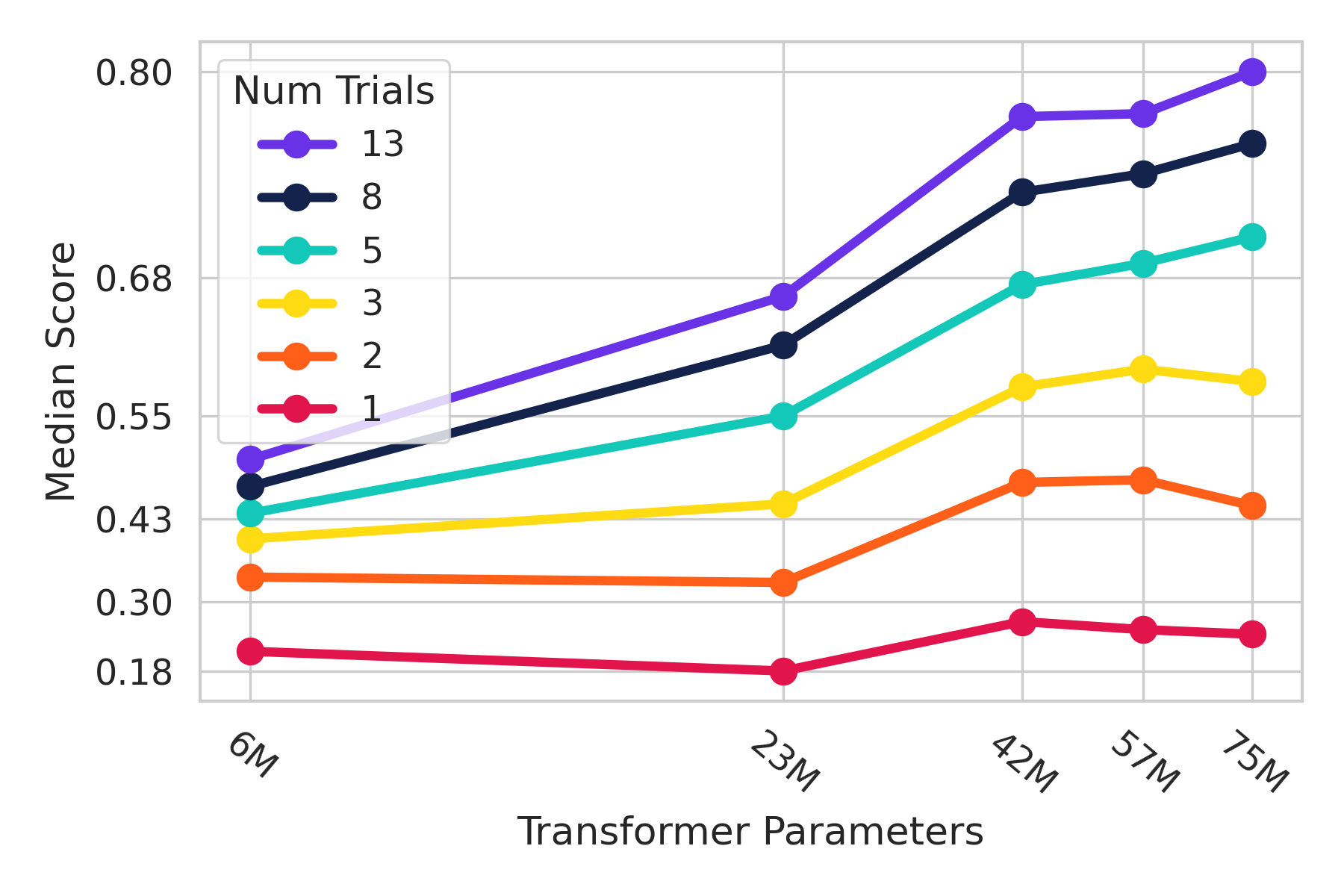}
        \caption{\centering Task distribution with many world topologies facilitates scaling (median).}
        \label{subfig:complexity-scaling-worlds-50th}
    \end{subfigure}
    \hfill
    \begin{subfigure}[b]{0.49\textwidth}
        \centering
        \includegraphics[width=\textwidth]{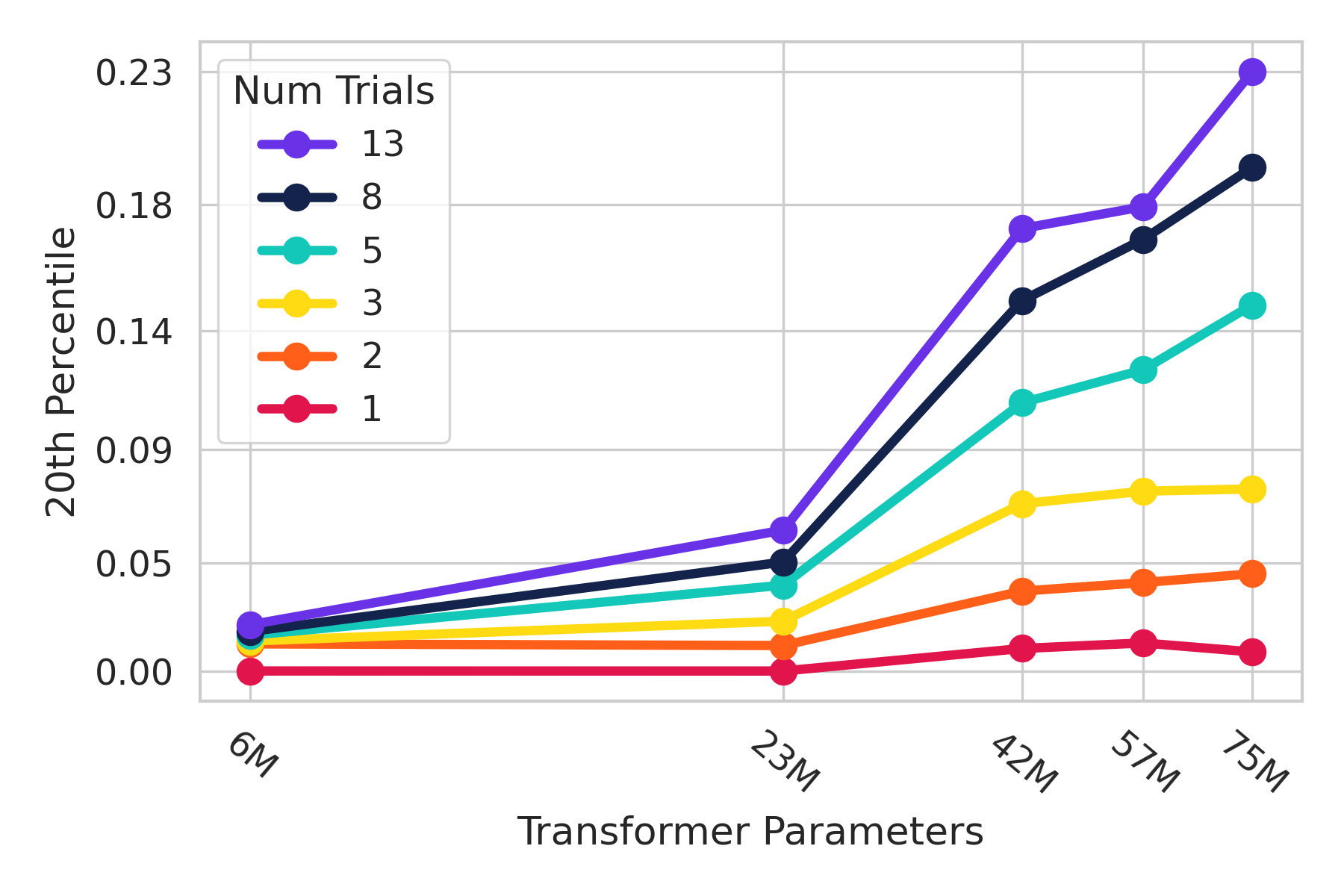}
        \caption{\centering Task distribution with many world topologies facilitates scaling (20th percentile).}
        \label{subfig:complexity-scaling-worlds-20th}
    \end{subfigure}
    \caption{The benefit of scaling model size is bottlenecked if the distribution is not complex enough, even if the total number of tasks is accounted for.}
\label{fig:task_complexity_scaling}
\end{figure}

In Figure \ref{fig:task_complexity_scaling}, we show that low environment complexity can be a bottleneck to scaling, by comparing the effectiveness of model scaling between agents trained on the two distributions, and each evaluated on their respective test sets. On both the median (Figure \ref{subfig:complexity-scaling-empty-room-50th}) and 20\textsuperscript{th} (Figure \ref{subfig:complexity-scaling-empty-room-20th}) percentiles in the empty room, we see that past a certain point (42M Transformer parameters), scaling model size begins to reduce performance. By contrast, in the distribution with many world topologies (Figures \ref{subfig:complexity-scaling-worlds-50th} and \ref{subfig:complexity-scaling-worlds-20th}), increased model size continues to improve performance far beyond this, showing improvements through at least 75M Transformer parameters.
Open-ended settings with unbounded environment complexity, such as multi-agent systems, may therefore be particularly important for scaling up adaptive agents.

\subsection{Computational cost}
\label{app:computational-cost-flops}
In the scaling experiments (Sections \ref{sec:scaling_results} and \ref{sec:results_task_scaling}), we compare agents after they have been trained for an equivalent number of steps. While this controls for sample efficiency of models, here we provide an analysis of the computational cost in FLOPs for a given experiment, and reproduce some of our scaling results, controlling for for compute cost. We see that bigger is not always better from this perspective. 
For each model size and memory length we use JAX \citep{jax2018github} cost analysis to estimate the number of FLOPs per frame of the learner step and actor step (Table \ref{tab:flops}).

\begin{figure}[t!]
    \centering
    \begin{subfigure}[b]{0.49\textwidth}
        \centering
        \includegraphics[width=\linewidth]{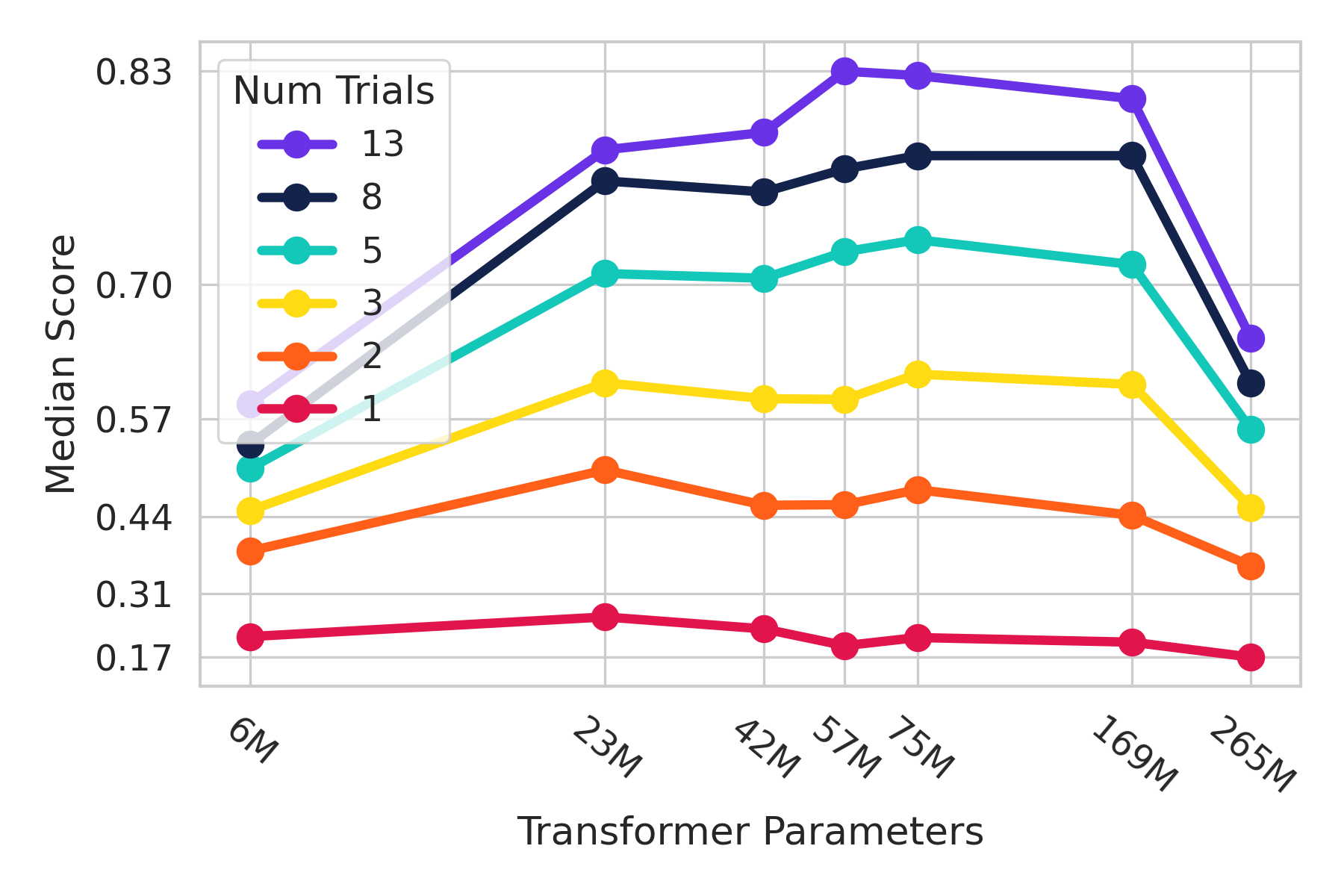}
        \vskip -3pt \caption{}
    \label{figure:scaling_parameters_total_flops_50th}
    \end{subfigure}
    \begin{subfigure}[b]{0.49\textwidth}
        \centering
        \includegraphics[width=\linewidth]{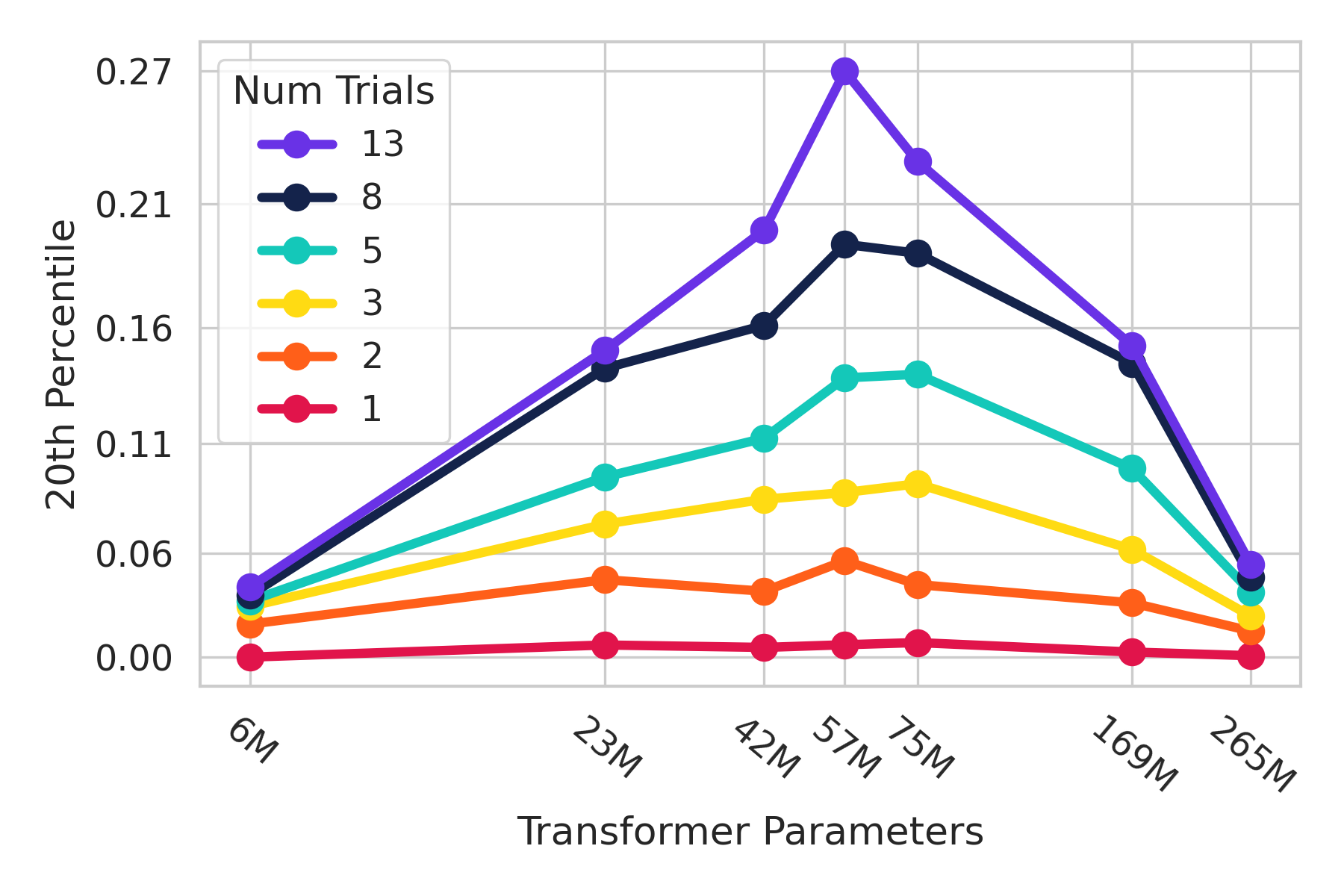}
        \vskip -3pt \caption{}
    \label{figure:scaling_parameters_total_flops_20th}
    \end{subfigure}
    \caption{
        Scaling Transformer model size controlling for the total number of FLOPs for the learner and actors, including auto-curriculum evaluation actors.
    \label{figure:scaling_parameters_total_flops}
    }
\end{figure}

\begin{figure}[t!]
    \centering
    \begin{subfigure}[b]{0.49\textwidth}
        \centering
        \includegraphics[width=\linewidth]{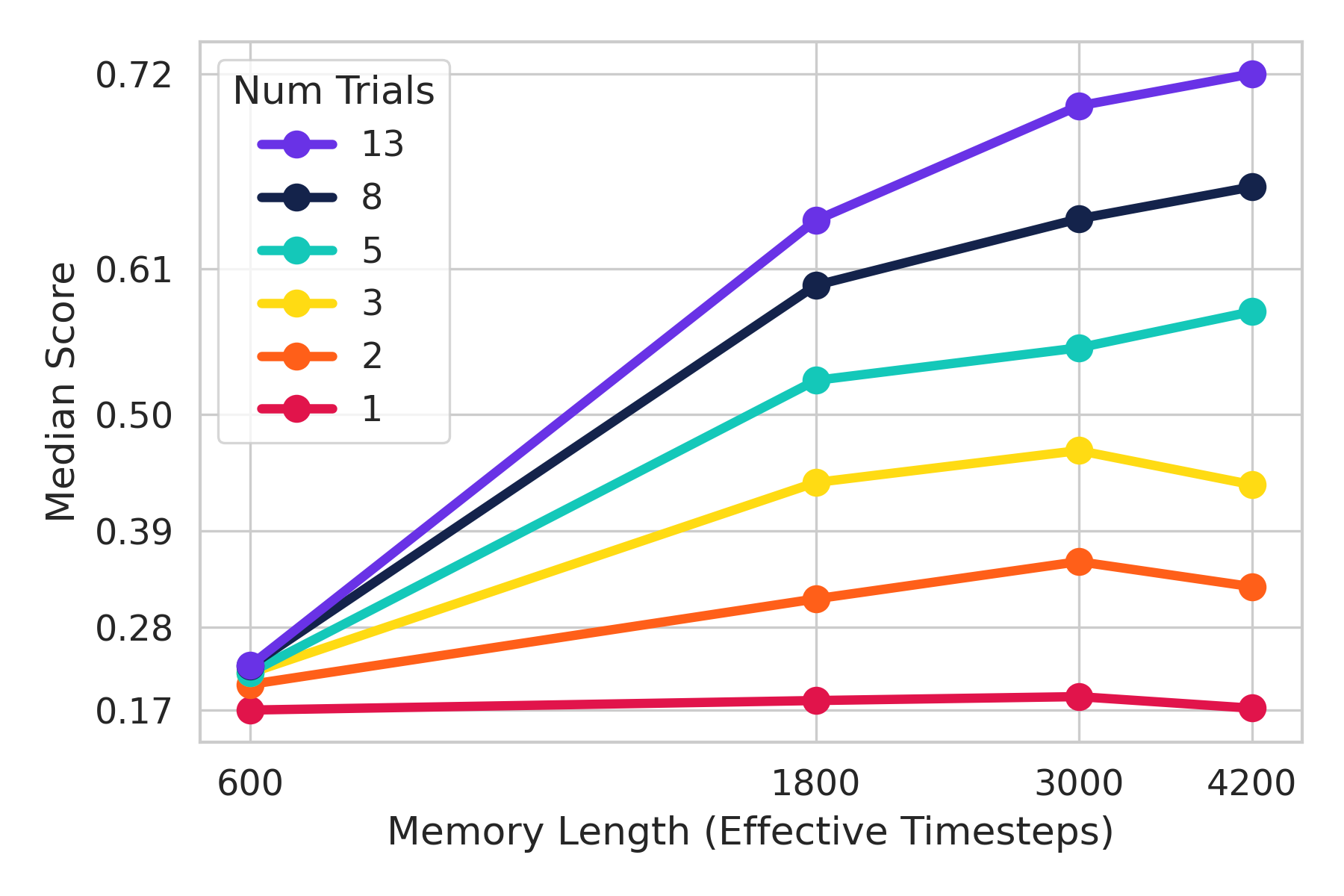}
        \vskip -3pt \caption{}
    \label{figure:scaling_memory_total_flops_50th}
    \end{subfigure}
    \begin{subfigure}[b]{0.49\textwidth}
        \centering
        \includegraphics[width=\linewidth]{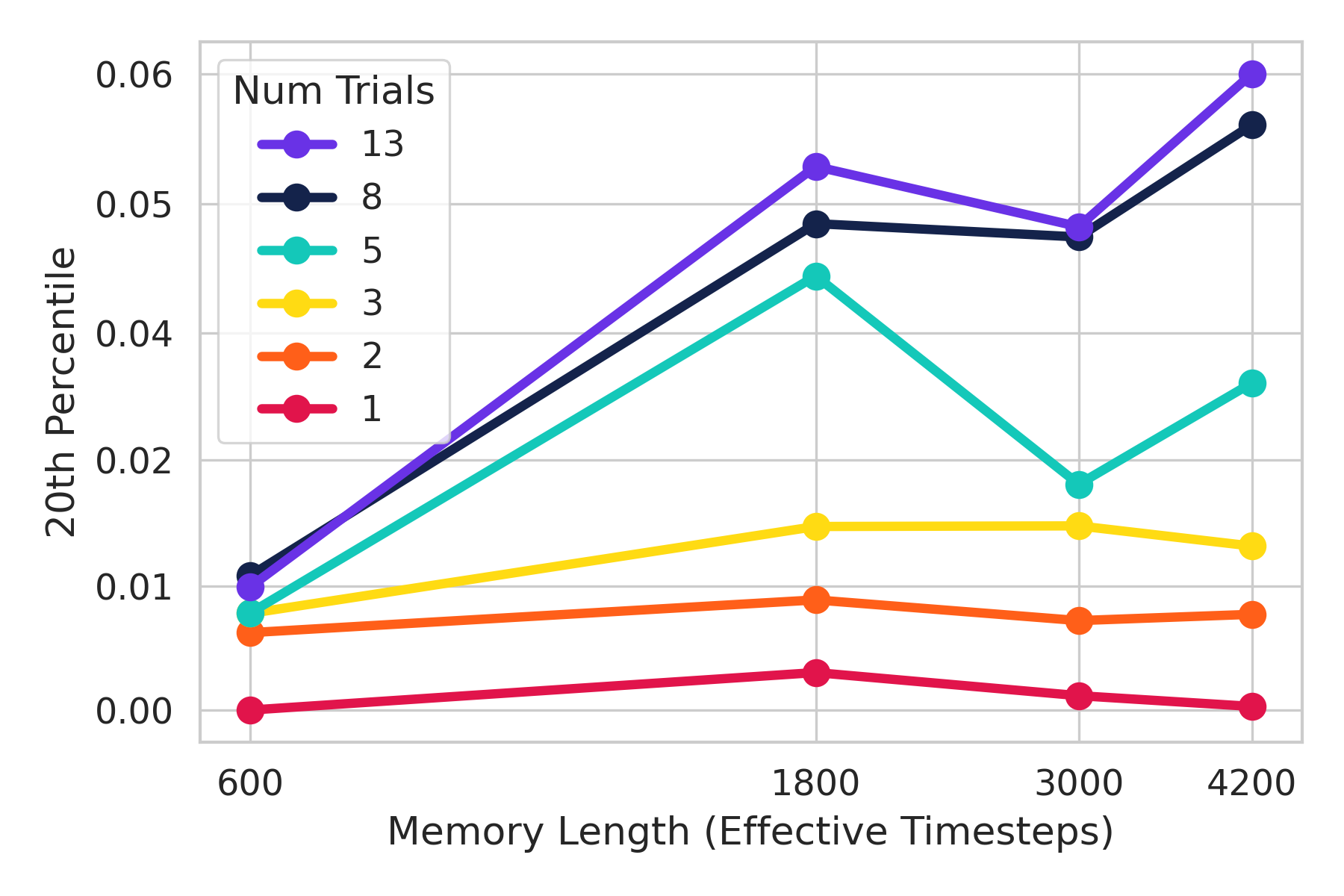}
        \vskip -3pt \caption{}
    \label{figure:scaling_memory_total_flops_20th}
    \end{subfigure}
    \caption{
        Scaling Transformer-XL memory controlling for the total number of FLOPs for the learner and actors, including auto-curriculum evaluation actors.
    \label{figure:scaling_memory_total_flops}
    }
\end{figure}

\begin{table}[t!]
  \begin{center}
    \caption{FLOPs per frame for different model sizes and memory lengths.}
    \begin{tabular}{c|c|c|c}
    \label{tab:flops}
      Model parameters & Memory & Learner FLOPs per frame & Actor FLOPs per frame \\
      \hline
      6M TXL / 41M total & \multirow{7}{*}{1800} & 1,138,005,632 & 736,987,392\\
      23M TXL / 76M total & & 1,380,255,078 & 2,580,445,440\\
      42M TXL / 112M total & & 1,623,461,663 & 4,489,239,552\\
      57M TXL / 141M total & & 1,821,422,230 & 6,063,742,976\\
      75M TXL / 175M total & & 2,048,253,663 & 7,879,863,808\\
      169M TXL / 353M total & & 3,243,189,525 & 17,560,326,144\\
      265M TXL / 533M total & & 4,443,008,234 & 27,358,181,376\\
      \multirow{3}{*}{23M TXL / 76M total } & 600 & 1,278,491,404 & 963,739,136\\
      & 3000 & 1,481,009,258 & 4,181,042,688\\
      & 4200 & 1,582,270,213 & 5,789,702,144\\
    \end{tabular}
  \end{center}
\end{table}

We multiply the values in Table \ref{tab:flops} by the number of learner/actor steps for each experiment, then,
for a given comparison, we take the largest such value common to all experiments (usually associated with the smallest model) as the total FLOPs, and make the comparison of each model at this number of FLOPs. The results for the FLOPs-matched model scaling experiments are shown in Figure \ref{figure:scaling_parameters_total_flops}. We see a reduction in performance as the model size grows beyond a ``sweet spot'' around 57M Transformer parameters (141M total parameters). Results for the FLOPs-matched memory scaling experiments in Figure \ref{figure:scaling_memory_total_flops} show that there is still benefit to increasing context lengths given a fixed computational budget. Details of the compute used for these experiments can be found in Tables \ref{tab:parameter_scaling_total_flops} (model size) and \ref{tab:memory_scaling_total_flops} (memory).

\begin{figure}[b!]
    \centering
    \begin{subfigure}[b]{0.49\textwidth}
        \centering
        \includegraphics[width=\linewidth]{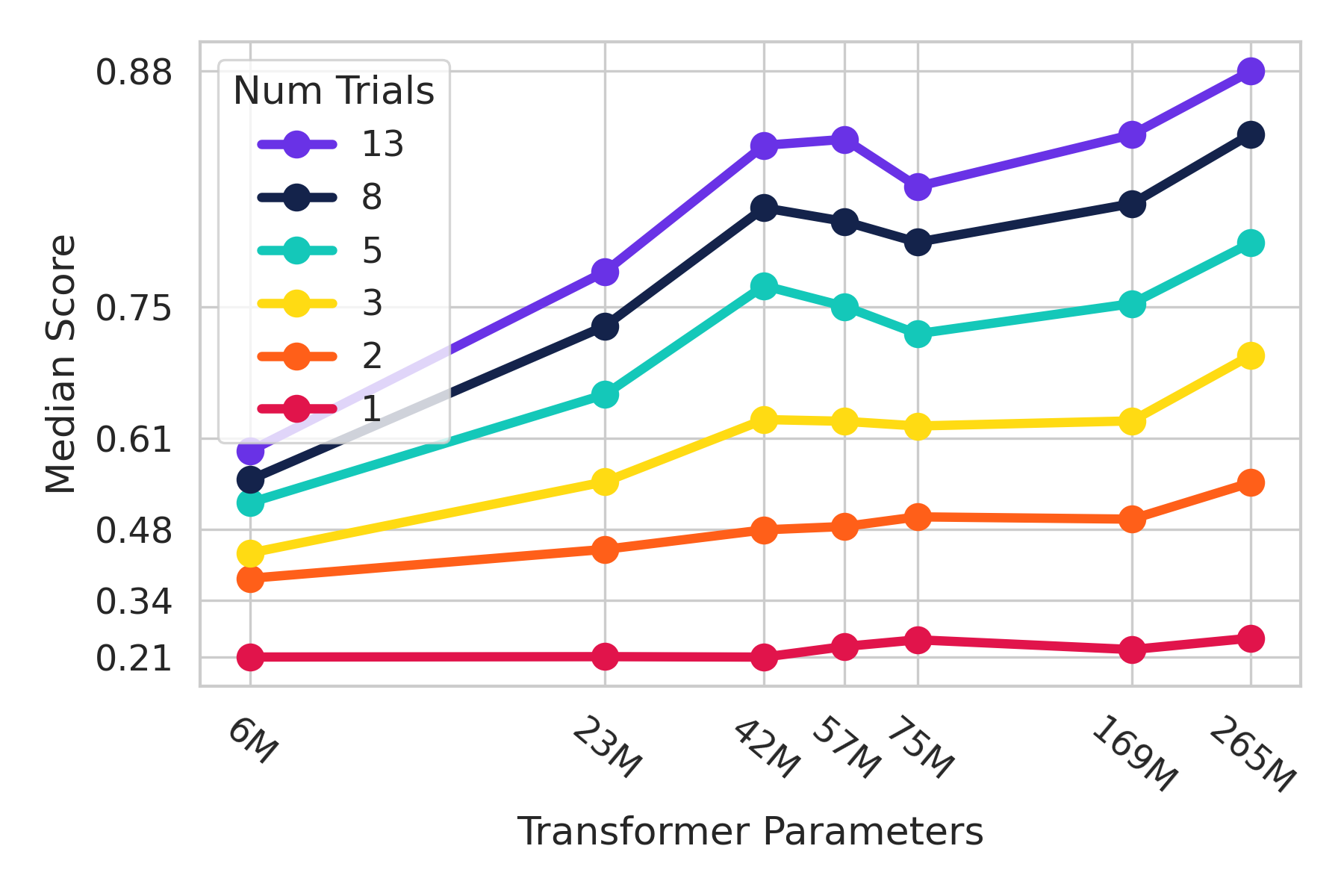}
        \vskip -3pt \caption{}
    \label{figure:scaling_parameters_learner_flops_50th}
    \end{subfigure}
    \begin{subfigure}[b]{0.49\textwidth}
        \centering
        \includegraphics[width=\linewidth]{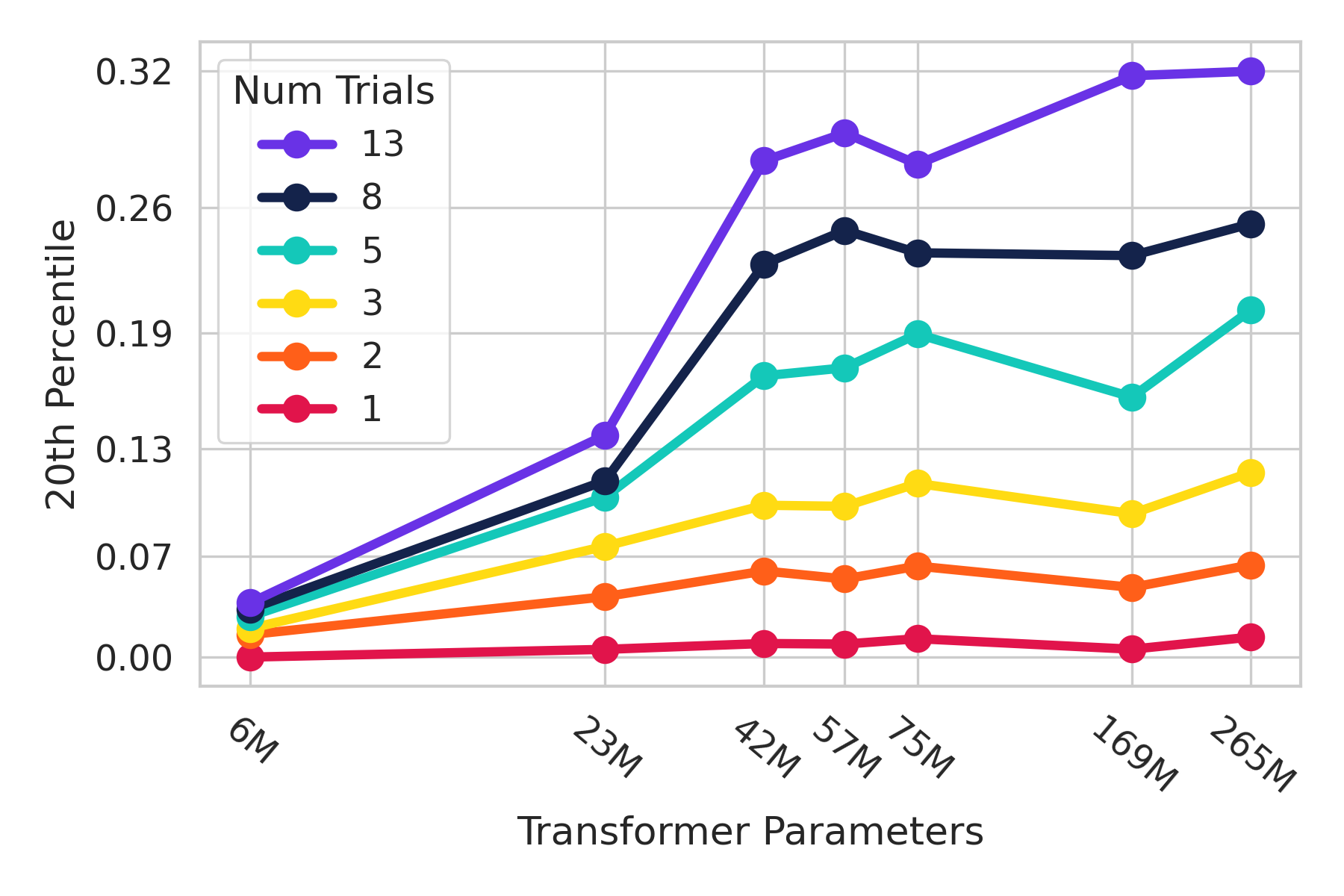}
        \vskip -3pt \caption{}
    \label{figure:scaling_parameters_learner_flops_20th}
    \end{subfigure}
    \caption{
        Scaling Transformer-XL model size controlling for the number of learner FLOPs.
    \label{figure:scaling_parameters_learner_flops}
    }
\end{figure}

\begin{figure}[b!]
    \centering
    \begin{subfigure}[b]{0.49\textwidth}
        \centering
        \includegraphics[width=\linewidth]{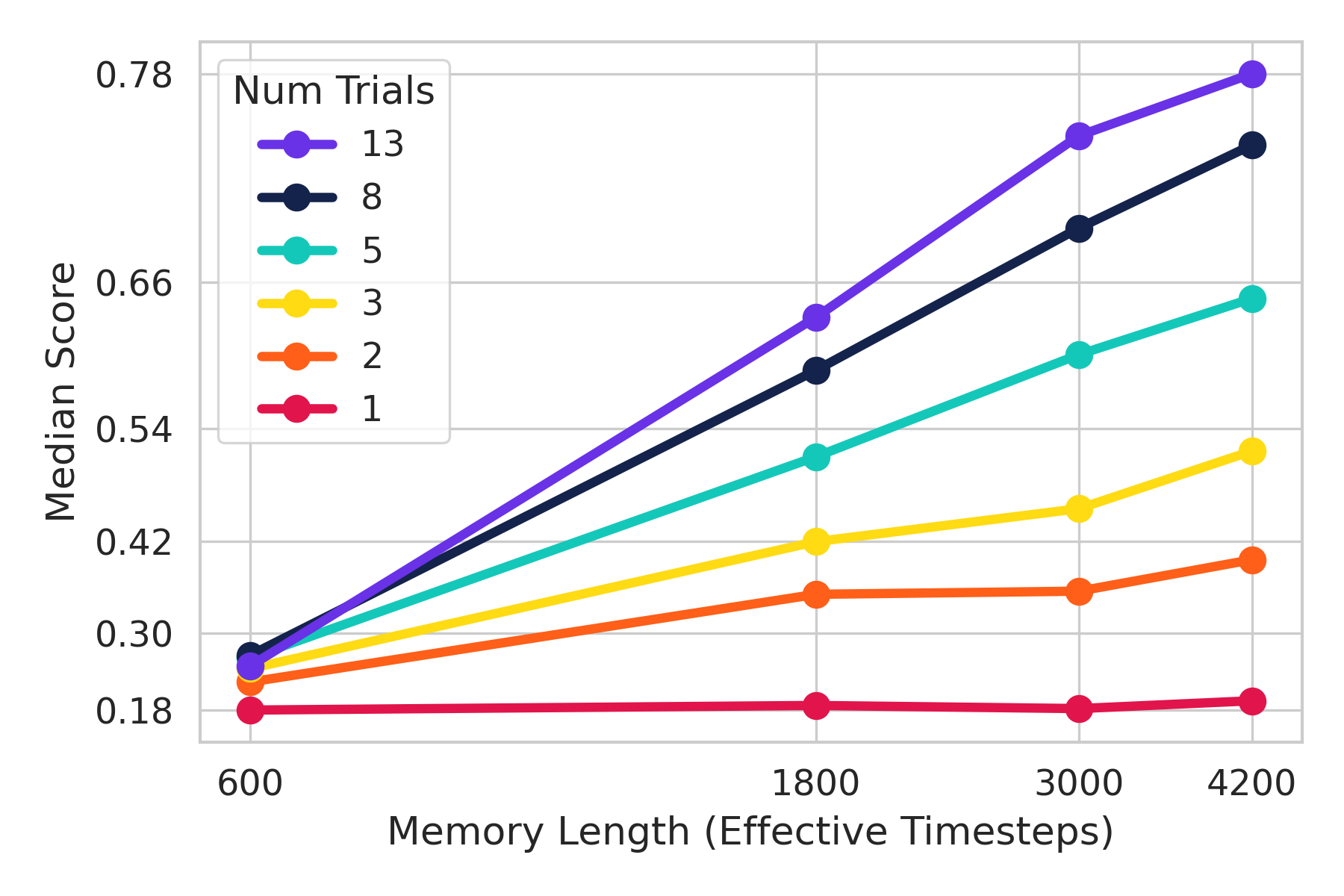}
        \vskip -3pt \caption{}
    \label{figure:scaling_memory_learner_flops_50th}
    \end{subfigure}
    \begin{subfigure}[b]{0.49\textwidth}
        \centering
        \includegraphics[width=\linewidth]{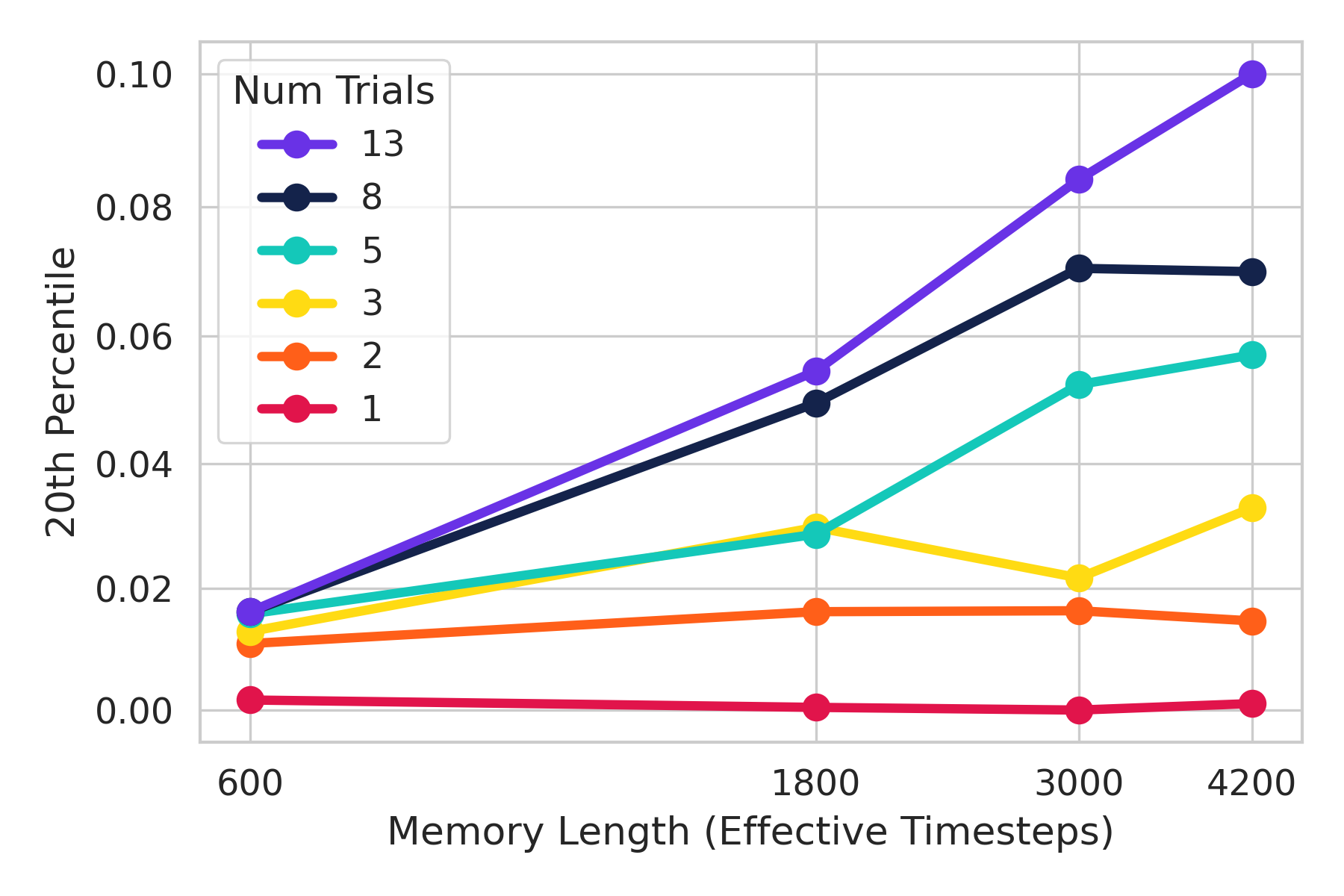}
        \vskip -3pt \caption{}
    \label{figure:scaling_memory_learner_flops_20th}
    \end{subfigure}
    \caption{
        Scaling Transformer-XL memory length controlling for the number of learner FLOPs.
    \label{figure:scaling_memory_learner_flops}
    }
\end{figure}

We note that Table \ref{tab:flops} indicates poor scaling of actor step FLOPs with model size, and suspect this could be due to poor optimisation of a single-step query of the Transformer on TPU, compared to the operation batching achieved with a rollout length of 80 on the learner. In order to account for this and for potential discrepancies in number of actor steps due to differences in curriculum evaluation based on model quality, we also provide plots which only account for the learner step FLOPs for each model: Figures  \ref{figure:scaling_parameters_learner_flops} (model size), and \ref{figure:scaling_memory_learner_flops} (memory length). These might be more informative, and show that performance still increases as a function of model size and memory length, albeit not as steeply as when controlling for sample efficiency directly. Details of the compute used for these experiments is shown in Tables \ref{tab:parameter_scaling_learner_flops} (model size) and \ref{tab:memory_scaling_learner_flops} (memory length).

\begin{table}[t!]
  \begin{center}
    \caption{Corresponding learner steps given total FLOPs for each model size.}
    \begin{tabular}{c|c|c|c}
    \label{tab:parameter_scaling_total_flops}
      Model parameters & Memory & Total FLOPs & Learner steps \\
      \hline
      6M TXL / 41M total & \multirow{7}{*}{1800} & \multirow{7}{*}{$2.0 \times 10^{20}$} & 97B\\
      23M TXL / 76M total & & & 44B\\
      42M TXL / 112M total & & & 27B\\
      57M TXL / 141M total & & & 23B\\
      75M TXL / 175M total & & & 18B\\
      169M TXL / 353M total & & & 9B\\
      265M TXL / 533M total & & & 5B\\
    \end{tabular}
  \end{center}
\end{table}

\begin{table}[t!]
  \begin{center}
    \caption{Corresponding learner steps given total FLOPs for each memory length.}
    \begin{tabular}{c|c|c|c}
    \label{tab:memory_scaling_total_flops}
      Model parameters & Memory & Total FLOPs & Learner steps \\
      \hline
      \multirow{4}{*}{23M TXL / 76M total} & 600 & \multirow{4}{*}{$7.9 \times 10^{19}$} & 31B\\
      & 1800 & & 17B\\
      & 3000 & & 11B\\
      & 4200 & & 8B\\
    \end{tabular}
  \end{center}
\end{table}

\begin{table}[t!]
  \begin{center}
    \caption{Corresponding learner steps given learner-step-only FLOPs for each model size.}
    \begin{tabular}{c|c|c|c}
    \label{tab:parameter_scaling_learner_flops}
      Model parameters & Memory & Learner step FLOPs & Learner steps \\
      \hline
      6M TXL / 41M total & \multirow{7}{*}{1800} & \multirow{7}{*}{$1.0 \times 10^{20}$}& 92B\\
      23M TXL / 76M total & & & 76B\\
      42M TXL / 112M total & & & 65B\\
      57M TXL / 141M total & & & 58B\\
      75M TXL / 175M total & & & 51B\\
      169M TXL / 353M total & & & 32B\\
      265M TXL / 533M total & & & 23B\\
    \end{tabular}
  \end{center}
\end{table}

\begin{table}[t!]
  \begin{center}
    \caption{Corresponding learner steps given learner-step-only FLOPs for each memory length.}
    \begin{tabular}{c|c|c|c}
    \label{tab:memory_scaling_learner_flops}
      Model parameters & Memory & Learner step FLOPs & Learner steps \\
      \hline
      \multirow{4}{*}{23M TXL / 76M total} & 600 & \multirow{4}{*}{$3.9 \times 10^{19}$} & 39B\\
      & 1800 & & 36B\\
      & 3000 & & 29B\\
      & 4200 & & 25B\\
    \end{tabular}
  \end{center}
\end{table}

\begin{figure}[t]
    \centering
    \begin{subfigure}[b]{0.49\textwidth}
        \centering
        \includegraphics[width=\linewidth]{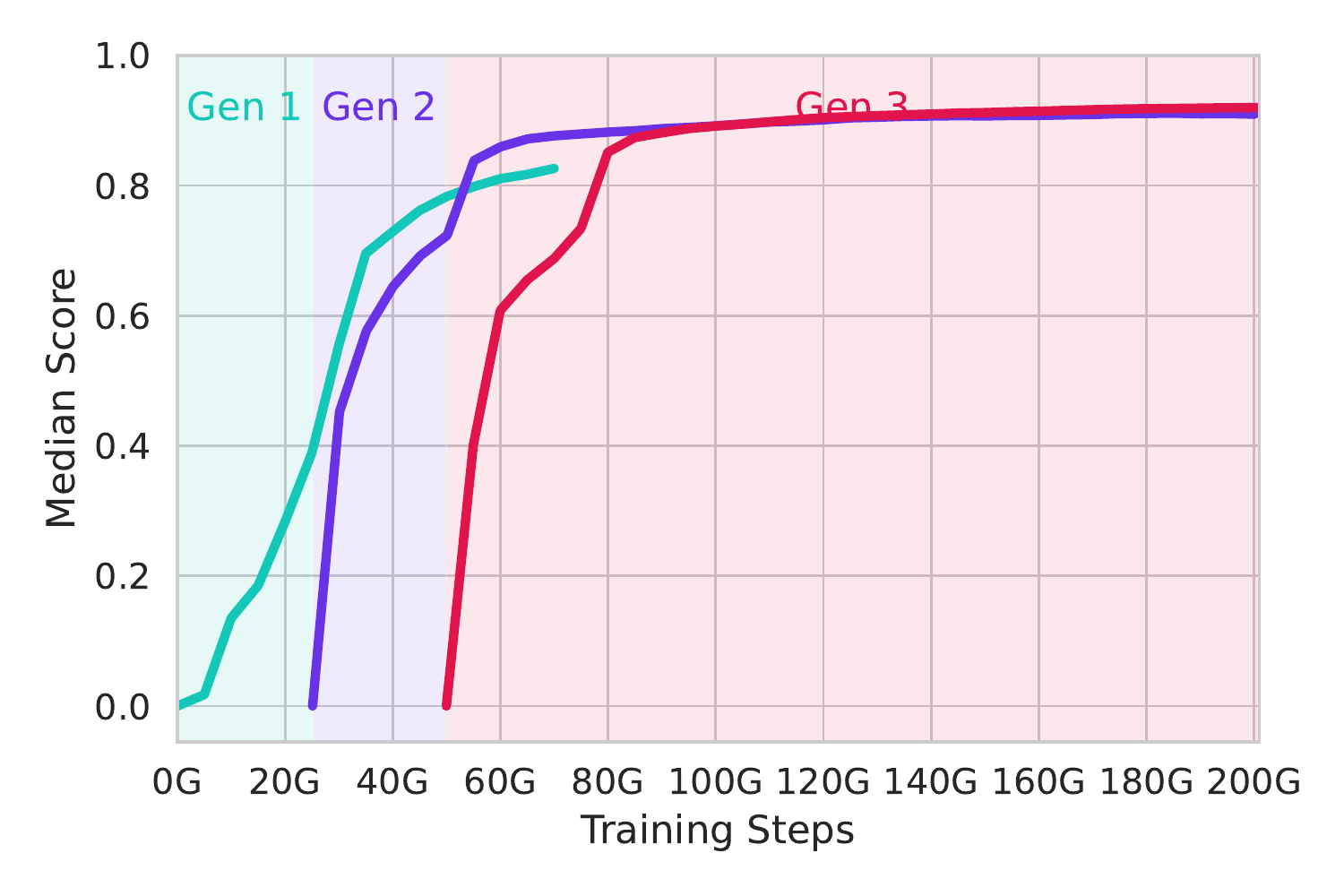} 
        \vskip -3pt \caption{}
    \end{subfigure}
    \begin{subfigure}[b]{0.49\textwidth}
        \centering
        \includegraphics[width=\linewidth]{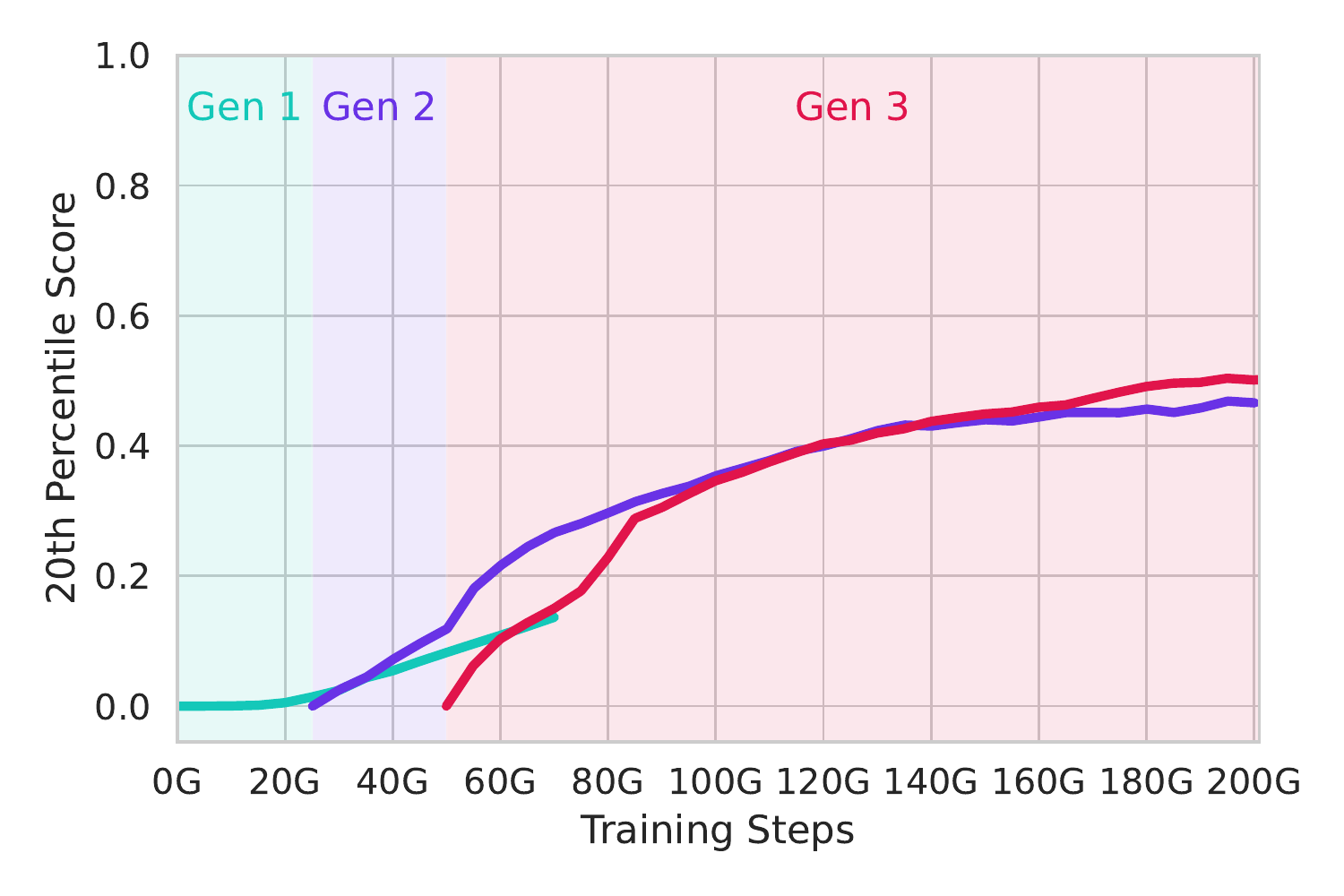}  
        \vskip -3pt \caption{}
    \end{subfigure}
    \caption{
        Normalised few-shot score over three generations using the 23M parameter Transformer-XL. The first and second generations correspond to the agents shown in Figure~\ref{fig:distillation}. The third generation is distilled from the second after it has been trained for 25 billion steps. The $x$-axis counts the combined amount of experience, starting from the experience collected by the original teacher. The third generation shows additional gains over the second, but to a lesser degree than the gap between the first and second generations.
    }
    \label{fig:app_distillation}
\end{figure}
\subsection{Repeated distillation}
\label{app:repeated_distillation}
In Section~\ref{sec:results_kickstarting}, we show that distilling an agent into an identical student can lead to large increases in the agent's performance. Here, we investigate the potential benefits of applying the procedure repeatedly. To this end, we continue the experiment shown in Figure~\ref{fig:distillation} and add a third generation, using a snapshot taken from the previous student after 25 billion frames, equivalent to 50 billion frames of total experience when taking the teacher's experience into account. Figure~\ref{fig:app_distillation} shows that applying this procedure repeatedly can indeed lead to additional benefits; however, we observe diminishing returns with successive generations. 

\section{Human-Timescale Adaptation}\label{appendix:human_scale_adaptation}
In this section we provide more details regarding our claim of human-timescale adaptation.

\subsection{Probe tasks}\label{appendix:probe_tasks}

Tables \ref{probe-task-desc-single-agent} and \ref{probe-task-desc-multi-agent} describe in detail the single-agent and multi-agent probe tasks respectively. Unless noted otherwise, all probe tasks are set in complex worlds with many objects in them, and use multiple production rules, which are fully hidden from the players.

\renewcommand*{\arraystretch}{1.4}
\begin{longtable}{|>{\raggedright}p{4cm}|p{12cm}|}
\caption{Single-agent probe tasks}
    \label{probe-task-desc-single-agent}
    \\ \hline
\textbf{Name} & \textbf{Description} \\ \hline
\textbf{Wrong pair disappears} & The player’s goal is to hold a black cube, which does not exist among the initial objects. There are two (hidden) rules. The player needs to identify the correct world state which triggers the rule that creates the cube and not the one which destroys the necessary inputs. All this is embedded in a challenging world layout with one-way drops and limited visibility. \\ \hline
\textbf{Wrong pair disappears, partial hiding} & `Wrong pair disappears', but instead of hiding all rules completely we only hide the input objects for both rules (outputs and conditions are fully visible). \\ \hline
\textbf{Irreversible production} & Similar to `Wrong pair disappears', but this task has multiple dead ends (rules which create unsolvable states). \\ \hline
\textbf{Irreversible production, all rules visible} & `Irreversible production', but with all rules fully visible to the player. \\ \hline
\textbf{Push, don't lift} & The vast majority of training and evaluation tasks require lifting objects. Here two hidden rules destroy any object when lifted. In order to create the goal state, some ``lateral thinking'' is necessary: the player needs to identify that pushing the cubes with their body is possible. \\ \hline
\textbf{Push, don't lift, with distractors} & Similar to `Push, don't lift', but here a large number of distractor objects in the world make this a much more challenging exploration task for any player ignoring the objects mentioned in the goal. \\ \hline
\textbf{Spacer tool} & Two objects need to be brought close together, but lifting them or touching them (with the avatar's body) destroy them. The solution is to use another object in the world as a tool to push them together. \\ \hline
\textbf{Transform to transport} & Again, two objects need to be brought close to each other, but lifting and touching them destroys them. The solution here is to exploit a set of rules that can turn one of the objects into something that can be carried safely, and then turn it back into the original object once it is in the right place. \\ \hline
\textbf{Small workstation} & A very hard object handling task. 8 objects near the player's spawn position need to be combined in different ways and 5 rules need to be triggered (some multiple times) to create the goal object. This task is set on top of a tall plateau and it is very easy to fail by touching an object too hard and throwing it off the edge of the plateau. \\ \hline
\textbf{Small workstation, all rules visible} & The same as `Small workstation', but all rules are visible to the player. \\ \hline
\textbf{Crafting pyramid} & Eight objects need to be recursively combined first into four, then two and then ultimately one final object. This requires triggering a chain of 7 rules. This seems easy for humans. But the lack of intermediate reward makes this a hard hierarchical credit assignment task for agents. \\ \hline
\textbf{Crafting tree, all rules visible} & Similar to the crafting trees in video games like Minecraft, this multi-step task requires triggering different rules in a chain to create the goal object. All objects exist in the world multiple times, making many different solutions viable. \\ \hline
\textbf{Crafting tree, hidden shortcut} & Identical to `Crafting tree, all rules visible', but one additional (hidden) rule exists. This allows the player to take a shortcut that lets them finish the task faster than executing only the visible rules. \\ \hline
\textbf{Antimatter} & In a world full of yellow and black spheres, the goal is for no pair of black and yellow spheres to ``see each other'' (no direct line of sight). Beyond moving the objects and blocking line of sight with the avatar there exists a production rule which destroys any yellow sphere and black sphere pair which touch (similar in spirit to matter and antimatter particles). This is all embedded in a world requiring advanced navigation skills. \\ \hline
\textbf{Antimatter with creation} & Similar to `Antimatter', only here a third object (purple pyramids) exists which duplicates any sphere touching it. In addition, this task is set on two plateaus, making it easier to break line of sight and reducing the navigation challenge. \\ \hline
\textbf{Pyramid in a haystack} & To create the necessary yellow pyramid, the player needs to find and hold the purple cube. There are several distractor objects and distractor rules in this world, requiring the player to deal not just with a hard exploration challenge but also a very noisy environment. \\ \hline
\textbf{Protect the egg} & The player is tasked to hold the single yellow sphere in the world. A large number of other spheres exist in the world. These destroy the yellow sphere on collision. As these touch each other or get near the player they get duplicated. This can lead to a constantly growing number of objects, filling up the world. \\ \hline
\textbf{3 spheres jumbled} & 3 spheres exist in the world. Holding one of them creates the goal object. Only one sphere can be reached within the 10-second time limit, meaning that the optimal policy on the first trial is to choose uniformly at random. \\ \hline
\textbf{Two doors} & The goal object is hidden behind one of the two large objects (the ``doors'') positioned at opposite ends of the world. Only one of them can be reached in time, so the player needs to decide between exploring one of them per trial. \\ \hline
\textbf{Same signature: match colours} & All `Same signature' tasks have the exact same world layout, goal and number of fully-hidden rules. This means they look exactly the same to a player starting out. This one only requires two objects of matching colour to be brought close together to create the goal object, using only one production rule out of three. \\ \hline
\textbf{Same signature: three steps} & This `Same signature' task variant requires the player to trigger all three (hidden) production rules to create the goal object. \\ \hline
\textbf{Same signature: two dead ends} & In this `Same signature' task variant, two of the tree rules are dead ends, leading to an unsolvable world state. Only one rule is helpful (and in fact required) to solve the task. \\ \hline
\textbf{Same signature: destroy to protect} & To solve this `Same signature' task variant the player first needs to destroy a black sphere (by getting near it) before creating the goal object (a yellow cube). Otherwise when the black sphere ``sees'' the yellow cube, both get destroyed. This is a very hard credit assignment challenge even for human players (it is hard to notice what is going on). \\ \hline
\textbf{Don't act} & This task is pre-solved: the player is getting reward from the very beginning. If they lift or touch one of the two objects in the world, the object gets destroyed and reward is now impossible. \\ \hline
\textbf{Don't peek} & This task is `pre-solved': the player is getting reward from the very beginning. However, the player will destroy any object they look at, at which point reward is impossible. So the optimal strategy is to not look at anything but the sky. \\ \hline
\textbf{Navigation: find the cube} & This memory task is not using production rules. It uses a large world with the goal object (a cube) hidden after a very winding path. \\ \hline
\textbf{Navigation: find the cube with teaser} & This memory task is not using production rules. It is set in a large world with the goal object (a cube) hidden after a very winding path. The object is visible from the spawn point but out of sight after starting to traverse the terrain. \\ \hline
\textbf{Navigation: hold up high} & This memory task is not using production rules. The goal object (a pyramid) is hidden on top of a plateau and can only be seen when nearly there. \\ \hline
\textbf{Object permanence: yellow cube} & This memory task is not using production rules. The goal object (a yellow cube) is visible from the spawn point. After moving for a bit, a decision between two paths has to be made, with the correct path being to the right. At this point the cube is no longer visible. \\ \hline
\textbf{Object permanence: black cube} & This memory task is not using production rules. This has same world layout as the task above, only here the player is asked to find the black cube, which requires going to the left. \\
\hline
\end{longtable}

\renewcommand*{\arraystretch}{1.4}
\begin{longtable}{|>{\raggedright}p{4cm}|p{12cm}|}
\caption{Multi-agent probe tasks}
    \label{probe-task-desc-multi-agent}
    \\ \hline
\textbf{Name} & \textbf{Description} \\ \hline
\textbf{Pass over the wall} & The players are separated by an opaque wall. They cannot see each other, only their half of the world. The solution requires ``passing'' the accessible objects on either player's side to other player to combine them into the goal object. Then the players must make sure the correct player holds this object. \\ \hline
\textbf{Pass over the wall repeatedly} & Similar to `Pass over the wall' but here 2 out of 3 initial objects are `frozen' (cannot be moved). This prescribes a very specific solution strategy that requires the players to pass 3 objects over the wall in a specific order. \\ \hline
\textbf{Coordinated production} & This task requires each player to be near a sphere to turn this sphere into a pyramid, and then for both pyramids to be touching each other to create the goal object. The spheres are slightly hidden in a complex world, requiring some exploration. \\ \hline
\textbf{Coordinated production with deadends} & Like `Coordinated production' but here each player will destroy one of the initial objects if they get near it. These dead-end rules make this a much harder exploration problem. \\ \hline
\textbf{Coordinated exchange} & This requires each player to create a new object by holding an existing object, then to hold the object created by the other player to turn it into another intermediate object and finally for both objects to be combined into the goal object. While the world is fully accessible to both players, this can only be solved if both players actively participate. \\ \hline
\textbf{Overcooked: coordination ring} & Inspired by the video game Overcooked ~\citep{https://doi.org/10.48550/arxiv.1910.05789, strouse2021collaborating}, in this task both players need to ``serve tomato soup to a hungry patron''. This is implemented as a repeatable four-step production rule chain which requires both players to traverse their shared space (a circular layout) carefully in order to not block the other player. \\ \hline
\textbf{Overcooked: coordination ring, all rules visible} & While in `Overcooked: Coordination ring' all production rules are hidden from both players, here they are fully visible (to match the dynamics of the original Overcooked game). \\ \hline
\textbf{Overcooked: cramped room} & Similar to `Overcooked: coordination ring' but with a different layout for the shared space and a different number of initial objects, using different shapes and colours. \\ \hline
\textbf{Overcooked: cramped room, all rules visible} & While in `Overcooked: cramped room' all production rules are hidden from both players, here they are fully visible (to match the dynamics of the original Overcooked game). \\ \hline
\textbf{Overcooked: forced coordination} & Similar to the other `Overcooked' task variants, but here both players are restricted to only a certain part of the world and so are forced to coordinate to solve this task. No player can solve this alone since they cannot reach all initial objects. \\ \hline
\textbf{Overcooked: forced coordination, all rules visible} & While in `Overcooked: forced coordination' all production rules are hidden from both players, here they are fully visible (to match the dynamics of the original Overcooked game). \\ \hline
\textbf{Kickball} & This task is set in large world with two frozen pyramids on opposite sides of the world. Both players want to bring all of the plentiful purple spheres to the yellow pyramid. But lifting them destroys the pyramids so they need to ``kick'' them by bouncing them off the avatar. Think ``soccer practice''. \\ \hline
\textbf{Lemon eater} & The first player destroys all yellow spheres (of which there are many) when bumping into them. This is also the goal for both players. So they need to cooperate to bring all yellow spheres to the first player as quickly as possible. \\ \hline
\textbf{Careful lemon eater} & Like `Lemon eater' but any collision between two spheres turns them from yellow to purple. Purple spheres are ``not edible'' and so the players need to be careful to not create those, otherwise they will lose out on reward. \\ \hline
\textbf{Lemon eater and maker} & Like `Lemon eater' but we start out with only purple spheres. Only the second player can turn purple into yellow spheres by lifting them, effectively having to ``create food'' for the first player. \\ \hline
\textbf{Antimatter for two} & Identical in nature to the single-player `Antimatter' task but set in a different world layout and with two players who share the same goal. \\ \hline
\textbf{Antimatter with creation for two} & Identical in nature to the single-player `Antimatter with creation' task but set in a different world layout and with two players who share the same goal. \\ \hline
\textbf{Antimatter with copies for two} & Identical to `Antimatter for two' but here any two spheres of the same colour colliding leads to the creation of another sphere of that colour. This can set in motion runaway growth in the number of spheres, making it very hard to solve the task. \\ \hline
\textbf{Irreversible production for two} & Identical in nature to the single-player `Irreversible production' task but set in a different world layout and with two players who share the same goal. \\ \hline
\textbf{Irreversible production for two, all rules visible} & Like `Irreversible production for two' but with all production rules visible to both players. \\ \hline
\textbf{Wrong pair disappears for two} & Identical in nature to the single-player `Wrong pair disappears' task but set in a different world layout and with two players who share the same goal. \\ \hline
\textbf{Wrong pair disappears for two, partial hiding} & Like `Wrong pair disappears for two' but with only the input objects of the productions rules being hidden, the outputs and condition being visible. \\ \hline
\textbf{Crafting pyramid for two} & Identical in nature to the single-player `Crafting pyramid' task but set in a different world layout and with two players who share the same goal. \\ \hline
\textbf{Information asymmetry} & A simple world in which the players are asked to execute a two-step production rule in the presence of multiple dead ends. While the first player knows all the rules, they are completely hidden from the second player. This task is intended to measure a specific flavor of third-person imitation. \\ \hline
\textbf{Information asymmetry with repetition} & While `Information asymmetry' only allows for up to four completions, this task variant is (given unlimited time) infinitely repeatable, providing more opportunities for imitation. \\ \hline
\textbf{Combine outputs} & Similar in nature to `Coordinated production' but set in a larger world and (through the use of frozen objects) requiring both players to repeatedly navigate quite far to create input objects for shared creation. \\ \hline
\textbf{Two machines} & Many objects of different colours and shapes litter this world. The frozen yellow pyramid on one end of the world transforms these objects into an intermediary object. The players then need to bring the intermediary object to the frozen black pyramid on the other end of the world to ``cash it in'' for instantaneous reward, at which point the intermediary object is destroyed. Therefore to get more reward, the players must repeat this process. \\ \hline
\textbf{Two machines with information asymmetry} & Like `Two machines', but all rules are visible to one player, and all rules are hidden from the other player. This creates an information asymmetry and thus an opportunity for third-person imitation. \\ \hline
\end{longtable}

\subsection{Comparing human and agent scores on every probe task}
\label{app:human-scale-adaptation}

\begin{figure}[h!]
    \centering
    \includegraphics[width=1\linewidth]{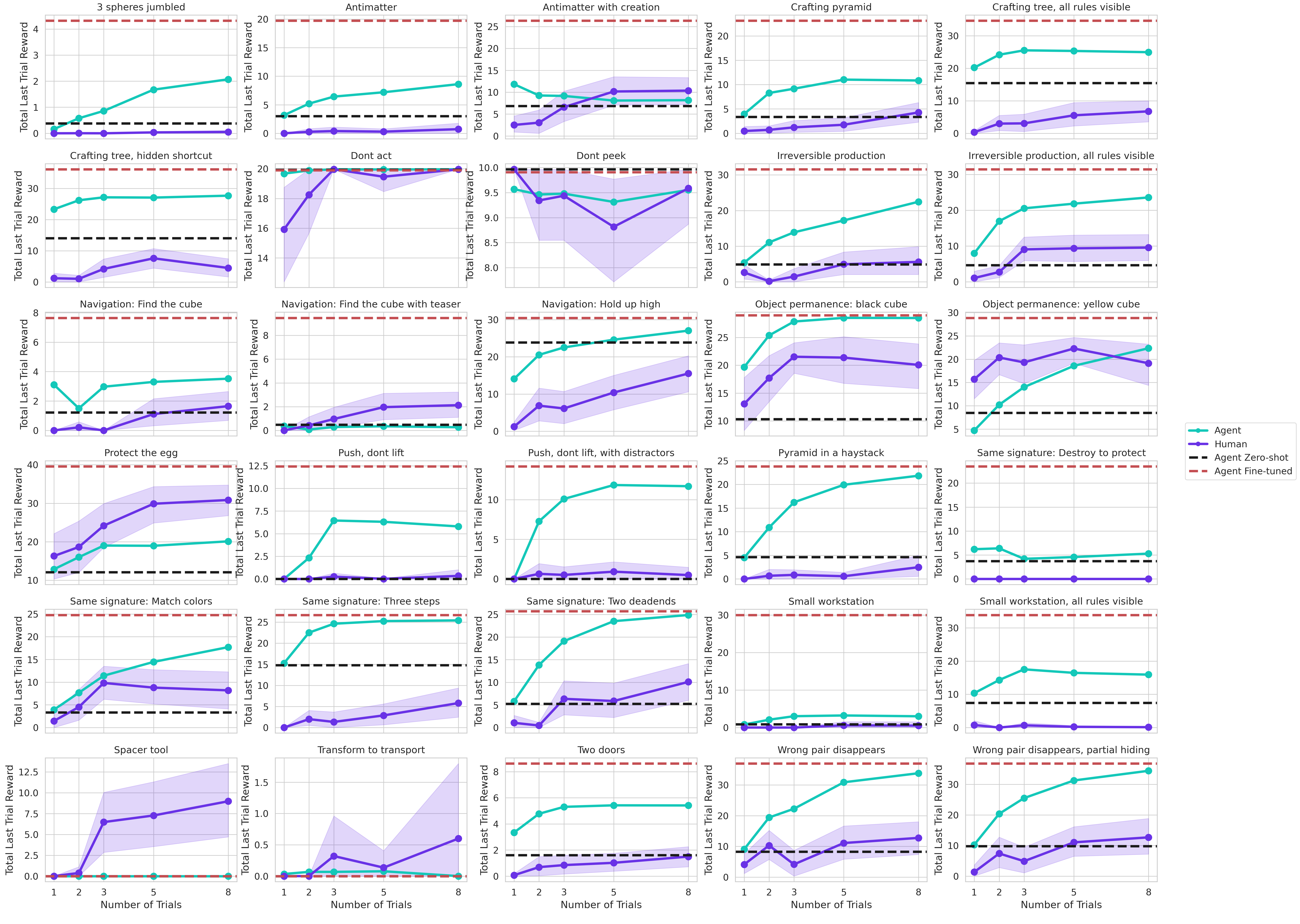}
    \caption{Comparison of AdA against 19 human players on each of the 30 held-out single-agent hand-authored tasks. The performance of a baseline agent trained to optimise zero-shot performance is shown as a black dashed line and indicates that AdA does not sacrifice its zero-shot generalisation to achieve adaptation. The reward of an agent fine-tuned on the hand-authored tasks is also shown as a red dashed line to provide some indication of the maximum reward achievable in a trial of each task.}
    \label{fig:human_scale_adaptation_all_tasks}
\end{figure}

In Figure \ref{fig:human_scale_adaptation_all_tasks} we show the raw last-trial total reward for humans and our agent as a function of number of trials across every one of the $30$ evaluation probe tasks. 
\subsection{Quantifying stochasticity}
\label{app:seed-variation}

\begin{figure}[b!]
    \centering
    \begin{subfigure}[b]{0.47\textwidth}
        \centering
        \includegraphics[width=\linewidth]{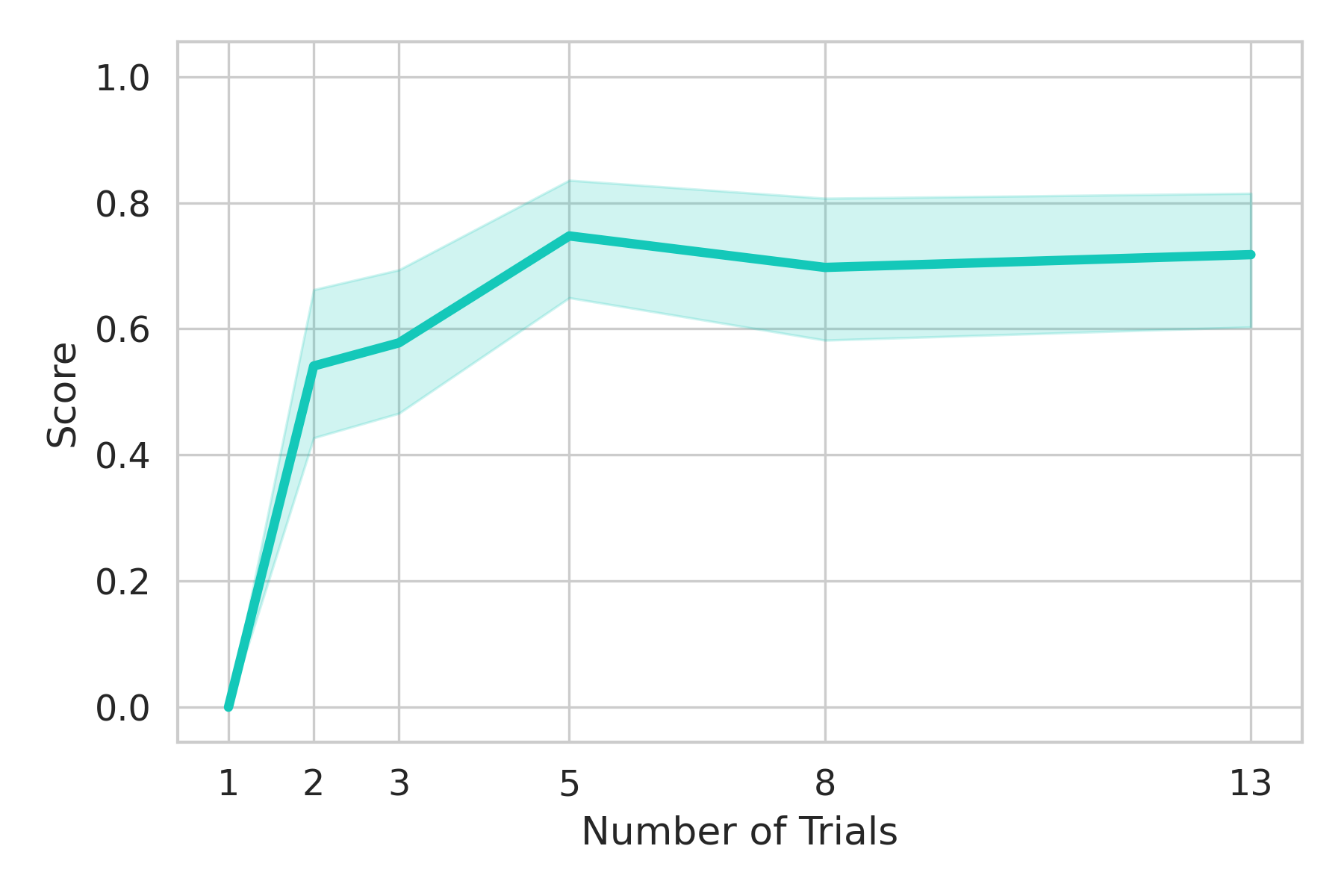}
        \vskip -3pt \caption{}
        \label{fig:task-variance}
    \end{subfigure}
    \begin{subfigure}[b]{0.47\textwidth}
        \centering
        \includegraphics[width=\linewidth]{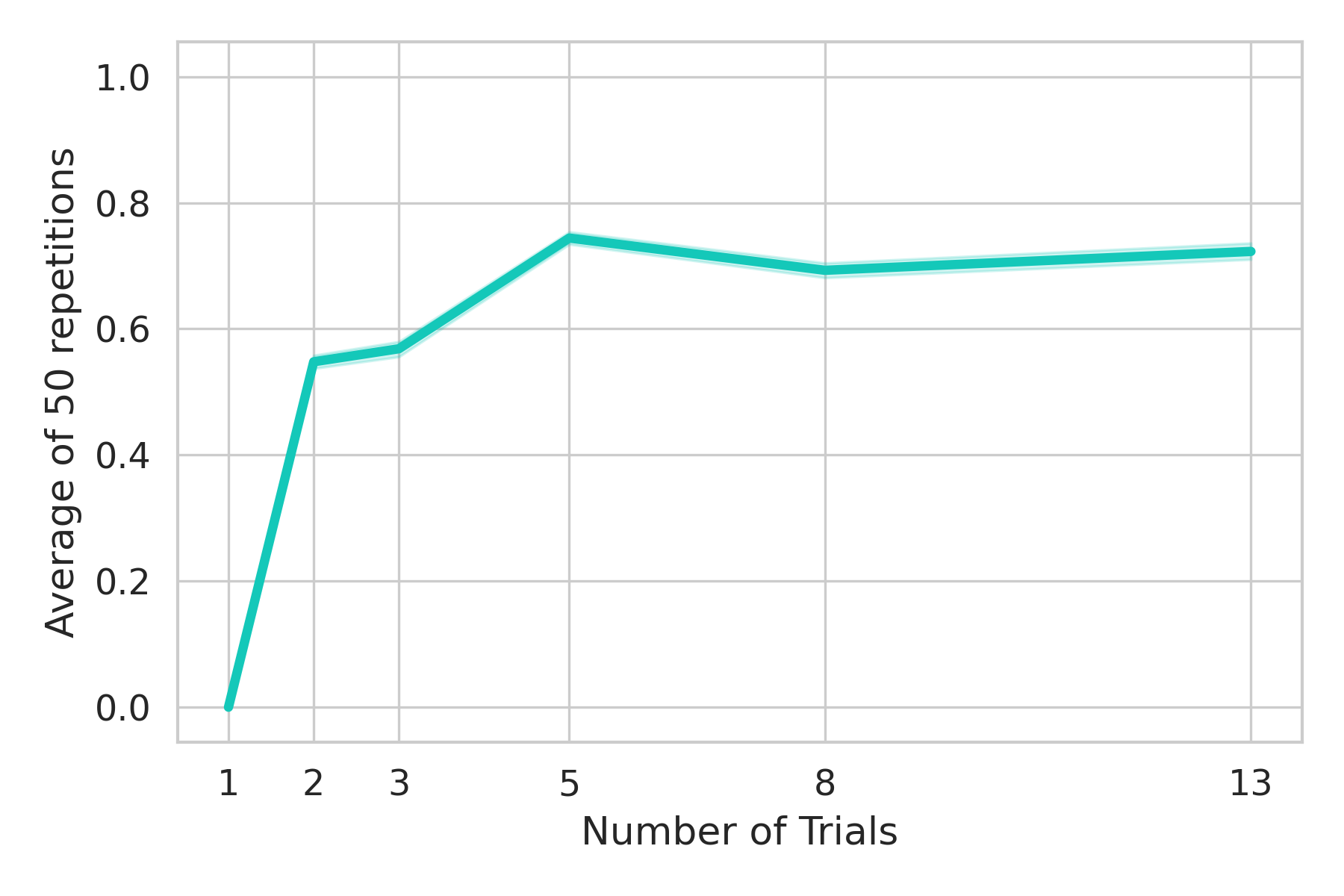}
        \vskip -3pt \caption{}
        \label{fig:task-variance-repetitions}
    \end{subfigure}

    \caption{
        \textbf{(a)} Mean single-repetition score and 95\% confidence intervals over 50 samples.
        \textbf{(b)} Mean of the aggregated 50-repetition score and 95\% confidence intervals over 50 samples.}
\end{figure}

\paragraph{Task variation.} Figure \ref{fig:task-variance} shows that there is fairly high variance in the score obtained by a single agent over 50 repetitions of a single task. This is due to random spawn rotations of the avatar following every environment reset and stochasticity in the agent's policy. To address this we run 50 repetitions of every (task, $k$) combination and average the score over these repetitions. This reduces the standard deviation to that shown in Figure \ref{fig:task-variance-repetitions}. Here, the standard deviation of the mean task score reduced from a maximum of 0.43 with 1 repetition, to a maximum of 0.06 with 50 repetitions.

\begin{figure}[htb]
    \centering
    \includegraphics[width=0.5\linewidth]{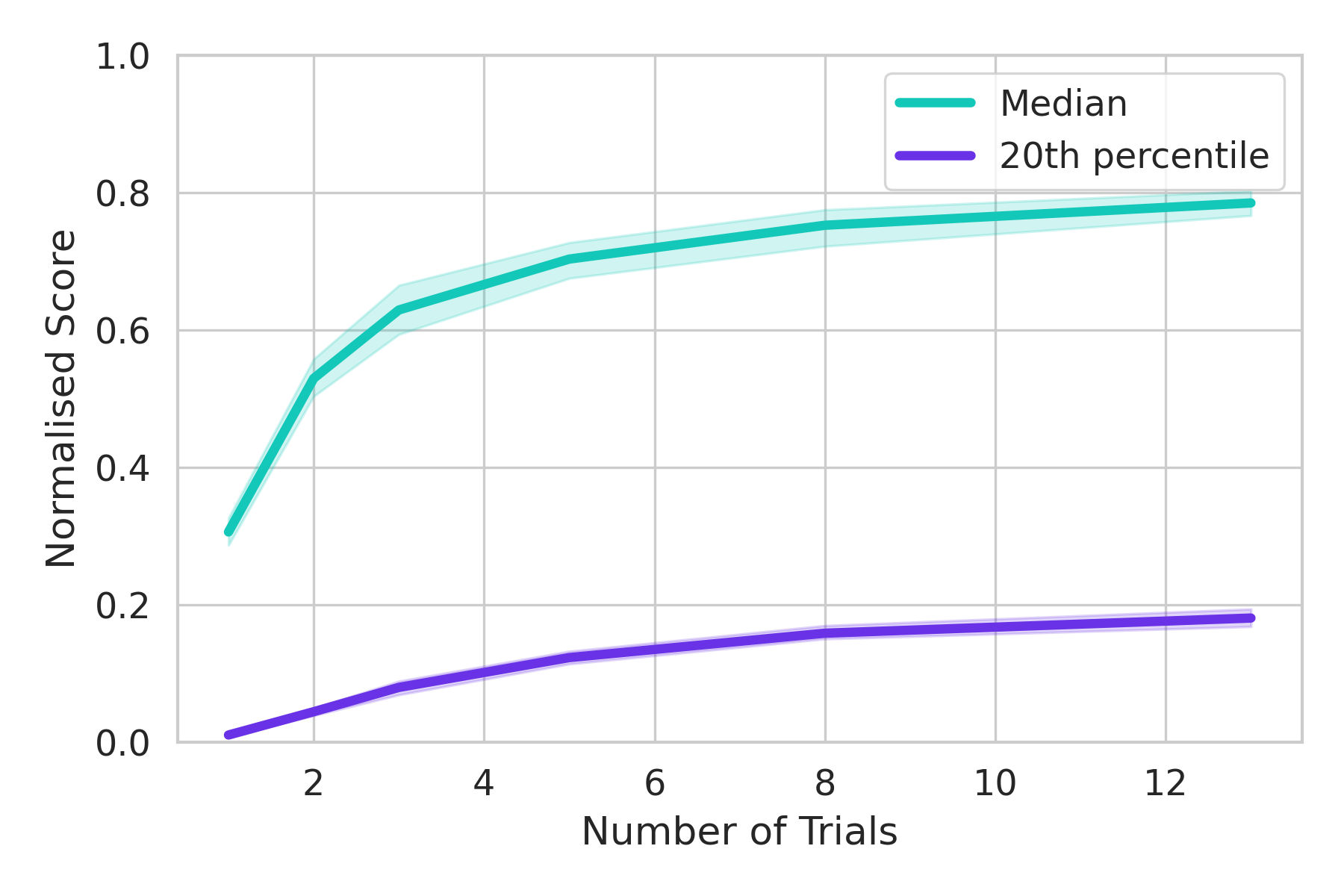}
    \caption{95\% bootstrap confidence intervals around the median and 20th percentile over our test set of 5 agents trained with different initialisation seeds. The maximum observed standard deviation was 0.04 around the median and 0.02 around the 20th percentile.}
    \label{fig:seed-variance}
\end{figure}

\paragraph{Agent initialisation variation.} After accounting for task variation, Figure \ref{fig:seed-variance} shows the variance due to agent initialisation seed during training. We plot the score as a function of $k$ on our test set for the 76M total parameter version of AdA with 5 different initialisation seeds. The maximum standard deviation observed for any number of trials in the median was 0.04 and for the 20\textsuperscript{th} percentile was 0.02. This low initialisation seed variance led us to run our ablations with one initialisation seed to save on compute. We note that the results shown in our ablation section have significantly larger than one standard deviation differences. 
\subsection{Prompting through first-person demonstrations}
\label{app:prompting}

\begin{figure}[t]
    \centering
    \includegraphics[width=1\linewidth]{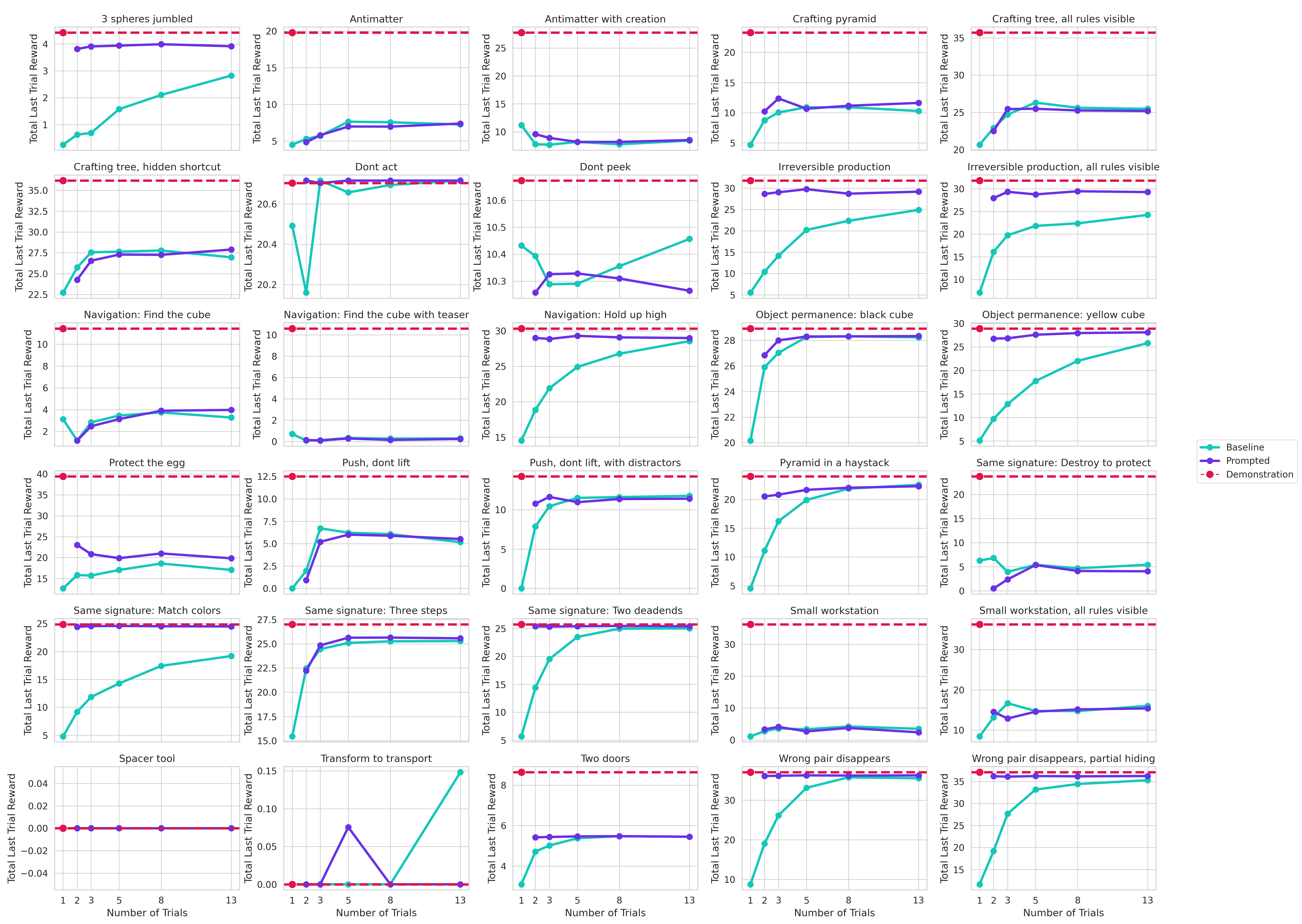}
    \caption{A comparison of a prompted agent and unprompted baseline for our full set of 30 hand-authored single-agent tasks. The dashed red lines indicate the score obtained by the teacher providing the first-person demonstration to the prompted agent in the first trial.}
    \label{fig:prompting-all-tasks}
\end{figure}

Figure \ref{fig:prompting-all-tasks} shows the performance of AdA prompted with a fine-tuned agent compared to an unprompted baseline on each of the 30 single agent hand-authored probe tasks. The figure reveals a set of tasks on which AdA is able to leverage information in the prompt, resulting in perfect or near perfect scores. There are also tasks where AdA does not seem to be able to do this. In all but one case, prompting does not hurt performance.

Analysing the tasks in Figure \ref{fig:prompting-all-tasks} suggests that prompting is useful for short navigational tasks such as \texttt{Navigation: hold up high} in which the agent follows a short and simple path to reach the goal object. Prompting does not, however, improve performance for longer and more complex navigation tasks like \texttt{Navigation: find the cube}, likely due to the full demonstration being too long to fit in the agent’s memory context.

We observe a similar pattern in tasks involving production rules. For tasks with up to 2 production rules in the solution path, such as \texttt{Same signature: match colors}, we observe the unprompted agent exploring different objects to determine the correct rule to trigger. When prompted with a demonstration it subsequently triggers the correct rule immediately and achieves a perfect score. An exception to this is \texttt{Same signature: destroy to protect} where one of the production rules involves destroying an object, which the agent does not appear to remember from the demonstration. For tasks using 3 or more production rules like \texttt{Same signature: three steps} (3 production rules in the solution path) and \texttt{Small workstation} (5 production rules), the agent tends to only remember a subset of the rules to trigger and continues engaging in exploratory behaviour following the demonstration. The performance on these tasks tend to match the unprompted baseline.

Another factor appearing to influence the effectiveness of prompting is the topology and configuration of objects in the world, as seen in the \texttt{Small workstation} tasks. While the teacher demonstrations for these tasks present a clean trajectory, the agent subsequently knocks into and displaces distractor objects, leading to environment states not observed during the demonstration (and thus not recallable from memory). On the other hand, a favourable configuration of objects appears to make the demonstration easier to learn from, as observed in \texttt{Irreversible production}. Here the objects required to trigger the last production rule are positioned close together. The agent shows optimal behaviour here despite the task requiring 3 production rules on the solution path. The agent also appears unable to infer from prompting certain more subtle requirements like relative positioning of objects or tool use. This is observed in tasks like \texttt{Antimatter} in which prompted AdA is able to trigger the destruction rule but we did not observe it immediately hiding objects from each other. 

Prompting may also help eliminate biases that the agent may have acquired during training. This is reflected in the lower score obtained by the unprompted agent for small $k$ in \texttt{Object permanence: yellow cube} compared to \texttt{Object permanence: black cube}. These tasks are identical except for the goal being to navigate to a yellow cube on the right, or a black cube on the left respectively. This suggests that the agent may have acquired a bias during training to either prefer black cubes or to navigate towards objects on its left. The significantly higher prompted scores on \texttt{Object permanence: yellow cube} for small $k$ suggest that a prompt may help the agent overcome these biases.

\begin{figure}[t!]
    \centering
    \includegraphics[width=1\linewidth]{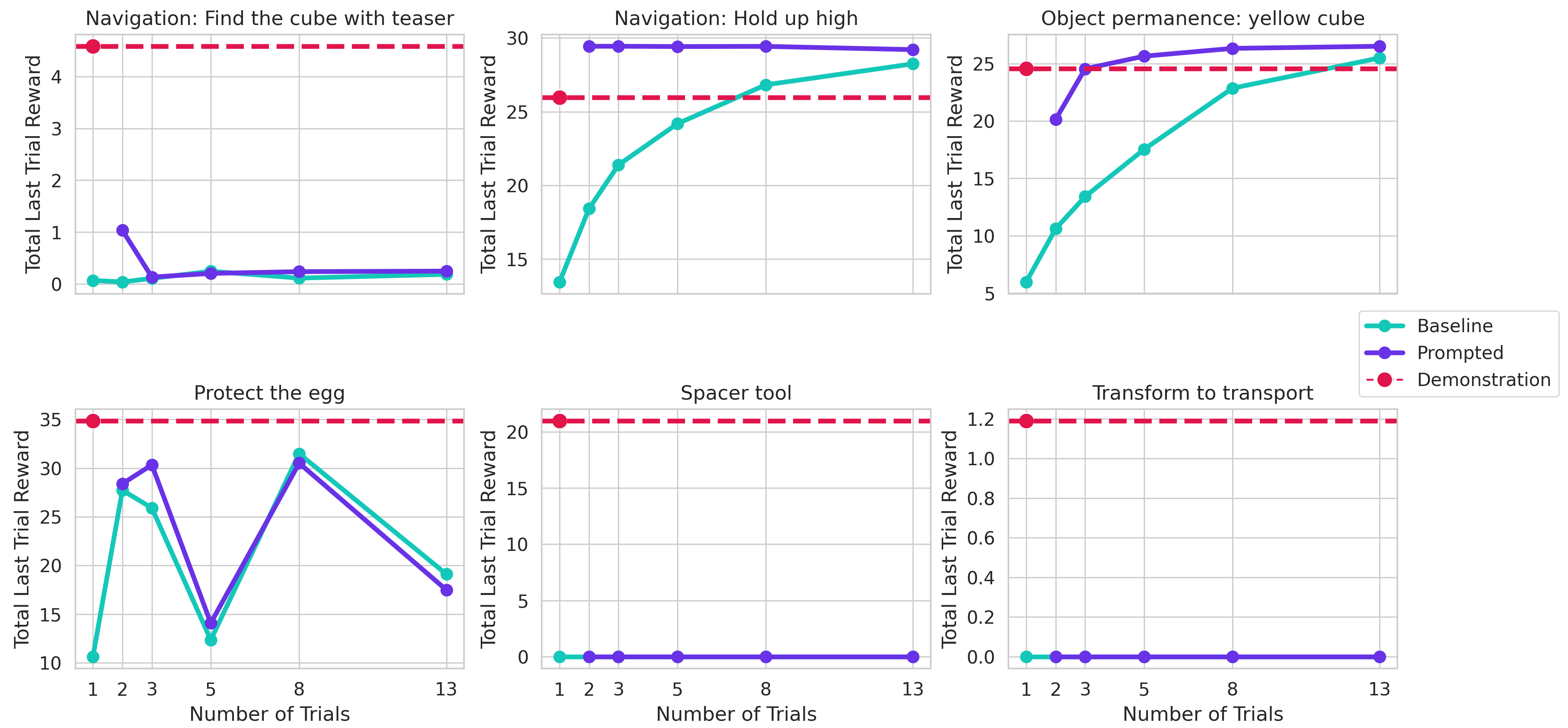}
    \caption{Performance of AdA on 6 hand-authored tasks when prompted with an expert first-person human demonstration, compared with an unprompted baseline.}
    \label{fig:prompting-with-humans}
\end{figure}

\begin{figure}[t!]
    \centering
    \includegraphics[width=0.6\linewidth]{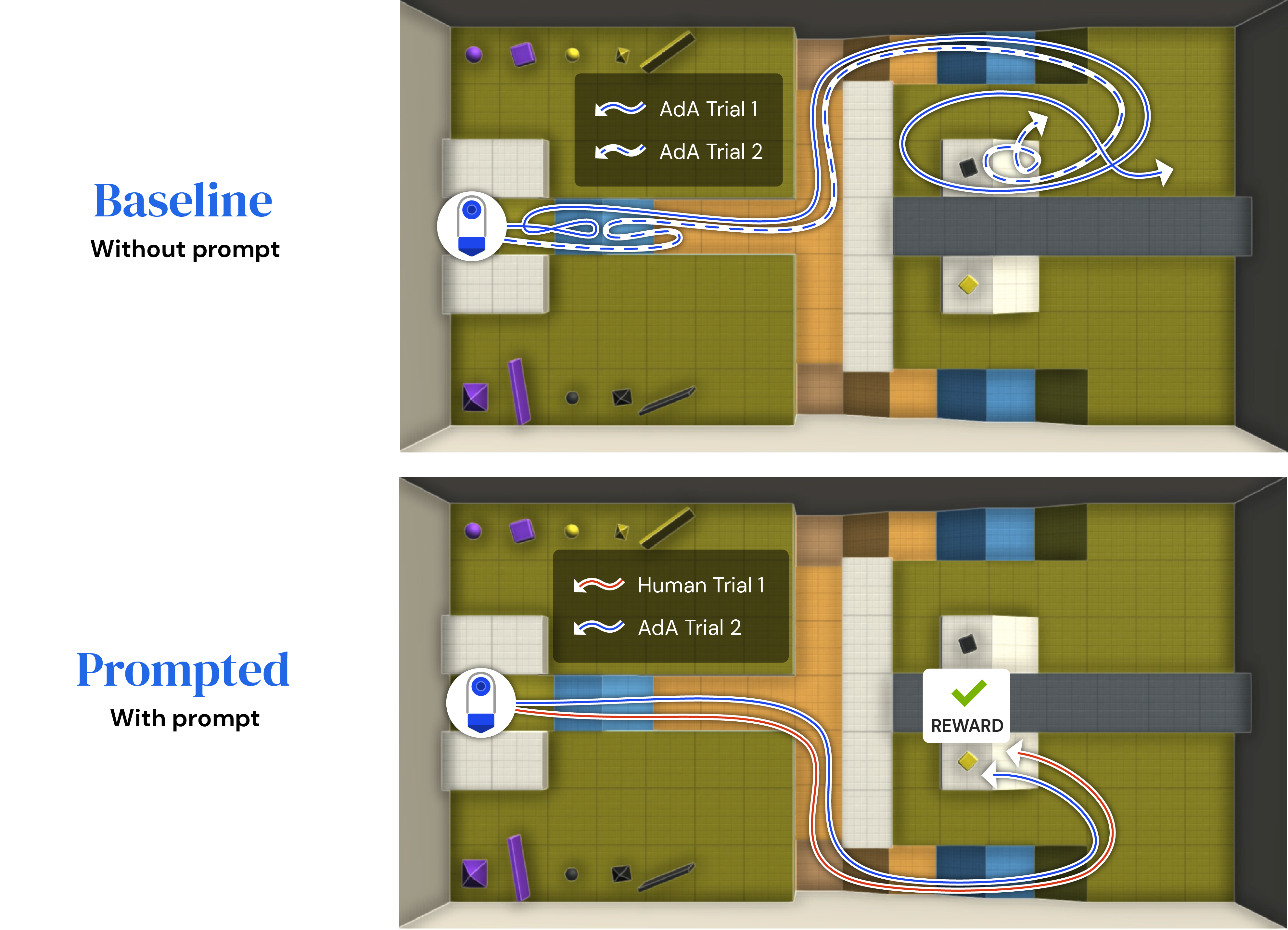}
    \caption{Top-down views depicting the behaviour of AdA with and without a human expert prompt on the task \texttt{Object permanence: yellow cube}. On its own, AdA appears to have a bias of navigating to the black cube which is a dead end in this task. When prompted with a human (or fine-tuned) expert trajectory, AdA is able to overcome this bias and navigate to the yellow cube in the second trial.}
    \label{fig:prompting-trajectory}
\end{figure}

\paragraph{Prompting with human demonstrations.} We prompted AdA with expert human demonstrations in a small selection of 6 hand-authored tasks, depicted in Figure \ref{fig:prompting-with-humans}. These tasks were chosen to be a mixture of tasks where AdA excelled with fine-tuned teacher prompting, where it failed with fine-tuned teacher prompting and where even the fine-tuned teacher failed. 

The results show the same pattern as those obtained when prompting with a fine-tuned teacher. In both \texttt{Navigation: hold up high} and \texttt{Object permanence: yellow cube}, prompted AdA achieved close to optimal performance, exceeding both the baseline and the human demonstration. Figure \ref{fig:prompting-trajectory} depicts the latter behaviour in detail. AdA continues to fail to learn from a demonstration in \texttt{Navigation: find the cube with teaser} and in both \texttt{Spacer tool} and \texttt{Transform to transport}, which our fine-tuned teacher also failed at. A successful human demonstration did not unlock any capabilities AdA was previously not capable of demonstrating, suggesting that these tasks are perhaps too far out-of-distribution with respect to AdA's training tasks. The fact that prompting with off-policy human demonstrations is partially successful is worthy of note, and opens an exciting area of future research. 

\end{appendix}
\end{document}